\documentclass[journal]{IEEEtran}
\usepackage{amsmath,amsfonts}
\usepackage{array}
\usepackage[caption=false,font=normalsize,labelfont=sf,textfont=sf]{subfig}
\usepackage{enumitem}
\usepackage{textcomp}
\usepackage{stfloats}
\usepackage{verbatim}
\usepackage{graphicx}
\graphicspath{{./figures/}}
\usepackage{balance}
\usepackage{hyperref}
\usepackage{xspace}
\usepackage[ruled,vlined,linesnumbered]{algorithm2e}
\usepackage{placeins}  
\usepackage[capitalize]{cleveref}  
\usepackage{tikz}
\usetikzlibrary{positioning,calc,arrows.meta}

\newcommand{\titleOfArticle}{Estimating temperatures with low-cost infrared cameras using deep neural networks}

\newcommand{\taucamera}{\href{https://www.flir.com/products/tau-2/}{\texttt{Tau2}}\xspace}
\newcommand{\blackbody}{\href{https://www.ci-systems.com/sr-800n-superior-accuracy-blackbody}{\texttt{SR-800N}}\xspace}
\newcommand{\scientificCamera}{\href{https://www.flir.com/products/a655sc/}{\texttt{A655sc}}\xspace}
\newcommand{\campbellCtrl}{\href{https://www.campbellsci.com/cr1000}{\texttt{Campbell CR1000}}\xspace}
\newcommand{\opPoint}{operating~point\xspace}
\newcommand{\opPoints}{operating~points\xspace}
\newcommand{\Tamb}{T_{amb}}
\newcommand{\tamb}{t_{amb}}
\newcommand{\Tobj}{T_{obj}}
\newcommand{\tobj}{t_{obj}}
\newcommand{\tobjapprox}{\Tilde{t}_{obj}}
\DeclareMathOperator*{\argmin}{argmin}
\newcommand{\mat}[1]{ \underline{\underline{#1}}}
\newcommand{\nChannels}{\delta}
\newcommand{\nLevels}{\alpha}
\newcommand{\TambAsVec}{\left[\Tamb[0],...,\Tamb[\NumFPA]\right]}

\Crefname{figure}{Fig.}{Figs.}
\crefname{section}{Section}{Sections}
\Crefname{table}{Table}{Tables}
\Crefname{equation}{Eq.}{Equations}

\newcommand{\sizePatches}{0.164}
\newcommand{\figPathces}[1]{
    \subfloat{\includegraphics[width=\sizePatches\linewidth]{patches/#1/sample.pdf}}
    \hfill
    \subfloat{\includegraphics[width=\sizePatches\linewidth]{patches/#1/label.pdf}}
    \hfill
    \subfloat{\includegraphics[width=\sizePatches\linewidth]{patches/#1/gxpd.pdf}}
    \hfill
    \subfloat{\includegraphics[width=\sizePatches\linewidth]{patches/#1/he.pdf}}
    \hfill
    \subfloat{\includegraphics[width=\sizePatches\linewidth]{patches/#1/admire.pdf}}
    \hfill
    \subfloat{\includegraphics[width=\sizePatches\linewidth]{patches/#1/snrwdnn.pdf}}
}
\newcommand{\figPathcesCaption}[1]{
    \subfloat[Sample]{\includegraphics[width=\sizePatches\linewidth]{patches/#1/sample.pdf}}
    \hfill
    \subfloat[Ground Truth]{\includegraphics[width=\sizePatches\linewidth]{patches/#1/label.pdf}}
    \hfill
    \subfloat[GxPD (ours)]{\includegraphics[width=\sizePatches\linewidth]{patches/#1/gxpd.pdf}}
    \hfill
    \subfloat[ADMIRE~\cite{admire}]{\includegraphics[width=\sizePatches\linewidth]{patches/#1/admire.pdf}}
    \hfill
    \subfloat[He et al.~\cite{He2018}]{\includegraphics[width=\sizePatches\linewidth]{patches/#1/he.pdf}}
    \hfill
    \subfloat[SNRWDNN~\cite{snrwdnn}]{\includegraphics[width=\sizePatches\linewidth]{patches/#1/snrwdnn.pdf}}
    }

\newcommand{\sizePatchesSuppMat}{0.164}
\newcommand{\figFloatsCaptionSuppMat}[2]{
    \setcounter{subfigure}{0}
    \subfloat[Sample]{\includegraphics[width=\sizePatchesSuppMat\linewidth]{supp_figures/#1/#2/sample.pdf}}
    \hfill
    \subfloat[Ground Truth]{\includegraphics[width=\sizePatchesSuppMat\linewidth]{supp_figures/#1/#2/label.pdf}}
    \hfill
    \subfloat[GxPD (ours)]{\includegraphics[width=\sizePatchesSuppMat\linewidth]{supp_figures/#1/#2/gxpd.pdf}}
    \hfill
    \subfloat[ADMIRE~\cite{admire}]{\includegraphics[width=\sizePatchesSuppMat\linewidth]{supp_figures/#1/#2/admire.pdf}}
    \hfill
    \subfloat[He et al.~\cite{He2018}]{\includegraphics[width=\sizePatchesSuppMat\linewidth]{supp_figures/#1/#2/he.pdf}}
    \hfill
    \subfloat[SNRWDNN~\cite{snrwdnn}]{\includegraphics[width=\sizePatchesSuppMat\linewidth]{supp_figures/#1/#2/snrwdnn.pdf}}
}
\newcommand{\figFloatsSuppMat}[2]{
    \subfloat{\includegraphics[width=\sizePatchesSuppMat\linewidth]{supp_figures/#1/#2/sample.pdf}}
    \hfill
    \subfloat{\includegraphics[width=\sizePatchesSuppMat\linewidth]{supp_figures/#1/#2/label.pdf}}
    \hfill
    \subfloat{\includegraphics[width=\sizePatchesSuppMat\linewidth]{supp_figures/#1/#2/gxpd.pdf}}
    \hfill
    \subfloat{\includegraphics[width=\sizePatchesSuppMat\linewidth]{supp_figures/#1/#2/admire.pdf}}
    \hfill
    \subfloat{\includegraphics[width=\sizePatchesSuppMat\linewidth]{supp_figures/#1/#2/he.pdf}}
    \hfill
    \subfloat{\includegraphics[width=\sizePatchesSuppMat\linewidth]{supp_figures/#1/#2/snrwdnn.pdf}}
}
\newcommand{\figPatchesLabelSuppMat}[7]{
    \centering
    \figFloatsSuppMat{#1}{#2}\\
    \figFloatsSuppMat{#1}{#3}\\
    \figFloatsSuppMat{#1}{#4}\\
    \figFloatsSuppMat{#1}{#5}\\
    \figFloatsSuppMat{#1}{#6}\\
    \figFloatsCaptionSuppMat{#1}{#7}
    \caption{}
}
\newcommand{\figPatchesFiveSuppMat}[5]{
    \centering
    \figFloatsSuppMat{#1}{#2}\\
    \figFloatsSuppMat{#1}{#3}\\
    \figFloatsSuppMat{#1}{#4}\\
    \figFloatsCaptionSuppMat{#1}{#5}
    \caption{}
}
\newcommand{\coefficientsPixelwise}[1]{\Tilde{b}_{#1}}
\newcommand{\coefficientsPixelwiseVec}{\underline{\coefficientsPixelwise{}}}
\newcommand{\coefficientsAmbRadii}[1]{\Tilde{\mathcal{B}}_{#1}}
\newcommand{\coefficientsAmbRadiiVec}{\underline{\coefficientsAmbRadii{}}}
\newcommand{\coefficientsRadii}[1]{\mathcal{B}_{#1}}
\newcommand{\coefficientsRadiiVec}{\underline{\coefficientsRadii{}}}
\newcommand{\coefficientsRadiance}[1]{\underline{\alpha}_{#1}}
\newcommand{\coefficientsSpatial}{\underline{\beta}}
\newcommand{\coefficientsSpatialRaw}{\underline{\beta}_R}
\newcommand{\coefficientsSpatialQuadratic}{\coefficientsSpatial_Q}
\newcommand{\coefficientsSpatialFine}{\coefficientsSpatial_F}
\newcommand{\coefficientsSpatialSkewless}{\coefficientsSpatial}
\newcommand{\simulatedCameraResponse}[2]{\Tilde{R}(#1, #2)}
\newcommand{\mGrayLevels}{{M_{GL}}}
\newcommand{\NumTempObj}{{L_{\tobj}}}
\newcommand{\mSpatial}{{M_{spatial}}}
\newcommand{\mRadial}{{M_{radial}}}
\newcommand{\mFPA}{{M_{ambient}}}
\newcommand{\NumFPA}{{L_{\tamb}}}  

\begin{document}
\title{\titleOfArticle}
\author{\href{mailto:navotoz@mail.tau.ac.il}{Navot Oz},
    \href{mailto:sochen@math.tau.ac.il}{Nir Sochen},
    \href{mailto:mend@eng.tau.ac.il}{David Mendelovich},
    and \href{mailto:iftach@volcani.agri.gov.il}{Iftach Klapp}
    \thanks{\href{mailto:navotoz@mail.tau.ac.il}{Navot Oz}
        and \href{mailto:mend@eng.tau.ac.il}{David Mendelovich} are with the School of Electrical Engineering, Tel Aviv University, Tel Aviv 69978, Israel.}%
    \thanks{\href{mailto:sochen@math.tau.ac.il}{Nir Sochen} is with the Department of Mathematics, Tel Aviv University, Tel Aviv 69978, Israel.}%
    \thanks{\href{mailto:navotoz@mail.tau.ac.il}{Navot Oz} and \href{mailto:iftach@volcani.agri.gov.il}{Iftach Klapp} are with the Department of Sensing, Information and Mechanization engineering, Agricultural Research Organization, Volcani Institute, P.O. Box 15159, Rishon LeZion 7505101, Israel.}%
    \thanks{This work was supported by the Israeli Ministry of Agriculture’s Kandel Program (grant number 20-12-0018).}}
\maketitle
\begin{abstract}
    Low-cost thermal cameras are inaccurate (usually $\pm 3^\circ C$) and have space-variant nonuniformity across their detector. Both inaccuracy and nonuniformity are dependent on the ambient temperature of the camera. The goal of this work was to estimate temperatures with low-cost infrared cameras, and rectify the nonuniformity.

    A nonuniformity simulator that accounts for the ambient temperature was developed. An end-to-end neural network that incorporates both the physical model of the camera and the ambient camera temperature was introduced.
    The neural network was trained with the simulated nonuniformity data to estimate the object's temperature and correct the nonuniformity, using only a single image and the ambient temperature measured by the camera itself.
    Results of the proposed method significantly improved the mean temperature error compared to previous works by up to $0.5^\circ C$.
    In addition, constraining the physical model of the camera with the network lowered the error by an additional $0.1^\circ C$.

    The mean temperature error over an extensive validation dataset was $0.37^\circ C$. The method was verified on real data in the field and produced equivalent results.
\end{abstract}
\section{Introduction}\label{sec:intro}
    \IEEEPARstart{I}{nfrared} (IR) imagery in the $8_{\mu m}-14_{\mu m}$ atmospheric window measures the thermal radiation emitted from an object. IR imagery is extensively used for various applications, such as - military night vision \cite{bolometer_night_vision}, medical fever screening \cite{bolometer_medial_screening} and machinery fault diagnosis \cite{bolometer_fault_diagnosis}, among many others. One interesting utilization of such an imaging system is agriculture, because the temperature of a plant is important in deducing information on its well-being \cite{ir_agri_est_crop_water, ir_agri_dought}.
\begin{figure}
    \centering
    \includegraphics[width=\linewidth]{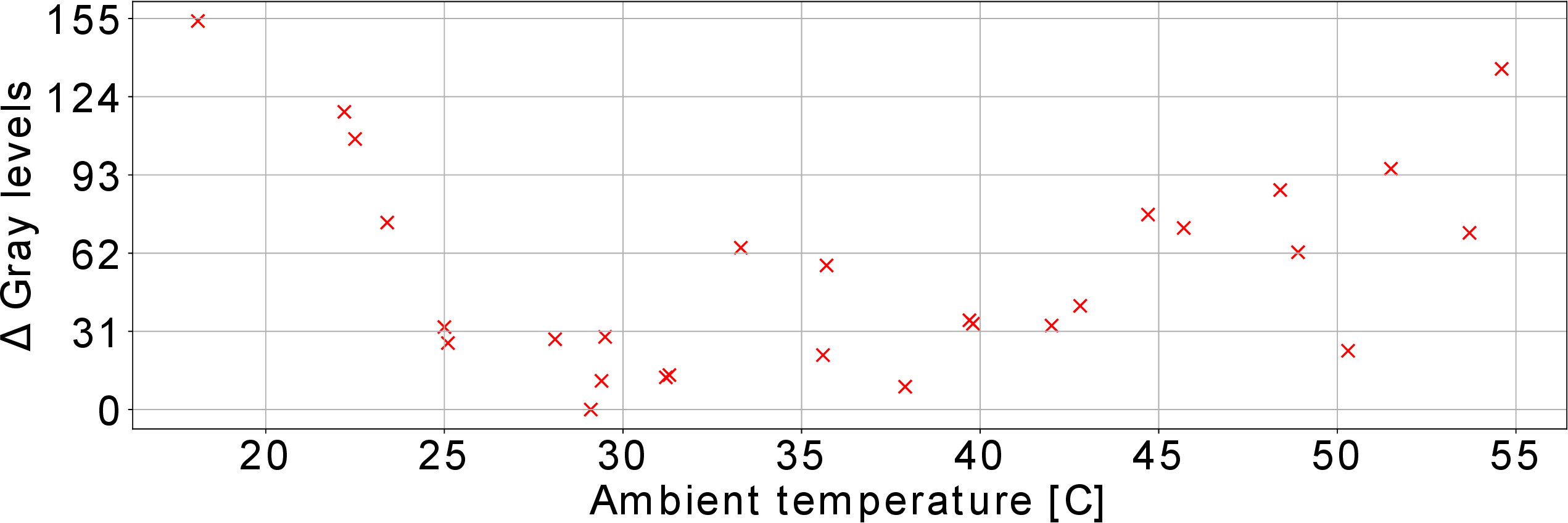}
    \caption{Change in gray-levels as a function of the ambient temperature, for a constant object temperature of $40^\circ C$.
    The measurements are of a scientific-grade \blackbody blackbody, taken with a low-cost \taucamera camera at different ambient temperatures.
    Each measurement was taken after performing the internal flat-field correction (FFC) calibration procedure.}
    \label{fig:intro:GraylevelsAsFunctionOfFPAs}
\end{figure}

Low-cost IR cameras are usually uncooled, and rely on microbolometer arrays as sensors. The microbolometer array enables the construction of inexpensive IR cameras with low energy requirements. Unlike the photon-counting detector arrays (e.g., CMOS in the visible range), microbolometers measure changes in the electrical resistance caused by the incident thermal radiation originating from an object \cite{bolometer}. The thermal radiation heats each microbolometer to a temperature that depends on the scene, and each microbolometer in the array has a slightly different temperature depending on the observed scene and the incident angle of the radiation. The resistance of each microbolometer changes according to the scene temperature. The minuscule changes in the resistance of each microbolometer in the array are used to construct an image corresponding to the temperature of the observed scene.

While accurate microbolometer-based IR camera exists (e.g., \scientificCamera which will be used extensively in this work), they are expensive and require complex calibration procedures~\cite{kruseIR}.
Low-cost IR cameras temperature estimation accuracy is usually low.
\cref{fig:intro:GraylevelsAsFunctionOfFPAs} demonstrates the effect of the camera's ambient temperature on the measurements made by a typical low-cost IR camera \taucamera. Notice that the change is dependent on the ambient temperature, as well as non-linear.
\begin{figure}
    \centering
    \includegraphics[width=\linewidth]{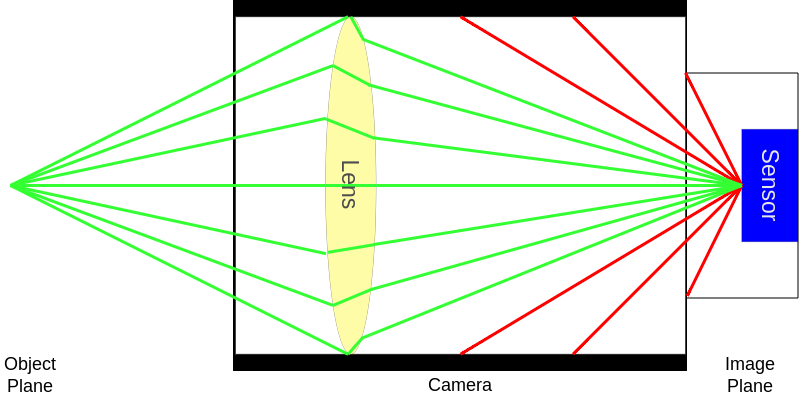}
    \caption{Side view of a thermal camera. Green lines are thermal radiation propagating from the object plane to the sensor plane. Red lines are thermal radiation emitted by the camera itself, which has a major effect on nonuniformity~\cite{Schulz1995}.}
    \label{fig:intro:selfHeating}
\end{figure}

\subsection{Nonuniformity and noise}\label{sec:intro:nonuniformity}
Microbolometer arrays are subject to space-variant nonuniformity and noise from various sources.
The microbolometer array is uncooled, and so a prominent source of nonuniformity is thermal radiation emitted by the camera itself~\cite{kruseIR}.
Another parasitic thermal radiation source is the narcissus effect, where unfocused reflection of the detector returns from the optical surfaces~\cite{infraredTheramlImaging}. The effect of the internal self-radiation (red lines) mixed with the incident thermal radiation from the scene (green lines) is schematically presented in \cref{fig:intro:selfHeating}.

These parasitic effects are dependent on ambient temperatures, meaning that their influence on the measurements changes with the environmental conditions of the camera.
\cref{fig:intro:ConstBlackbodyDiffFPAs} demonstrates the nonuniformity effect. The images presented are of a scientific-grade blackbody target set at a constant temperature of $40^\circ C$, and were taken with a \taucamera camera.
\cref{fig:intro:ConstBlackbodyDiffFPAs:18} was taken at an ambient temperature of $18.1^\circ C$, and \cref{fig:intro:ConstBlackbodyDiffFPAs:53} was taken at an ambient temperature of $53.7^\circ C$.
Two effects are visible in the images. First, the gray-levels of the image are dependent on the ambient temperature. Second, the nonuniformity is spatially variant.
The effect of the ambient temperature on the middle horizontal line of both images is illustrated in \cref{fig:intro:ConstBlackbodyDiffFPAs:lineCmp}, with both lines on the same axis.
The nonuniformity exhibits spatial variation, as evidenced by the different relations between the lines for each pixel.
\newcommand{\heightCmpGlDiffFPAs}{10em}
\begin{figure}
    \centering
    \subfloat[$\tamb=18.1^\circ C$]{\includegraphics[height=\heightCmpGlDiffFPAs]{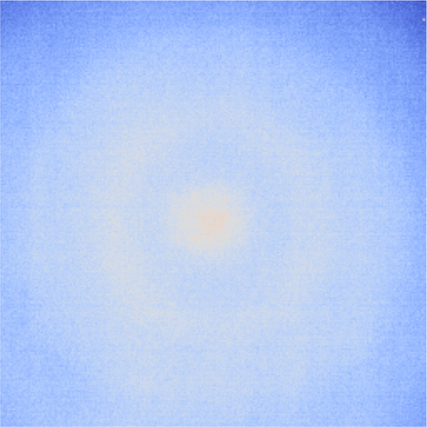}\label{fig:intro:ConstBlackbodyDiffFPAs:18}}
    \hfill
    \subfloat[$\tamb=53.7^\circ C$]{\includegraphics[height=\heightCmpGlDiffFPAs]{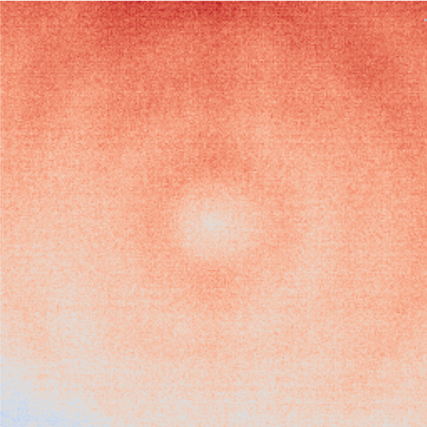}\label{fig:intro:ConstBlackbodyDiffFPAs:53}}
    \hfill
    \subfloat{\includegraphics[height=\heightCmpGlDiffFPAs]{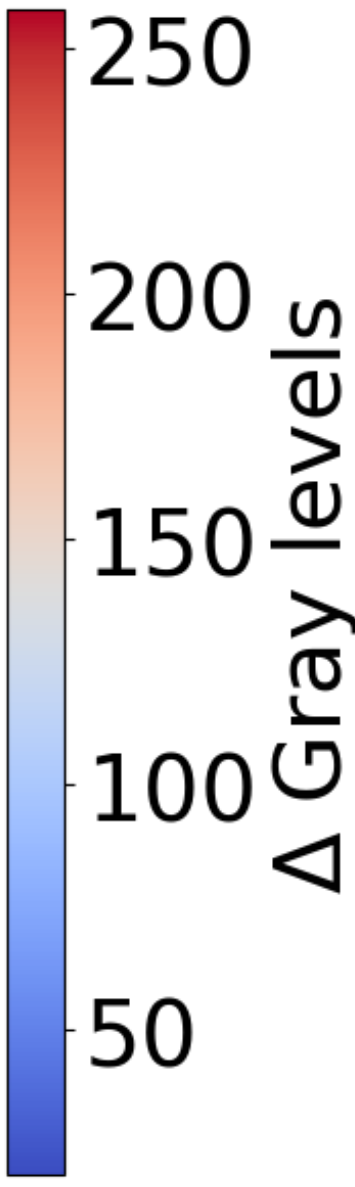}}
    \\
    \setcounter{subfigure}{2} 
    \subfloat[]{\includegraphics[width=\linewidth]{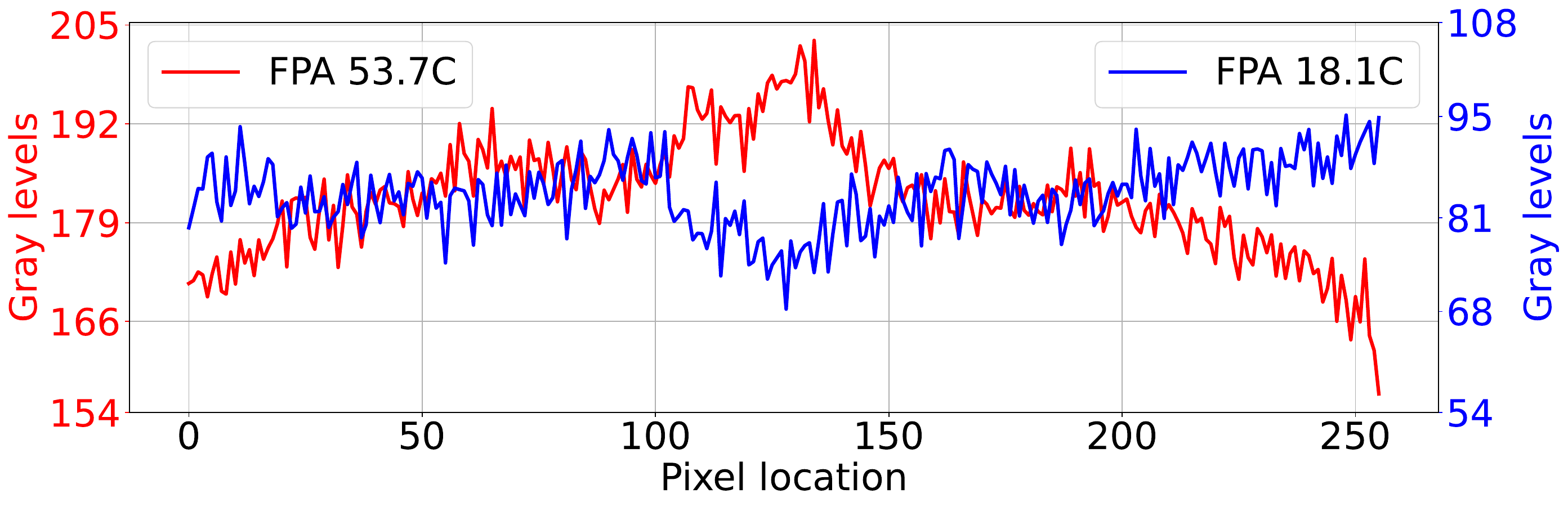}\label{fig:intro:ConstBlackbodyDiffFPAs:lineCmp}}
    \caption{Images of a uniform heat source \blackbody blackbody at different ambient temperatures.
    The blackbody is set to $40^\circ C$ for both images.
    (a) was the response at the ambient temperature $\tamb=18.1^\circ C$ and (b) was the response at the ambient temperature $\tamb=53.7^\circ C$.
    (c) is the middle horizontal lines of (a) and (b).
    }
    \label{fig:intro:ConstBlackbodyDiffFPAs}
\end{figure}

Another source of nonuniformity is fixed-pattern noise (FPN). The readout circuitry of the microbolometer array is usually line-based (similar to charge coupled devices). Slight changes between readers on the same array can lead to considerable disparity between lines on the image \cite{Riou2004}.

Finally, the signal-to-noise ratio of the camera is often low due to readout and electronic noise~\cite{kruseIR}.

\subsection{Image acquisition model}
The thermal radiation emitted by a body for all wavelengths can be found using the Stefan-Boltzmann law, whereby the emitted radiation can be approximated by the fourth power of the object's temperature~\cite{kruseIR}:
\begin{equation}\label{eq:stefanBoltzmann}
    L(T)\approx\epsilon\sigma T^4  \quad \left[   \frac{W}{m^2}   \right]
\end{equation}
where $T$ is the object's temperature, $\epsilon$ is a proportional constant and $\sigma$ is the Boltzmann constant.

In a small environment near a reference temperature $T_0$, the Stefan-Boltzmann law can be expanded by Taylor series:
\begin{equation}\label{eq:stefanBoltzmannTaylor}
    \begin{split}
        L(T)&=\epsilon\cdot\sigma T^4=\epsilon\cdot\sigma(T_0+\Delta T)^4\\
        &\approx\epsilon\cdot\sigma(T_0)^4+4\epsilon\cdot\sigma(T_0)^3\Delta T
        \approx a_1\cdot\tobj + a_0
    \end{split}
\end{equation}
where $a_0$, $a_1$ are the coefficients and $T_0$ is a reference temperature. $\Delta T$ was changed to $\tobj$ for brevity.

\cref{eq:stefanBoltzmannTaylor} demonstrates that the radiation can be approximated as linear in scene temperature for a small environment around a reference temperature. This result means that the incident thermal radiation on the sensor has a temperature-\textit{dependent} element and a temperature-\textit{independent} element.

The ambient temperature of the camera has a profound effect on the measurements that it produces. 
Thus, the model in \cref{eq:stefanBoltzmannTaylor} must also account for changes in ambient temperature.
The linear approximation of the overall reading of the camera depends on both the ambient temperature and the object temperature:
\begin{equation}\label{eq:finalWithFPA}
    L(\tobj, \tamb) = G(\tamb)\cdot L(\tobj)+D(\tamb)
\end{equation} where $\tamb,\tobj$ are the ambient and object temperatures, respectively.

$G(\tamb)$ and $D(\tamb)$ in \cref{eq:finalWithFPA} are polynomials of $\tamb$. The polynomial model has been previously shown to be representative of the underlying physical thermal radiation model (e.g,\cite{Nugent2013,Tempelhahn2016}). For the remainder of this work, higher-order polynomials, mainly quadratic, will be used for approximations.

Separating the coefficients from the object temperature in \cref{eq:finalWithFPA} is complicated when only the camera response is given \cite{Papini2018}.
However, some mathematical functions can separate a product into a summation, such as the $\log()$ function \cite{math_handbook}. The existence of a separation function suggests the use of neural networks, which can approximate any function \cite{nn_alg_book}. Thus, this work attempts to leverage a neural network for representing the physical model in \cref{eq:finalWithFPA}.
\section{Prior work}\label{sec:prior}
    \noindent Nonuniformity correction is an ongoing area of research. Different approaches are described in \cref{sec:prior:calib,sec:prior:singleimage}.

\subsection{Calibration-based methods}\label{sec:prior:calib}
\noindent The process of calibration requires collecting data of a known heat source under different environmental conditions. This process is usually conducted with a scientifically calibrated blackbody in an environmental chamber. The data are used to find coefficients that solve an equation for the calibration. 

The baseline for the calibration methods is a one-point correction. These methods usually assume a known and constant gain across ambient temperatures, and only solve for the offset (e.g.,~\cite{Schulz1995}). The natural extension is the two-point correction where no assumption is made for the gain (e.g.,~\cite{Riou2004}). These early methods solved for the coefficients using a simple linear regression model.

Contemporary methods formulate the nonuniformity correction (NUC) as an inverse problem. 
Nugent et al.~\cite{Nugent2013} solved it as a least-squares problem, with the offset and gain modeled as polynomials of the object's temperature. 
In ref.~\cite{Nugent2014}, they used the internal shutter of the camera to periodically update the results of the calibration. 
Liang et al.~\cite{Liang2017} based the solution on interpolation of a predefined offset table for each ambient temperature, and the offset values for this table were found using a two-point correction.
Chang and Li~\cite{Chang2019Intergration} solved for both the ambient temperature and integration time of the camera.

Calibration-based methods produce good results but relay on the collection of extensive data. The data must be accurate and contain both varying object temperatures and ambient temperatures, requiring the use of scientific-grade equipment. Moreover, these methods are valid only for the camera used to collect the data, meaning that the data-collection process must be performed for every camera to be calibrated. Any attempt to apply the calibration data on another camera will be noisy and have noticeable FPN because the coefficients of the calibration will not be suitable between cameras.

\subsection{Scene-based methods}
\noindent Scene-based methods exploit redundant data in and between frames, rendering calibration unnecessary. 
The redundant data can be movement between frames, camera jitter between images, or a constraint on the dataset itself.

Most of these methods assume that the change in ambient temperature is slow, thus the gain and offset changes slowly, and both can be regarded as constant between frames. 
This assumption holds true, but only for a limited time and ambient temperature span.

Averbuch et al.~\cite{Averbuch2007} used the motion between frames. Consecutive frames were registered to add data on each pixel, and an inverse problem was solved to find the offset. The solution was updated using a Kalman filter.
Papini et al.~\cite{Papini2018} used pairs of blurred and sharp images to approximate the gain and offset.
Saragadam et al.~\cite{Saragadam2021} used a neural network as prior information for solving an optimization problem. The input to the network was jittered frames of the same object. The physical constraint shown in \cref{eq:stefanBoltzmannTaylor} was imposed as part of the optimization problem.

These approaches offer good approximations for the temperatures but are expensive to calculate and require redundancy between frames.

\subsection{Single image-based methods}\label{sec:prior:singleimage}
\noindent The idea of this approach is to use only information that is already embedded in the frame itself.
Scribner et al.~\cite{nnOld} used a neural network to find offset and gain. The neural network acted as a locally adaptive filter on a small neighborhood.
Tendero and Gilles~\cite{admire} equalized the frame using the cumulative histogram of all of the columns in the frame, and then used the discrete cosine transform to denoise the results.

Recently, methods that utilize neural networks in general, and convolutional neural networks (CNN) in particular, have been suggested.
He et al.~\cite{He2018} suggested using a U-Net-type CNN trained end-to-end.
Jian et al.~\cite{Jian2018} filtered the frame with a bilateral filter to allow the network to concentrate only on the high-frequency information.
Chang et al.~\cite{Chang2019} shared multiscale information between layers of the network to improve the NUC results.
Guan et al.~\cite{snrwdnn} used the wavelet transform to decompose the frame into different scales and then used the scales as input to the network. The network output was the wavelets coefficients of the corrected frame.

Estimating scene temperature accurately using only a single frame at different ambient temperatures, without the need to calibrate for each camera, has yet to be achieved.

\newcommand{\yDistFromSep}{2em}
\newcommand{\xLengthOfSep}{24em}
\newcommand{\heightTextBox}{2em}
\newcommand{\widthTextBox}{5em}
\newcommand{\widthImages}{4em}
\newcommand{\distFromFig}{0.01em}
\newcommand{\fontAboveBelowFigs}{\normalsize}
\newcommand{\fontAboveArrows}{\footnotesize}
\newcommand{\captionGlNotes}{GL}
\newcommand{\captionTemperatureNotes}{T}
\newcommand{\captionAboveFrame}{Gray levels}
\newcommand{\captionBoxFrame}{Frame\\ IR camera}

\begin{figure}
    \begin{tikzpicture}[node distance=auto, auto, thick]
        \coordinate (separator_start) at (0,0);
        \coordinate (separator_end) at ($(separator_start) + (\xLengthOfSep, 0)$);
        \draw [dashed, thick] (separator_start) to (separator_end) ;

        \coordinate (separator_start_image_width) at ($(separator_start) + (\widthImages / 2, 0)$);
        \coordinate (separator_end_image_width) at ($(separator_end) - (\widthImages / 2, 0)$);

        \node[draw, rectangle, minimum height=\heightTextBox,  minimum width=\widthTextBox, align=center] (orig_image_prev) at ($(separator_start) + (\widthImages / 2, \yDistFromSep)$)  {\captionBoxFrame};
        \node[draw, rectangle, minimum height=\heightTextBox, minimum width=\widthTextBox,  align=center] (nuc_image_prev) at ($(separator_start_image_width)!0.33!(separator_end_image_width) + (0, \yDistFromSep)$)  {NUC};
        \node[draw, rectangle, minimum height=\heightTextBox,  minimum width=\widthTextBox, align=center] (calib_image_prev) at ($(separator_start_image_width)!0.66!(separator_end_image_width) + (0, \yDistFromSep)$) {Calibration};
        \node[draw, rectangle, minimum height=\heightTextBox,  minimum width=\widthTextBox, align=center] (est_image_prev) at ($(separator_end) + (-\widthImages / 2, \yDistFromSep)$) {Estimation};

        \node[above=\distFromFig of orig_image_prev, font=\fontAboveBelowFigs] {\captionAboveFrame};
        \draw[-Latex] (orig_image_prev.east)  |-  (nuc_image_prev.west);
        \node[above, rotate=0, font=\fontAboveArrows] at ($(orig_image_prev.east)!0.5!(nuc_image_prev.west)$)         {\captionGlNotes};
        \draw[-Latex] (nuc_image_prev.east)  |-  (calib_image_prev.west);
        \node[above, rotate=0, font=\fontAboveArrows] at ($(nuc_image_prev.east)!0.5!(calib_image_prev.west)$)         {\captionGlNotes};
        \node[above, rotate=0, font=\fontAboveArrows] at ($(calib_image_prev.east)!0.5!(est_image_prev.west)$)         {\captionTemperatureNotes};
        \node[above=\distFromFig of nuc_image_prev, align=center, text width=5em, font=\footnotesize] {ADMIRE~\cite{admire}\\ He et al.~\cite{He2018}};
        \draw[-Latex] (calib_image_prev.east) |-  (est_image_prev.west);
        \node[above=\distFromFig of calib_image_prev, align=center, text width=6em, font=\footnotesize] {Riou et al.~\cite{Riou2004}\\ Nugent et al.~\cite{Nugent2013}};
        \node[above=\distFromFig of est_image_prev, font=\fontAboveBelowFigs] {Temperature};

        \node[draw, rectangle, minimum height=\heightTextBox,  minimum width=\widthTextBox, align=center] (orig_image_ours) at ($(separator_start) - (-\widthImages / 2, \yDistFromSep)$)  {\captionBoxFrame};
        \node[draw, rectangle, minimum height=\heightTextBox,  minimum width=\widthTextBox, align=center] (est_ours) at ($(separator_end) - (\widthImages / 2, \yDistFromSep)$) {Estimation};
        \node[draw, rectangle, text width=\widthTextBox, minimum height=\heightTextBox, align=center, font=\large] (method_ours) at ($(orig_image_ours.east)!0.5!(est_ours.west) - (0, 0)$) {Proposed method};

        \node[below, rotate=0, font=\fontAboveArrows] at ($(orig_image_ours.east)!0.5!(method_ours.west)$){\captionGlNotes};
        \draw[-Latex] (orig_image_ours.east)  |-  (method_ours.west);
        \node[below, rotate=0, font=\fontAboveArrows] at ($(est_ours.east)!0.5!(method_ours.west)$){\captionTemperatureNotes};
        \draw[-Latex] (method_ours.east)  |-  (est_ours.west);
        \node[below=\distFromFig of est_ours, font=\fontAboveBelowFigs] {Temperature};
        \node[below=\distFromFig of orig_image_ours, font=\fontAboveBelowFigs] {\captionAboveFrame};

        \node[font=\large] () at  ($(separator_start)!0.5!(separator_end) + (0, 1.5*\widthImages)$) {Previous methods};
        \node[font=\large] () at  ($(separator_start)!0.5!(separator_end) - (0, 1.15*\widthImages)$) {Ours};
    \end{tikzpicture}
    \caption{The difference between the proposed method to previous methods.
        Previous methods required two steps: first nonuniformity correction (NUC) and second is calibration. This required a known heat source and extensive data collection for each individual camera.
        Our method performs both steps simultaneously, and does not require calibration for different cameras.
        GL stands for gray levels, and T stands for temperature.
        The citations above the blocks are the methods used for each step.}
    \label{fig:prior:cmpOtherMethodsScheme}
\end{figure}
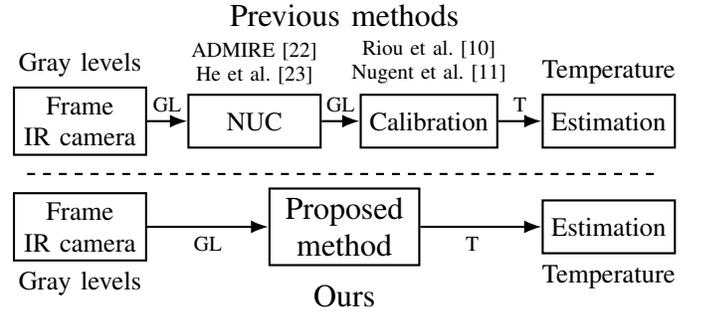
\cref{fig:prior:cmpOtherMethodsScheme} illustrates the difference between our method and previous methods for estimating the temperature of an object from a thermal image.
Previous methods consisted of two steps: first, applying a nonuniformity correction (NUC) to the thermal image to remove the effects of sensor fixed-pattern noise (FPN); second, calibrating the image using a known heat source to map the gray levels to temperature values.
This process required a specific calibration procedure for each individual camera, and a reliable heat source with a known temperature and emissivity.
Our method, on the other hand, performs both NUC and calibration in a single step.
Our method does not require any calibration for different cameras, and can estimate the temperature of an object from a single thermal image, without any prior knowledge of the object (e.g., the object's emissivity).

This work aims to both estimate the scene temperature and correct the nonuniformity in frames captured by low-cost uncooled microbolometer-based cameras.
We introduce a model for the nonuniformity in IR cameras based on the physical acquisition model, which accounts for the ambient temperature of the camera.
The model utilizes prior knowledge on the physics of the domain, namely radial spatial dependence, which is incorporated into the mathematical modeling of the nonuniformity. The nonuniformity model is general and represents different cameras, unlike previous calibration methods.
The model is used to train a neural network to correct the nonuniformity and produce accurate scene temperatures based only on a single frame and the ambient temperature of the camera.
We also compare a neural network with a physical constraint based on \cref{eq:finalWithFPA} to an end-to-end temperature-estimation network and show that the physical constraint improves performance.
Finally, we demonstrate our method on real data collected with a low-cost uncooled microbolometer camera and compare it to measurements taken with an uncooled scientific-grade accurate radiometric camera to show that the method indeed works and can provide generalizations.
To summarize the findings of this work:
\begin{enumerate}
    \item Estimating scene temperature accurately from a low-cost camera with only a single gray-level frame and ambient temperature as input.
    \item Elimination of the need to calibrate each camera separately by developing a nonuniformity simulator that uses physical prior knowledge of radial pixel dependence.
    The simulator is general and can faithfully represent multiple cameras and situations for a wide range of ambient temperatures.
    \item Investigation of the effect of the physical constraint introduced in \cref{eq:stefanBoltzmannTaylor} on the scene temperature estimation.
    \item Published a \href{https://drive.google.com/drive/folders/1tu_hMJR1SPunttWM65EyuCs7DJs2K6ah?usp=drive_link}{dataset} of thermal images with ground-truth temperature measurements.
\end{enumerate}
\section{Proposed method}\label{sec:method}
    \noindent The proposed method is aimed at estimating an accurate scene temperature from a single-image, while also correcting the space-variant degradation in microbolometer arrays.
%
\newcommand{\pathToFigs}{figures/schemeOfCharacterization}
\newcommand{\distanceOfRadiiFromSim}{14em}
\newcommand{\distanceBetweenRadiiSubfigs}{1.8em}
\newcommand{\distanceBetweenRadiiPack}{5.5em}
\newcommand{\distanceOfRadiiFromSum}{4em}
\newcommand{\xPixCoefFromSum}{4em}
\newcommand{\xFinalSumFromMult}{6em}
\newcommand{\yRealFromMulti}{3em}
\newcommand{\widthRadiiFigs}{1.5em}
\newcommand{\xMultFromCoef}{8em}
\newcommand{\widthRealFig}{4em}
\begin{figure*}
\begin{tikzpicture}[node distance=1cm, auto, thick]
\node[draw, rectangle, text width=2.1cm, minimum height=1cm, align=center] (sim) {Nonuniformity\\ simulator};
\newcommand{\heightInputsTextFromSim}{40pt}
\draw [dashed, thick] ($(sim.west) + (0, \heightInputsTextFromSim)$) -- ($(sim.east) + (0, \heightInputsTextFromSim)$) ;
\node[above =10pt+\heightInputsTextFromSim/2 of sim] (tfpa_caption) {$\tamb$};
\node[above =-2pt of tfpa_caption] (tobj_caption) {$\tobj$};
\node[above =-2pt of tobj_caption, align=center] (simulator_caption) {simulator:};
\node[above =-2pt of simulator_caption, align=center] {Inputs of the};

\node[inner sep=0pt] (a0) at ($(sim.west)+(\distanceOfRadiiFromSim,\distanceBetweenRadiiPack)$) {$\coefficientsAmbRadii{0}(\tamb)\times\mat{P}^0$};
\node[inner sep=0pt] (b0) at ($(sim.west)+(\distanceOfRadiiFromSim,-\distanceBetweenRadiiPack)$) {$\coefficientsAmbRadii{0}(\tamb)\times\mat{P}^0$};
\draw[-Latex] (sim.east) -- ++(0.2, 0) |-  (a0.west);
\draw[-Latex] (sim.east) -- ++(0.2, 0) |-  (b0.west);
\foreach \i in {1,...,2}
{
    \node[inner sep=0pt] (a\i) at ($(sim.west)+(\distanceOfRadiiFromSim,\distanceBetweenRadiiPack-\i*\distanceBetweenRadiiSubfigs)$) {$\coefficientsAmbRadii{\i}(\tamb)\times$\includegraphics[width=\widthRadiiFigs]{\pathToFigs/radius_order_\i.pdf}};
    \node[inner sep=0pt] (b\i) at ($(sim.west)+(\distanceOfRadiiFromSim,-\distanceBetweenRadiiPack+\i*\distanceBetweenRadiiSubfigs)$) {$\coefficientsAmbRadii{\i}(\tamb)\times$\includegraphics[width=\widthRadiiFigs]{\pathToFigs/radius_order_\i.pdf}};
    \draw[-Latex] (sim.east) -- ++(0.2, 0) |-  (a\i.west);
    \draw[-Latex] (sim.east) -- ++(0.2, 0) |-  (b\i.west);
}

\draw[red, dashed, thick, font=\small] ($(a0.west)-(2ex, -2em)$) rectangle ($(a2.east) + (1ex, -\widthRadiiFigs)$) node[midway, above=2.2em, font=\footnotesize] {$m=1$};
\draw[red, dashed, thick, font=\small] ($(b0.west)-(2ex, 2em)$) rectangle ($(b2.east) + (1ex, \widthRadiiFigs)$) node[midway, below=2.2em, font=\footnotesize] {$m=2$};

\node[draw, circle, inner sep=2pt] (sum_upper) at ($(a2.east)+(\distanceOfRadiiFromSum,0.2*\widthRealFig)$) {$+$};
\node[draw, circle, inner sep=2pt] (sum_lower) at ($(b2.east)+(\distanceOfRadiiFromSum,-0.2*\widthRealFig)$) {$+$};
\draw[-Latex] (a0.east) -- (sum_upper.north west);
\draw[-Latex] (a1.east) -- ([yshift=1mm]sum_upper.west);
\draw[-Latex] (a2.east) -- (sum_upper.west);
\draw[-Latex] (b2.east) -- (sum_lower.west);
\draw[-Latex] (b1.east) -- ([yshift=-1mm]sum_lower.west);
\draw[-Latex] (b0.east) -- (sum_lower.south west);

\node[inner sep=0pt] (coef_upper) at ($(sum_upper.east)+(\xPixCoefFromSum,0.)$) {\includegraphics[width=\widthRealFig]{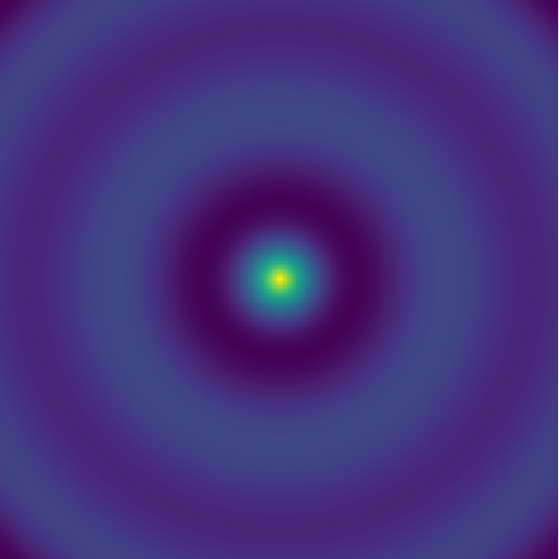}};
\node[above =-2pt of coef_upper, font=\small] {$\coefficientsPixelwise{m=1}(\tamb)$};
\draw[-Latex] (sum_upper.east) --  (coef_upper.west);
\node[inner sep=0pt] (coef_lower) at ($(sum_lower.east)+(\xPixCoefFromSum,0.)$) {\includegraphics[width=\widthRealFig]{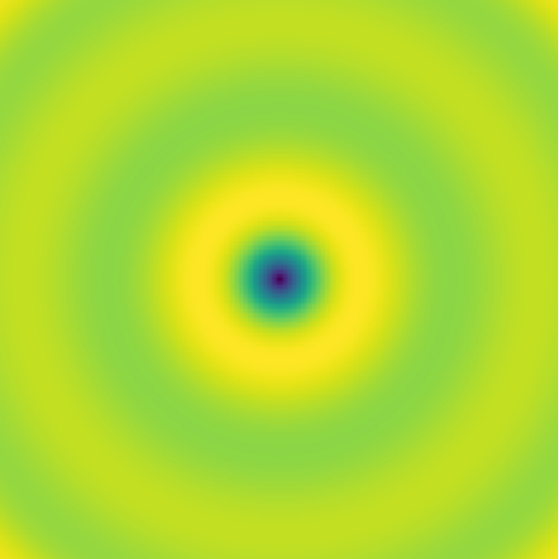}};
\node[below =-2pt of coef_lower, font=\small] {$\coefficientsPixelwise{m=2}(\tamb)$};
\draw[-Latex] (sum_lower.east) --  (coef_lower.west);

\node[draw, circle, inner sep=2pt] (mult_upper) at ($(coef_upper.east)+(\xMultFromCoef,-0.3)$) {$\times$};
\draw[-Latex] (coef_upper.east) -- (mult_upper.west);
\node[draw, circle, inner sep=2pt] (mult_lower) at ($(coef_lower.east)+(\xMultFromCoef,0.3)$) {$\times$};
\draw[-Latex] (coef_lower.east) -- (mult_lower.west);

\newcommand{\scaleOfPowerBrackets}{1.8}
\newcommand{\scaleOfPower}{1.3}
\node[inner sep=0pt] (real_upper) at ($(mult_upper.north)+(0,\yRealFromMulti)$) {\includegraphics[width=\widthRealFig]{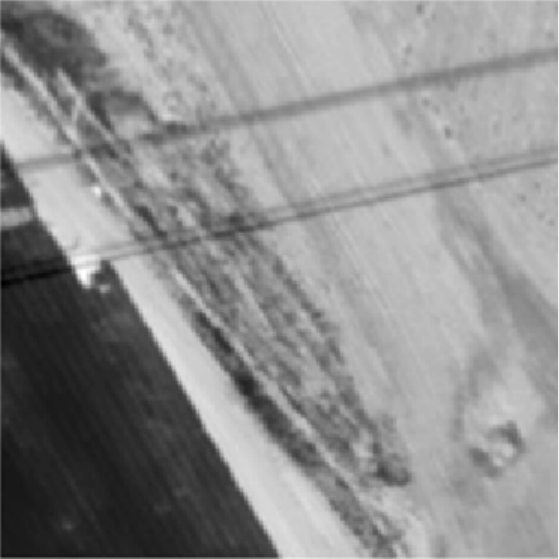}};
\node[inner sep=0pt, font=\Huge, scale=\scaleOfPowerBrackets] () at ($(real_upper.west)-(0.11,0)$) {$[$};
\node[inner sep=0pt, font=\Huge, scale=\scaleOfPowerBrackets] (real_upper_bracket_right) at ($(real_upper.east)+(0.1,0.0)$) {$]$};
\node[inner sep=0pt, font=\Huge, scale=\scaleOfPower] () at ($(real_upper_bracket_right.east)+(0.07,0.3)$) {$^1$};
\draw[-Latex] (real_upper.south) -- (mult_upper.north) ;
\node[inner sep=0pt] (real_lower) at ($(mult_lower.south)-(0,\yRealFromMulti)$) {\includegraphics[width=\widthRealFig]{\pathToFigs/real.pdf}};
\node[inner sep=0pt, font=\Huge, scale=\scaleOfPowerBrackets] () at ($(real_lower.west)-(0.11,0)$) {$[$};
\node[inner sep=0pt, font=\Huge, scale=\scaleOfPowerBrackets] (real_lower_bracket_right) at ($(real_lower.east)+(0.1,0.0)$) {$]$};
\node[inner sep=0pt, font=\Huge, scale=\scaleOfPower] () at ($(real_lower_bracket_right.east)+(0.07,0.3)$) {$^2$};
\draw[-Latex] (real_lower.north) -- (mult_lower.south) ;

\node[draw, circle, inner sep=2pt] (sum_final) at ($(mult_lower.north)!0.5!(mult_upper.south) + (\xFinalSumFromMult, 0)$) {$+$};
\draw[-Latex] (mult_upper.south east) -- (sum_final.west);
\draw[-Latex] (mult_lower.north east) -- (sum_final.west);
\node[inner sep=0pt] (sample) at ($(sum_final.east)+(2,0)$) {\includegraphics[width=\widthRealFig]{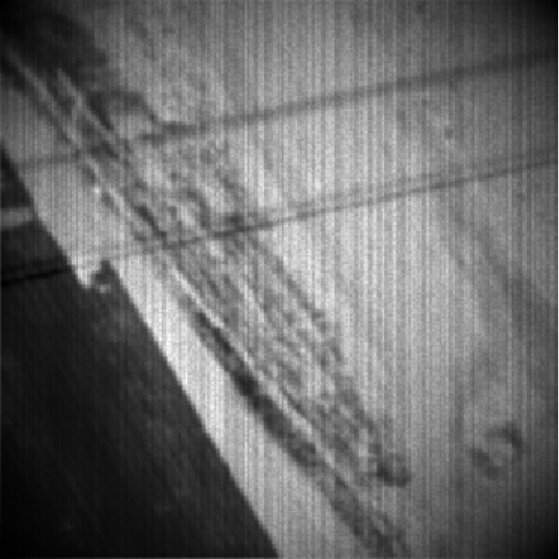}};
\node[below =0em of sample, align=center] {Synthetic\\ Gray levels};
\draw[-Latex] (sum_final.east) -- (sample.west);

\newcommand{\yDashedLine}{6.4cm}
\coordinate (top_left_border) at ($(a0.west)!0.65!(sim.east) + (0,3.1cm)$);
\draw [ dashed, thin] (top_left_border) -- ($(top_left_border) - (0, \yDashedLine)$) ;
\coordinate (top_radial_border) let \p1 = (top_left_border), \p2=(sum_lower.east), \p3=(coef_lower.west) in (\x2/2+\x3/2, \y1) coordinate (top_radial_border);
\draw [ dashed, thin] (top_radial_border) -- ($(top_radial_border) - (0,\yDashedLine)$) ;
\coordinate (top_real_border) let \p1 = (top_radial_border), \p2=(coef_lower.east), \p3=(real_upper.west) in (0.3*\x2+0.7*\x3, \y1) coordinate (top_real_border);
\draw [ dashed, thin] (top_real_border) -- ($(top_real_border) - (0,\yDashedLine)$) ;
\coordinate (top_right_border) let \p1 = (top_real_border), \p2=(sum_final.east), \p3=(sample.west) in (\x2/2+\x3/2, \y1) coordinate (top_right_border);
\draw [name =borderRight, dashed, thin] (top_right_border) -- ($(top_right_border) - (0,\yDashedLine)$) ;

\node[align=center, font=\small]  at ($(top_left_border)!0.5!(top_radial_border) - (0, 1em)$) {Radial coefficients\\ $\mRadial=3$};
\node[align=center, font=\small, text width=9em]  at ($(top_radial_border)!0.5!(top_real_border) - (0, 1em)$) {Pixel-wise coefficients\\ $\mGrayLevels=2$};
\node[align=center, font=\small]  at ($(top_real_border)!0.5!(top_right_border) - (0, 1em)$) {Powers of temperature map\\$\tobj$ (input)};

\node[draw, rectangle, text width=2.2cm, minimum height=0.4cm, align=center] (noise_block) at ($(sample.north)+(0,1)$) {Noise\\ FPN + additive};
\draw[-Latex] (noise_block.south) -- (sample.north);

\end{tikzpicture}
\caption{Illustration of the simulation process of a gray-level frame from an accurate temperature map.
The camera characterization method detailed in \cref{sec:method:characterize} and in \cref{sec:supp:characterization} of the supplementary material.
Specifically, the figure depicts \cref{eq:supp:charactarization:simulation} in the supplementary material.
During the training phase of the simulator, $\mRadial$ the number of radial coefficients and $\mGrayLevels$ the number of polynomial coefficients are set.
During the inference phase, the nonuniformity simulator takes as input $\tobj$ a temperature map measured by a scientific IR camera, and $\tamb$ the ambient temperature.
First, the simulator produces the radial coefficients $\coefficientsAmbRadii{m}^i(\tamb)$ which are multiplied by the corresponding power of $\mat{P}$ the meshgrid 2D radial map.
Each group of $\mRadial$ radial maps is summed pixel-wise to create the polynomial 2D coefficients $\coefficientsPixelwise{m}(\tamb)$.
Each 2D coefficient $\coefficientsPixelwise{m}(\tamb)$ is multiplied by the corresponding power of an accurate temperature map.
The $\mGrayLevels$ multiplications are summed pixel-wise to create the gray level image.
The image is then corrupted by additive noise and fixed-pattern noise (FPN) to mimic the sensor degradation and noises (\cref{sec:method:preprocessing}).}
\label{fig:method:simulator}
\end{figure*}

\noindent The proposed method is composed of four steps:
\begin{enumerate}[label=\Alph*.]
    \item Characterize the nonuniformity in an uncooled microbolometer thermal camera described by four steps (\cref{sec:method:characterize}).
          \begin{enumerate}[label=\arabic*)]
              \item Model the camera response to a set of object temperatures (\cref{sec:method:characterization:temperatureDependency}).
              \item Use the spatial dependency between pixels as a constraint (\cref{sec:method:characterization:spatialDependency}).
              \item Exploit symmetry around the middle of the frame (\cref{sec:method:characterization:axisSymmetry}).
              \item Apply the method to new frames (\cref{sec:method:characterization:applyToNew}).
          \end{enumerate}
    \item Acquire a large dataset of accurate temperature maps.
    \item Create samples using the accurate temperature maps and the synthetic nonuniformity  (\cref{alg:applyNonUniformity}).
    \item Train a neural network to perform NUC in a supervised manner (\cref{sec:method:net}).
\end{enumerate}

\subsection{Characterization of the nonuniformity}\label{sec:method:characterize}
    The goal of the characterization was to create a general model for the camera, that can be used to estimate the nonuniformity for any possible combination ambient temperature and object temperature.

    In this work, we used the low-cost uncooled microbolometer thermal camera FLIR \taucamera, because it allows access to raw measurements of thermal radiation.
    To estimate the nonuniformity for various ambient temperatures, the camera was placed in an environmental chamber (described in \cref{sec:materials:equipment}), and focused on a \blackbody blackbody, which served as the object of the setup.
    The \taucamera was set to $60$ frames per second (FPS). The camera output was set to radiation flux, so the raw measurement of each microbolometer is represented as a $14_{bit}$ integer. To acquire the rawest possible radiation flux and without any image processing, all the automatic image enhancements were disabled before each measurement (details are in \cref{tab:cameraParams} in the supplementary materiel).
    \cref{sec:materials:equipment} elaborates on the equipment used.

    An extensive dataset of camera responses was collected, comprised of the measured radiance in gray-levels at a known object temperature for different ambient temperatures.
    The radiance was measured for a series of \opPoints denoted as $R(\tamb, \tobj)_i$, where $\tamb$ is the ambient temperature, $\tobj$ is the object temperature and $i\in[1,\hdots,N]$. The measurements were made on a predefined set of temperatures such that $\tamb\in\Tamb$ and $\tobj\in\Tobj$.
    $N$ images were averaged for each \opPoint to lower the noise per pixel~$\left(\propto N^{-0.5}\right)$.

    We exploited domain knowledge about pixel-dependence in the nonuniformity. Specifically, the radial pixel dependence stemming from the self-heating described in \cref{sec:intro:nonuniformity}.
    
    To find the different coefficients for the simulator,
    first we found pixel-wise coefficients connecting between the radiance and the real temperature of the \blackbody,
    and then we found approximated pixel-wise coefficients with the radial pixel dependency constraint.
    The algorithm for finding the coefficients is described in \cref{alg:estimateNonUniformity} in the supplementary material.
    The algorithm also handles error in the measured data, such as skewing around the middle of the image.
    
    \cref{fig:method:simulator} illustrates the process of simulating a gray-level frame with nonuniformity from a temperature map.
    The process is also detailed in \cref{alg:applyNonUniformity} of the supplementary material.

    The sets of ground-truth temperatures and their corresponding degradation maps enable training different supervised algorithms for the nonuniformity correction. The maps are noise-less and perfectly symmetrical around the middle of the image. Different augmentations can be applied by the user to simulate noises of varying degrees, directional heating on the camera which results in skewing, and FPN on the frames. These augmentations increase generalization, making it possible to perform NUC on different cameras with the same training. Details on the exact augmentations performed in this work are given in \cref{sec:method:preprocessing}.

    The characterization is shown to be general in \cref{sec:experiments:realdata}.
    We used the characterization process to create synthetic nonuniformity maps and train a neural network (\cref{sec:method:net}), and used the network to estimate temperatures from scenes taken by a \emph{different camera} than the one used for the characterization. These experiments with real-world data are displayed and discussed in \cref{sec:experiments:realdata}.

    Detailed information on the characterization process and various challenges in it are given in \cref{sec:supp:characterization} of the supplementary material.
    The dataset used for characterization can be found \href{https://drive.google.com/drive/folders/1tu_hMJR1SPunttWM65EyuCs7DJs2K6ah?usp=drive_link}{Here}.

\subsection{Network}\label{sec:method:net}
    \noindent This work focuses on the physical modeling of the IR camera and the use of a physical constraint to improve the temperature estimation by using a single-frame.
To highlight the effect of the physical constraint, we use the well-known U-Net architecture~\cite{unet} because it is a well-established architecture for image-to-image translation tasks. The neural network itself is a backbone to test the physical constraint, and the results are not dependent on the network architecture.
More information on the network architecture is provided in \cref{sec:supp:network} of the supplementary material.
The architecture is presented in \cref{fig:supp:network} of the supplementary material.

The network shown in \cref{fig:supp:network} operates as an end-to-end function:
\begin{equation}\label{eq:netEnd2End}
    \tobjapprox = F(I(\tobj), \tamb)
\end{equation} where $I(\tamb)$ is a gray-level map taken at known ambient temperature $\tamb$, and $F$ is the output of the network blocks. We name this configuration \textbf{E2E} (end to end).

\cref{eq:stefanBoltzmannTaylor} shows that the radiance is a linear function of the scene temperature, and \cref{eq:finalWithFPA} shows a linear dependence on the ambient temperature.
To plug this prior knowledge into the network, the final block in the backbone network was replaced with two identical blocks.
Both blocks have the same input, which is the output of the layer before the split.
These blocks extract the estimated object temperature $\tobjapprox$ from the linear approximation of the radiance shown in \cref{eq:finalWithFPA}:
\begin{equation}
    \begin{split}\label{eq:netPysical}
        I(\tamb) &= G(\tamb)\cdot\tobj+D(\tamb) \longrightarrow \\
        \tobjapprox &= \mathcal{G} \cdot I(\tamb) + \mathcal{D}
    \end{split}
\end{equation}
where $I(\tamb)$ is the input to the network, and $\mathcal{G}\approx\frac{1}{G}, \mathcal{D}\approx\frac{D}{G}$ and the outputs of the respective blocks. We name this configuration \textbf{GxPD} (Gx + D).

The network is trained on the synthetic dataset created by the simulator described in \cref{sec:method:characterize}, and the inference is performed on a single IR gray-level frame.

The effects of both networks are elaborated in \cref{sec:experiments}.

\subsection{Loss functions}
\noindent The loss function is comprised of a fidelity term, a structural term, and a noise-reduction term. The fidelity term is the mean absolute error (MAE) which is robust to outliers \cite{anwar2020}, applied on the difference between the accurate temperature map $\tobj$ and the output of the network $\tobjapprox$ from \cref{eq:netPysical}:
\begin{equation}\label{eq:lossFid}
    \mathcal{L}_{Fid} = \frac{1}{h\cdot w}\sum_{i,j}\left|\tobj\left[i,j\right] - \tobjapprox\left[i,j\right]\right|
\end{equation}
where $h,w$ are height and width respectively.

The structural term measures the dissimilarity index, based on the structural similarity metric (SSIM). The SSIM is aimed at providing a good metric for the human visual-perception system. Use of the DSSIM method has been shown to improve network performance in image-restoration tasks \cite{ssimLoss2017}. It is calculated as:
\begin{equation}\label{eq:lossSSIM}
    \mathcal{L}_{DSSIM} = \frac{1-\text{SSIM}(\tobj,\tobjapprox)}{2}
\end{equation}

The noise-reduction term is total variation loss \cite{totalvariation}. The underlying assumption is that the sum of absolute gradients for noisy images is higher than for clean images:
\begin{subequations}
    \begin{align*}
        \mathcal{L}_{TV}(\tobjapprox) = \frac{1}{h\cdot w}\sum_{i,j}&\left|\tobjapprox\left[i,j+1\right]-\tobjapprox\left[i,j\right]\right|+\\+&\left|\tobjapprox\left[i+1,j\right]-\tobjapprox\left[i,j\right]\right|
    \end{align*}
\end{subequations}
where $i,j$ denotes the pixel position.

The overall loss term for the network training is:
\begin{equation}\label{eq:loss}
    \mathcal{L} = \mathcal{L}_{Fid} + \beta\cdot\mathcal{L}_{DSSIM} + \gamma\cdot\mathcal{L}_{TV}
\end{equation}
where $\beta, \gamma$ are the hyperparameters that balance the loss terms.

\subsection{Preprocessing}\label{sec:method:preprocessing}
\noindent The input to the network is a gray-level map created from an accurate temperature map using the synthetic nonuniformity as described in \cref{alg:applyNonUniformity}. The input to the network can be described as:
\begin{equation}\label{eq:netInput}
    I(\tamb) = \Hat{R}(\tamb, \tobj) + \mathcal{N}(0, \sigma^2)
\end{equation}
where $\Hat{R}(\tamb, \tobj)$ is the synthetic gray-level map (\cref{alg:applyNonUniformity}), and $\mathcal{N}$ is the additive Gaussian noise.

The input of the network is a frame representing the radiation flux measured by the microbolometer. These are represented by 14-bit gray-levels. To normalize them to the range of [0,1], the maximal and minimal values of gray-levels in the entire training and validation sets were obtained, and all inputs were normalized by:
\begin{equation}
    \Bar{I}(\tamb) = \frac{I(\tamb) - I_\text{min}}{I_\text{max}-I_\text{min}}
\end{equation}
where $\Bar{I}(\tamb)$ is the normalized input and $I_\text{min},I_\text{max}$ are the minimal and maximal gray-levels over the datasets.

The accurate temperature maps must also be normalized to the range [0,1]. Again, the maximal and minimal temperatures were found over all datasets and both the output of the network and the original accurate temperature maps were normalized:
\begin{equation}
    \Bar{T} = \frac{T - T_\text{min}}{T_\text{max}-T_\text{min}}
\end{equation}
where $\Bar{T}$ is the normalized accurate temperature map and $T_\text{min},T_\text{max}$ are the minimal and maximal temperatures over all datasets.

Augmentations were applied during training and validation to enrich the dataset further. These included cropping to $256\times256$ pixels, random horizontal and vertical flips, and $90^\circ$ rotations.

Random Gaussian noise with $\sigma^2=5_{GL}$ and FPN were generated for each frame. FPN was generated as:
\begin{equation}
    M_{FPN} = \begin{bmatrix}
        1 \\ \vdots  \\ 1
    \end{bmatrix}_{h\times1}\cdot
    \left(\begin{bmatrix}
        \mathcal{U}[v_{\min}, v_{\max}] \\ \vdots  \\ \mathcal{U}[v_{\min}, v_{\max}]
    \end{bmatrix}^T\right)_{1\times w}
\end{equation} where $\mathcal{U}$ is uniform distribution. $v_{\min}, v_{\max}$ were chosen as $v_{\min}=0.9, v_{\max}=1$.

During training, all augmentations were generated and applied randomly, i.e, random cropping and flipping, and randomly generated noise and FPN\@.
During validation, the cropping was a $256\times 256$ pixels rectangle around the center of the frame, to make the validation process deterministic.
Moreover, the Gaussian noise and FPN were generated once for each frame and used throughout the entire validation process. This was done to allow a fair comparison between experiments.

To construct the input of the network, first cropping and flipping augmentations were applied to a temperature map $T$. Second, a random $\tamb$ was generated and used with the augmented temperature map in \cref{eq:estimationOfMeas:fpa} to obtain a simulated camera response $\simulatedCameraResponse{\tamb}{T}$. Then, normalization was applied to this simulated response to get $\Bar{I}(\tamb)$. The last step was to apply the Gaussian noise $\left(\mathcal{N}(1, \sigma^2)\right)$ and FPN $\left(M_{FPN}\right)$ to the normalized simulated camera response:
\begin{equation}
    I^{in}_{\tamb} = \mathcal{N}(1, \sigma^2) \otimes M_{FPN} \otimes \Bar{I}(\tamb)
\end{equation} where $I^{in}_{\tamb}$ is the normalized gray-level input to the network and $\otimes$ is the element-wise multiplication.

\subsection{Training details}\label{sec:experiments:training}
\noindent The network was trained using the ADAM optimizer \cite{adamOpt2015} with a learning rate of $10^{-4}$. The learning rate was halved on a validation loss plateau of more than 3 epochs. The network was run for a 100 epochs, but early stopping was applied for a validation loss plateau of 8 epochs.
The weights were initialized using the orthogonal scheme \cite{orthoInit2014} with a scaling of $50^{-2}$.
The training was run on a single Nvidia 2080Ti. The network was written in Python3.8 using Pytorch 1.10.  The hyperparameters of the network are given in \cref{tab:hyperparams} at the supplementary material.
The optimal hyper-parameters by optimizing on the average MAE (\cref{eq:lossFid}) of the validation sets.

The convergence results for the validation MAE of the E2E and GxPD network is found in \cref{fig:results:convergence} at the supplementary materiel.
\section{Experimental results}\label{sec:experiments}
    \noindent The methods described in \cref{sec:method} were used to estimate temperature maps and correct nonuniformity in microbolometer-based thermal cameras. The presented experiments are organized as follows: 
\begin{enumerate}
    \item The data and equipment used to develop the proposed method.
    \item The results for characterization of the nonuniformity as presented in \cref{sec:method:characterize}.
    \item The results of the NUC performed by the network, including the effect of the physical constraint.
    \item The results of the NUC performed by the network on real data.
\end{enumerate}

\subsection{Data}\label{sec:materials:data}
\noindent The data for the environmental chamber were measured at ambient temperatures $\Tamb$= \{27, 31, 37.2, 38.9, 40.4, 41.5, 43.6, 44.7, 46.2, 46.8, 48, 50.8\}$^\circ C$. The \blackbody temperatures at each \opPoint were $\Tobj$= \{20, 25, 30, 35, 40, 45, 50, 55, 60\}$^\circ C$.

Noise variance was determined from the environmental chamber measurements:
\begin{equation}\label{eq:varNoise}
    \sigma^2[\tamb,\tobj] = \frac{1}{h\cdot w}\sum_{i=0}^h\sum_{j=0}^w\text{Var}(R[\tamb,\tobj][i,j])_N
\end{equation}
As seen in \cref{eq:varNoise}, the noise variance  used as input to train the network in \cref{eq:netInput} was the average over the spatial dimension of the variance map obtained from $N$ images. The effects of $\tamb,\tobj$ on $\sigma^2$ were found to be negligible, so the average of all $\sigma^2$ was $\sigma^2=5$ gray levels.

As for the training of the network, the datasets were temperature maps collected using a FLIR \scientificCamera camera, which is a scientific-level bolometer-based radiometric camera. The \scientificCamera accuracy is only $2\%$ of the temperature range in each frame.

The training dataset was $12,897$ frames. The validation set was comprised of $4,723$ frames. All frames were of different agricultural fields in Israel, taken from an unmanned aerial vehicle (UAV) flying $70_m-100_m$ above the ground.
Only sharp frames were used, hand-picked by a human user.

The validation sets were captured at the same locations as the training sets, but on different days. This validation procedure was chosen to eliminate data leakage between the training and validation sets, so that the metrics represent the ability of the network to generalize to different data. 
The training and validation dataset split remained the same for all training schemes, to allow a fair comparison between different experiments.

\subsection{Equipment}\label{sec:materials:equipment}
\begin{figure}
    \centering
    \includegraphics[width=0.8\linewidth]{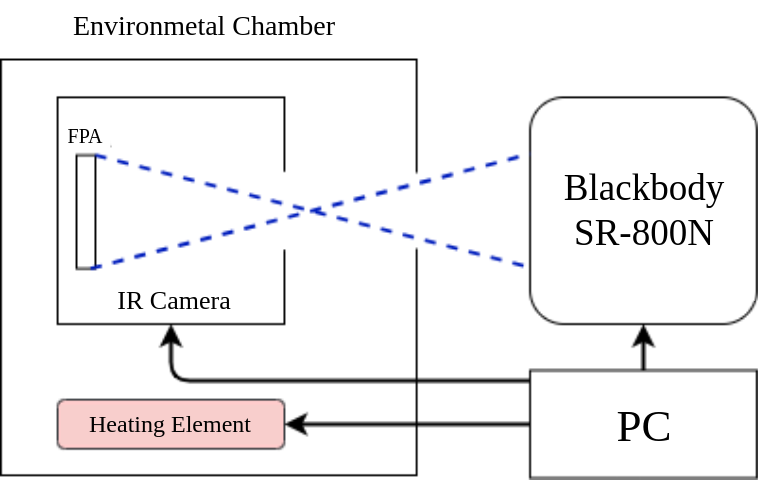}
    \caption{Schematic of the environmental chamber. The chamber is controlled via a PC and is comprised of a \taucamera camera, a heating element and a \blackbody blackbody. FPA is the focal plane array of the camera.}
    \label{fig:envChamber}
\end{figure}
\noindent The environmental chamber used for the characterization process (\cref{sec:method:characterize}) was designed and built at the Agricultural Research Organization, Volcani Institute. A cooking oven was adapted by controlling the heating element with a \campbellCtrl controller. A PID control loop was implemented on the \campbellCtrl to achieve a stable ambient temperature for the camera inside the oven. A schematic of the environmental chamber is presented in \cref{fig:envChamber}.

The \campbellCtrl, \taucamera and \blackbody were all controlled via Python3.8 from a Linux Ubuntu 20.04 computer.

The camera was calibrated using FLIR ThermalResearch v2.1. The configuration of the camera can be seen in \cref{tab:cameraParams} at the supplementary material and information on the various functions can be found in Tau2 Quark Software IDD.

\subsection{Camera characterization}
The number of coefficients for the radial fit (\cref{eq:spatialRadiiFit}) was set to $\mRadial=8$. The number of coefficients for the FPA fit of the radial coefficients (\cref{eq:fitRadiiFPA}) was set to $\mFPA=3$. These values were chosen empirically.

\begin{figure}
    \centering
    \subfloat[]{\includegraphics[width=0.49\linewidth]{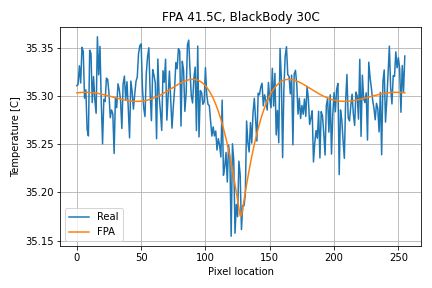}}\label{fig:fitRes:1}
    \hfill
    \subfloat[]{\includegraphics[width=0.49\linewidth]{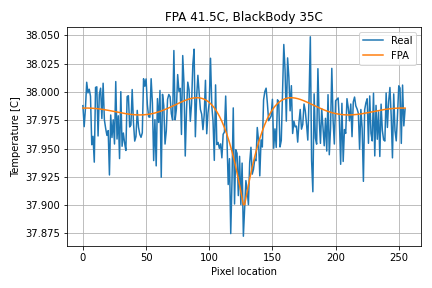}}\label{fig:fitRes:2}\\
    \subfloat[]{\includegraphics[width=0.49\linewidth]{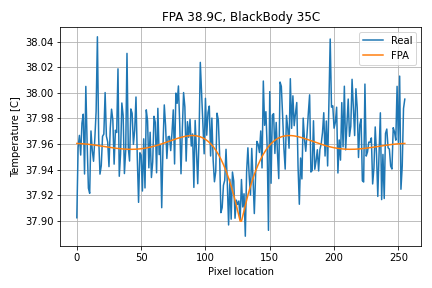}}\label{fig:fitRes:3}
    \hfill
    \subfloat[]{\includegraphics[width=0.49\linewidth]{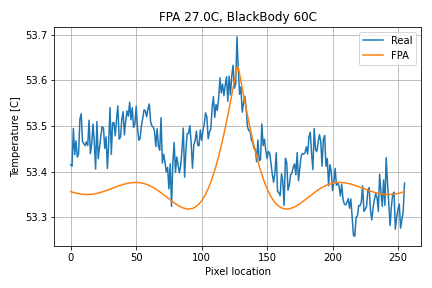}}\label{fig:fitRes:4}
    \caption{Side view of the fitting described in \cref{alg:applyNonUniformity}. Panel (d) demonstrates how the skewing is corrected.}
    \label{fig:fitRes}
\end{figure}

The results of the nonuniformity characterization process described in \cref{sec:method:characterize} as summarized by \cref{alg:estimateNonUniformity} are shown in \cref{fig:fitRes}. 
Four examples from different \opPoints are shown. These results illustrate that the fitting is both valid and corrects the skew in the measurements.

\subsection{Nonuniformity correction}\label{sec:results:nuc}
\begin{figure*}
    \centering
    \figPathces{gilat}\\
    \figPathces{neveyaar}\\
    \setcounter{subfigure}{0}
    \figPathcesCaption{ramon}
    \caption{Comparison of the results. A zoomed patch from the sample is presented (surrounded by a red square). From left to right - (a) the input sample, (b) the ground truth temperature map, (c) GxPD (ours), (d) ADMIRE~\cite{admire} (e) He et al.~\cite{He2018} and (f) SNRWDNN~\cite{snrwdnn}.}
     \label{fig:results:zoomin}
\end{figure*}
\begin{figure*}
    \centering
    \subfloat[Mevo Beitar]{\includegraphics[width=0.32\linewidth]{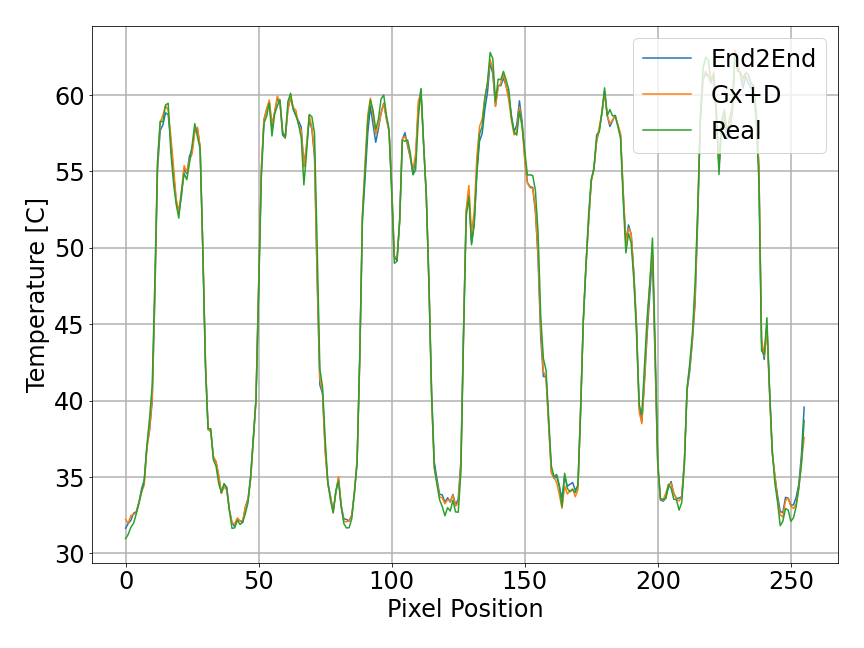}}
    \hfill
    \subfloat[Nir Eliyahu]{\includegraphics[width=0.32\linewidth]{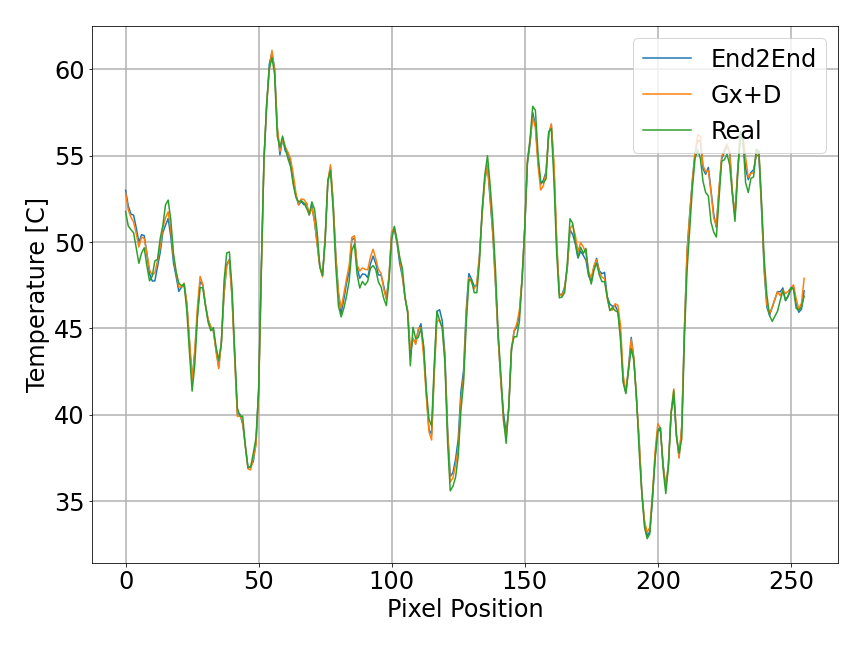}}
    \hfill
    \subfloat[Tzora]{\includegraphics[width=0.32\linewidth]{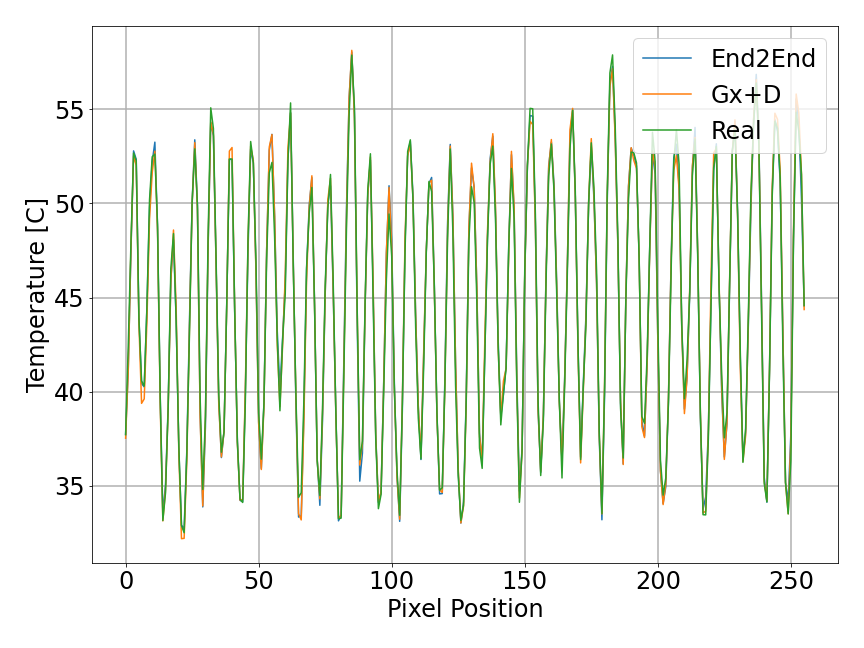}}
    \caption{Side view comparison of the network temperature estimation using the end-to-end (E2E) configuration and the physically constrained (GxPD) configurations.}
    \label{fig:results:plots}
\end{figure*}
\newcommand{\heightFigCmpMethodsReal}{25ex}
\newcommand{\colorRealMethodCmpCaptions}{white}
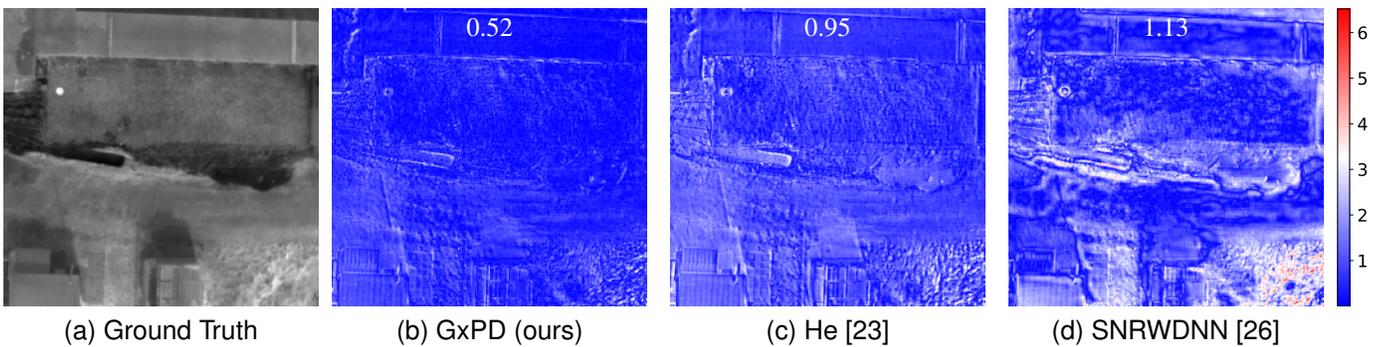
\begin{figure*}
    \subfloat[Ground Truth]{\includegraphics[height=\heightFigCmpMethodsReal]{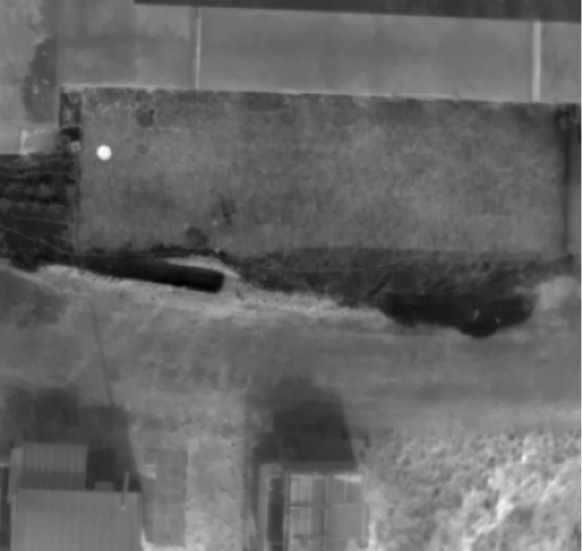}\label{fig:results:realDataCmp:gt}}
    \hfill
    \subfloat[GxPD (ours)]{
    \begin{tikzpicture}
        \node[anchor=south west, inner sep=0] at (0,0) {\includegraphics[height=\heightFigCmpMethodsReal]{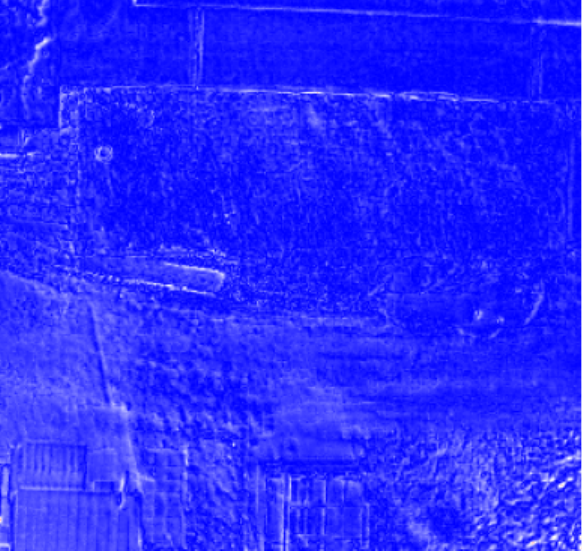}};
        \node[anchor=north, align=center, \colorRealMethodCmpCaptions] at (current bounding box.north) {0.52};
    \end{tikzpicture}\label{fig:results:realDataCmp:gxpd}
    }
    \hfill
    \subfloat[He~\cite{He2018}]{
    \begin{tikzpicture}
        \node[anchor=south west, inner sep=0] at (0,0) {\includegraphics[height=\heightFigCmpMethodsReal]{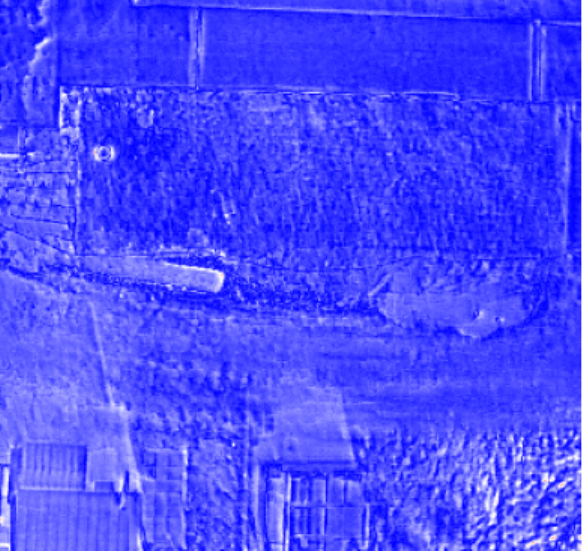}};
        \node[anchor=north , align=center, \colorRealMethodCmpCaptions] at (current bounding box.north) {0.95};
    \end{tikzpicture}\label{fig:results:realDataCmp:he}
    }
    \hfill
    \subfloat[SNRWDNN~\cite{snrwdnn}]{
        \begin{tikzpicture}
            \node[anchor=south west, inner sep=0] at (0,0) {\includegraphics[height=\heightFigCmpMethodsReal]{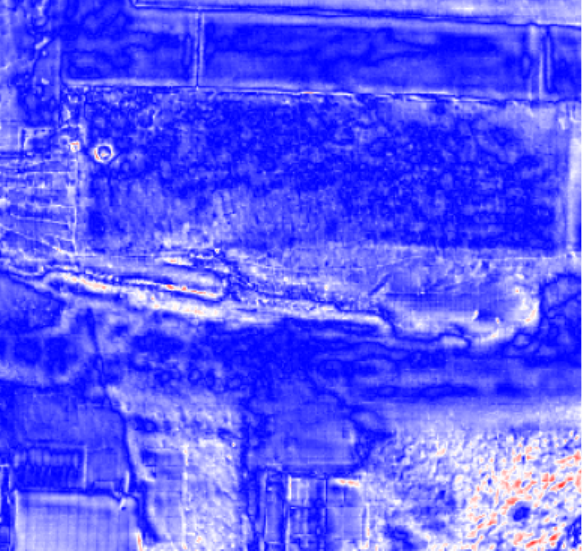}};
            \node[anchor=north , align=center, \colorRealMethodCmpCaptions] at (current bounding box.north) {1.13};
        \end{tikzpicture}\label{fig:results:realDataCmp:snrwdnn}
    }
    \hfill
    \subfloat{\includegraphics[height=\heightFigCmpMethodsReal]{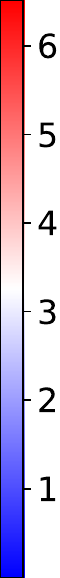}}
    \caption{Comparison between GxPD to other methods on real data.
    The ground truth temperature map is taken with the \scientificCamera, and the colored map is the difference between the results of the method with the temperature map.
    The number in white as the MAE in $\circ C$ between the temperature map and the results of the method.}
    \label{fig:results:realDataCmp}
\end{figure*}

A visual comparison of NUC between the proposed method and other methods is presented in \cref{fig:results:zoomin}. The left-most figure is the input to the network. The patch in the red square is zoomed-in and presented for GxPD, ADMIRE~\cite{admire}, He et al.~\cite{He2018} and SNRWDNN~\cite{snrwdnn}. 

Observing the results, both He et al.~\cite{He2018} and SNRWDNN~\cite{snrwdnn} does not thoroughly removes the NUC, and ADMIRE~\cite{admire} increases noise and adds surplus edges and details thus limiting the fidelity of its estimation. GxPD appears similar to the ground truth data.
More visual results are in \cref{fig:supp:patch:1,fig:supp:patch:2,fig:supp:patch:3,fig:supp:patch:4,fig:supp:patch:5,fig:supp:patch:6,fig:supp:patch:7,fig:supp:patch:8,fig:supp:patch:9,fig:supp:patch:10,fig:supp:patch:11,fig:supp:patch:12} in the \emph{supplementary material}.

A side-view of the results of the temperature estimation can be seen in \cref{fig:results:plots}. These figures contain the real temperature, and the estimations made by the results of the E2E network and the physically constrained network GxPD\@. As can be seen, both estimations are accurate and both network configurations are similar. The input to the network cannot be displayed with the plots, because it is in gray levels, whereas the network output temperatures are in $^\circ C$.

\cref{tab:resultsWithWithoutAmb} compares the metrics of the estimations between the different configurations and compares them to He et al~\cite{He2018} and to SNRWDNN~\cite{snrwdnn}. The latter results were retrained on the same data using the training scheme suggested by those authors. For a fair comparison, we constrained our network to the same depth and number of filters as He et al~\cite{He2018}.
The results of the E2E network without the ambient temperature are also compared. The metrics in the table are an average of the metrics from all validation sets.
Although ADMIRE~\cite{admire} is compared visually in \cref{fig:results:zoomin}, its metrics cannot be compared in the table because the method does not estimate the temperature, only corrects nonuniformity.

The physical constraint on the network (GxPD) improves the results by $12\%$ in MAE compared to the E2E network. This improvement is significant, and shows that the physical constraint is beneficial for the network.
GxPD achieved a $12\%$ improvements in MAE over E2E.
Although GxPD is superior, E2E still offers a significant improvement over other SOTA methods.
This improvement can be explained by the expressive power of the neural network. The network in E2E can intrinsically represent the GxPD network~\cite{nn_alg_book}.
Still, the $12\%$ decrease in MAE between E2E and GxPD means that the physical constraint still has a measurable effect on the results.

Another result in \cref{tab:resultsWithWithoutAmb} is that incorporating the ambient temperature into the network significantly improves the performance of the network, reducing the MAE by $13\%$.

\begin{table}[ht]
    \centering
    \caption{Estimation results for the different configurations (end-to-end (E2E), with and without (w/o) $\tamb$; and physically constrained (GxPD)).}
    \begin{tabular}{|c|c|c|c|}
    \hline
    Network & MAE [$^\circ C$] & PSNR [dB] & SSIM \\
    \hline
    He et al.~\cite{He2018} & 0.93 & 37.21 & 0.95 \\
    \hline
    SNRWDNN~\cite{snrwdnn} & 0.77 & 37.68 & 0.97 \\
    \hline 
    E2E w/o $\tamb$ &  0.48 & 43.25 & 0.99 \\
    \hline
    E2E with $\tamb$ &  0.42 & 44.50 & 0.99 \\
    \hline
    GxPD & 0.37 & 45.43 & 0.99 \\
    \hline
    \end{tabular}
    \label{tab:resultsWithWithoutAmb}
\end{table}

\subsection{Real data}\label{sec:experiments:realdata}
\begin{figure*}
    \centering
    \subfloat[Front of a car ($45_m$)]{\includegraphics[width=0.4\linewidth]{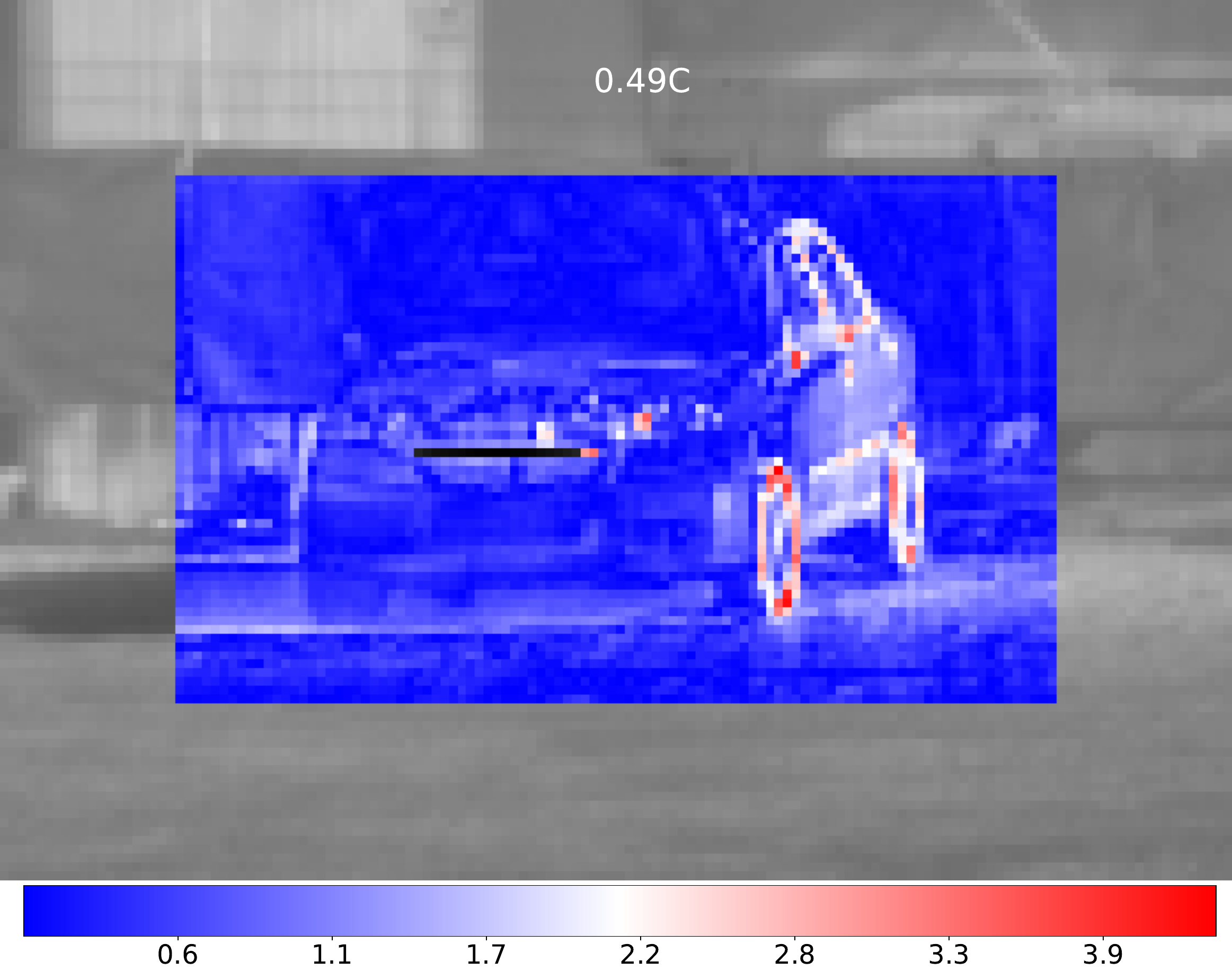}}
    \hfill
    \subfloat[Side view of a car ($45_m$)]{\includegraphics[width=0.48\linewidth]{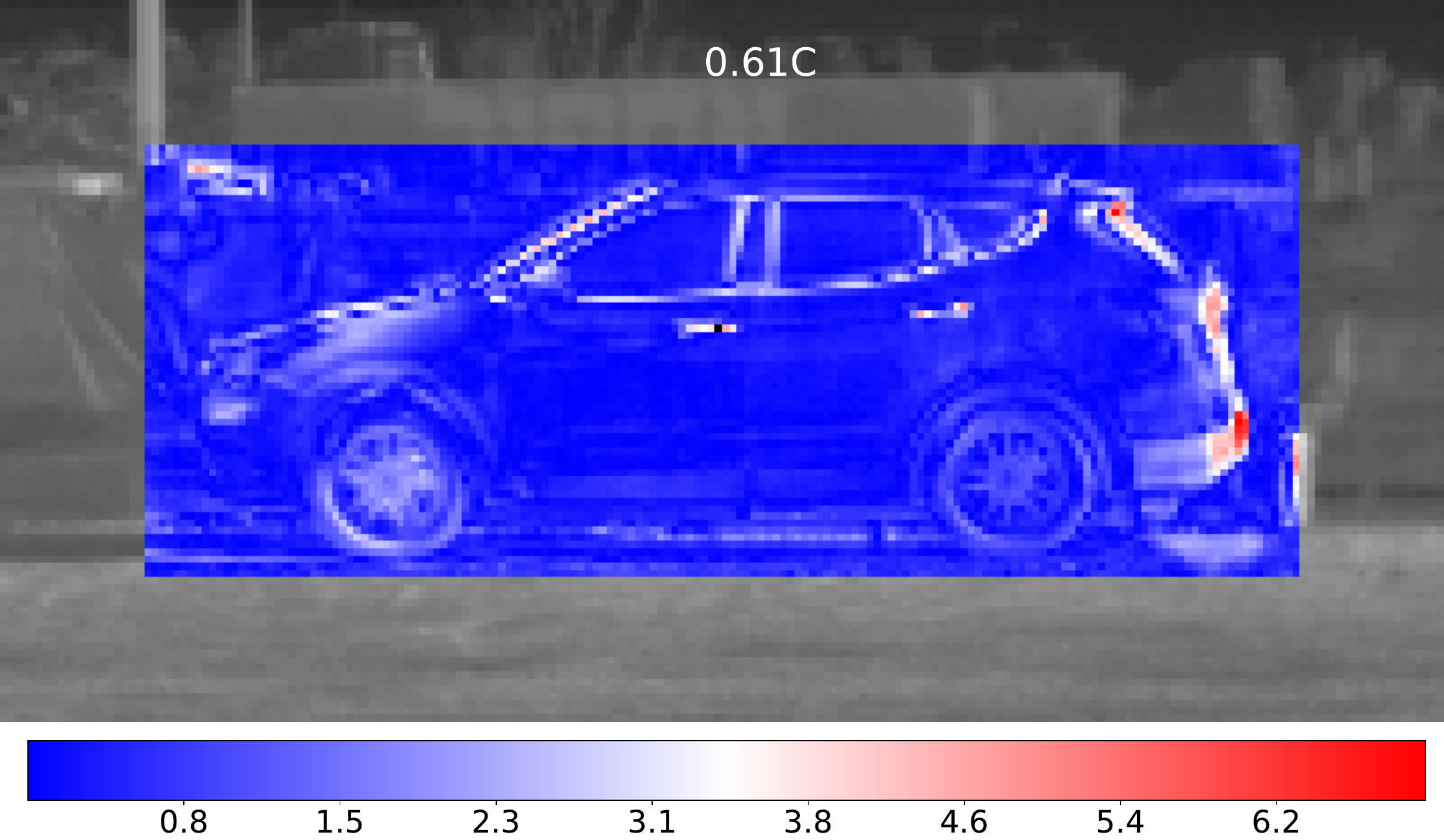}}\\
    \subfloat[Concrete shed with tin doors ($70_m$)]{\includegraphics[width=0.48\linewidth]{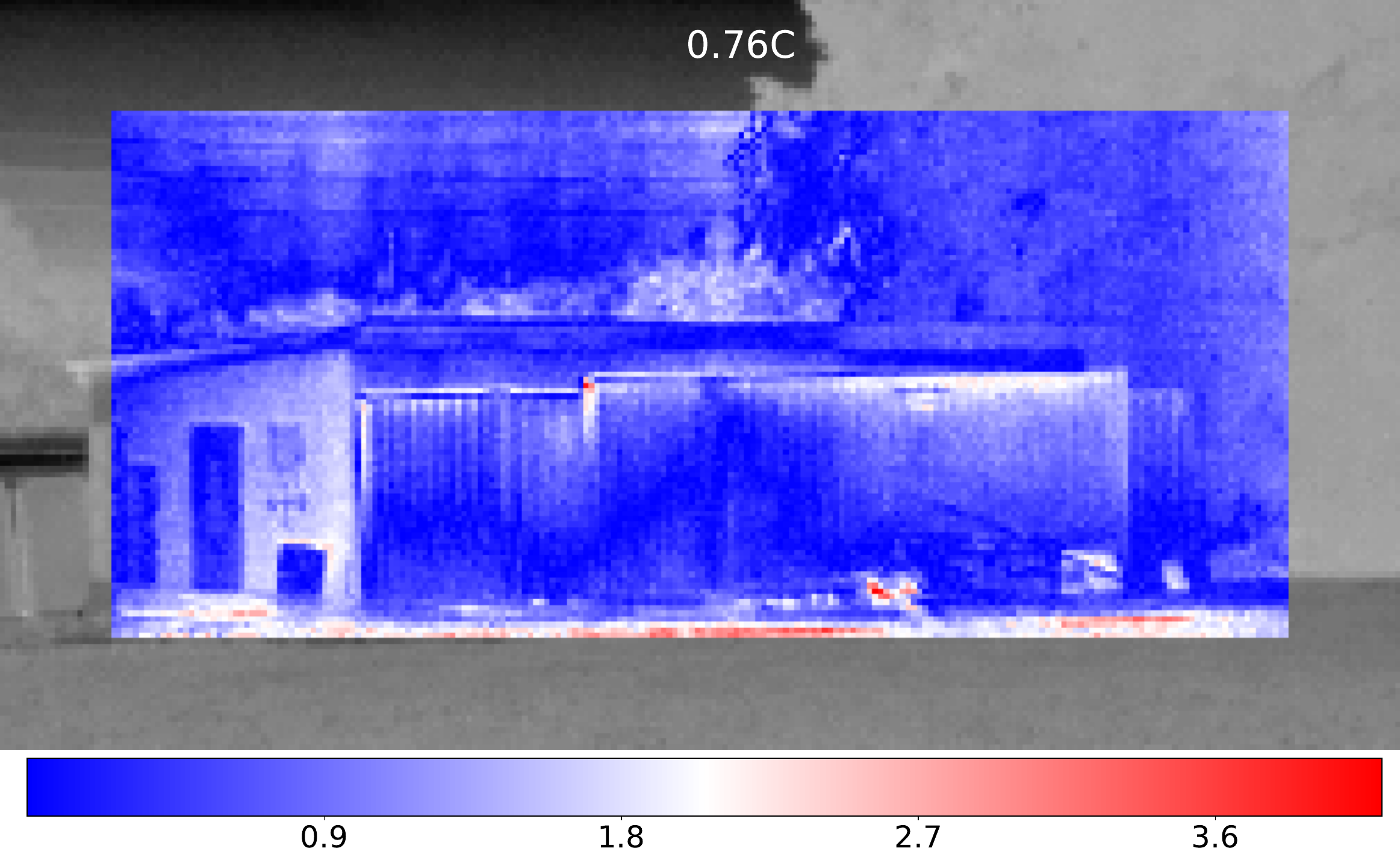}}
    \hfill
    \subfloat[Buildings ($500_m$)]{\includegraphics[width=0.48\linewidth]{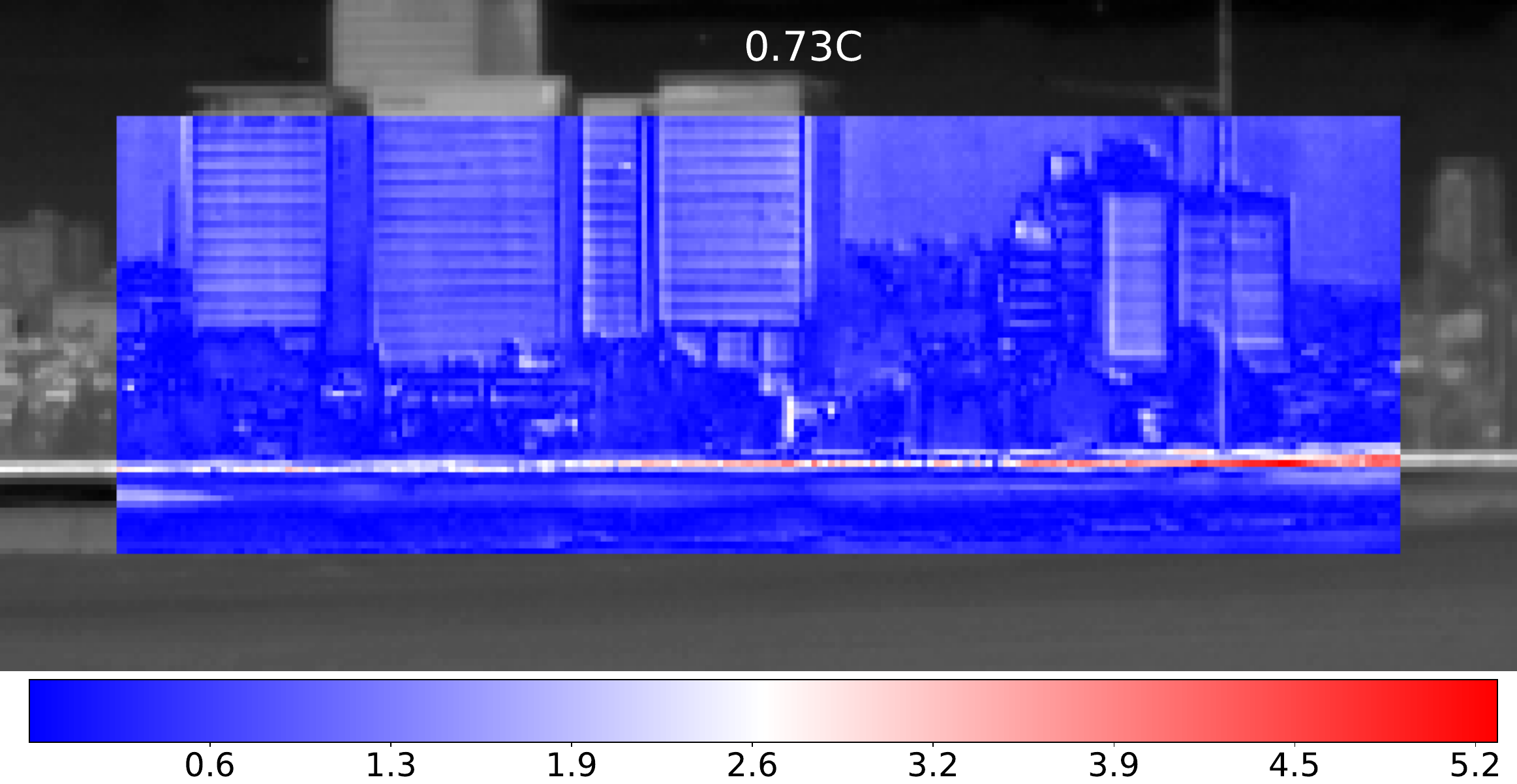}}\\
    \subfloat[Trees ($50_m$)]{\includegraphics[width=0.4\linewidth]{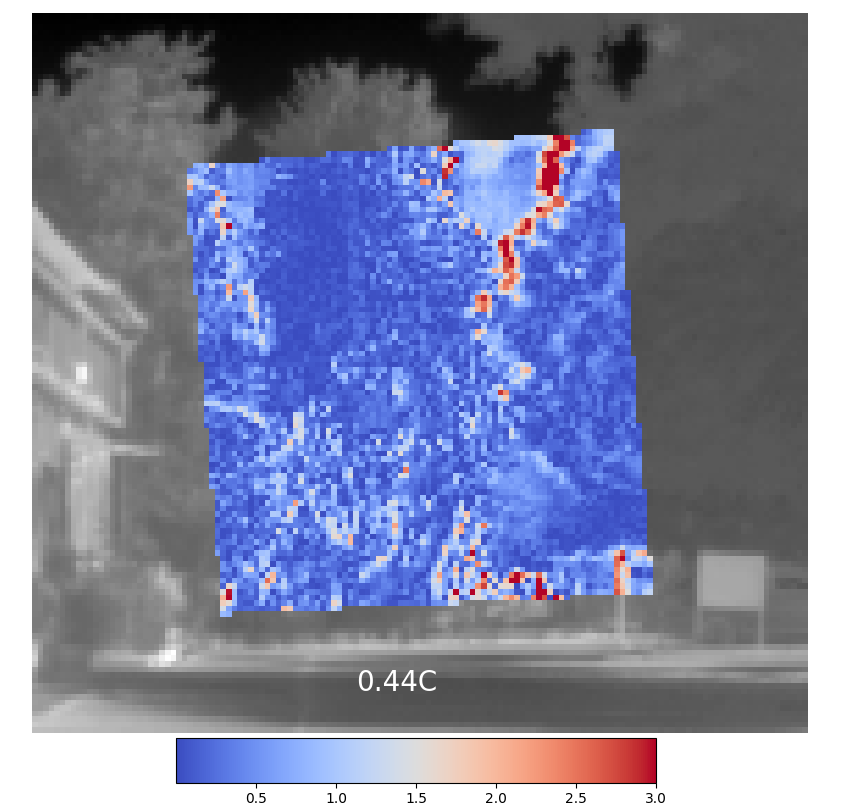}}
    \hfill
    \subfloat[Buildings ($500_m$)]{\includegraphics[width=0.6\linewidth]{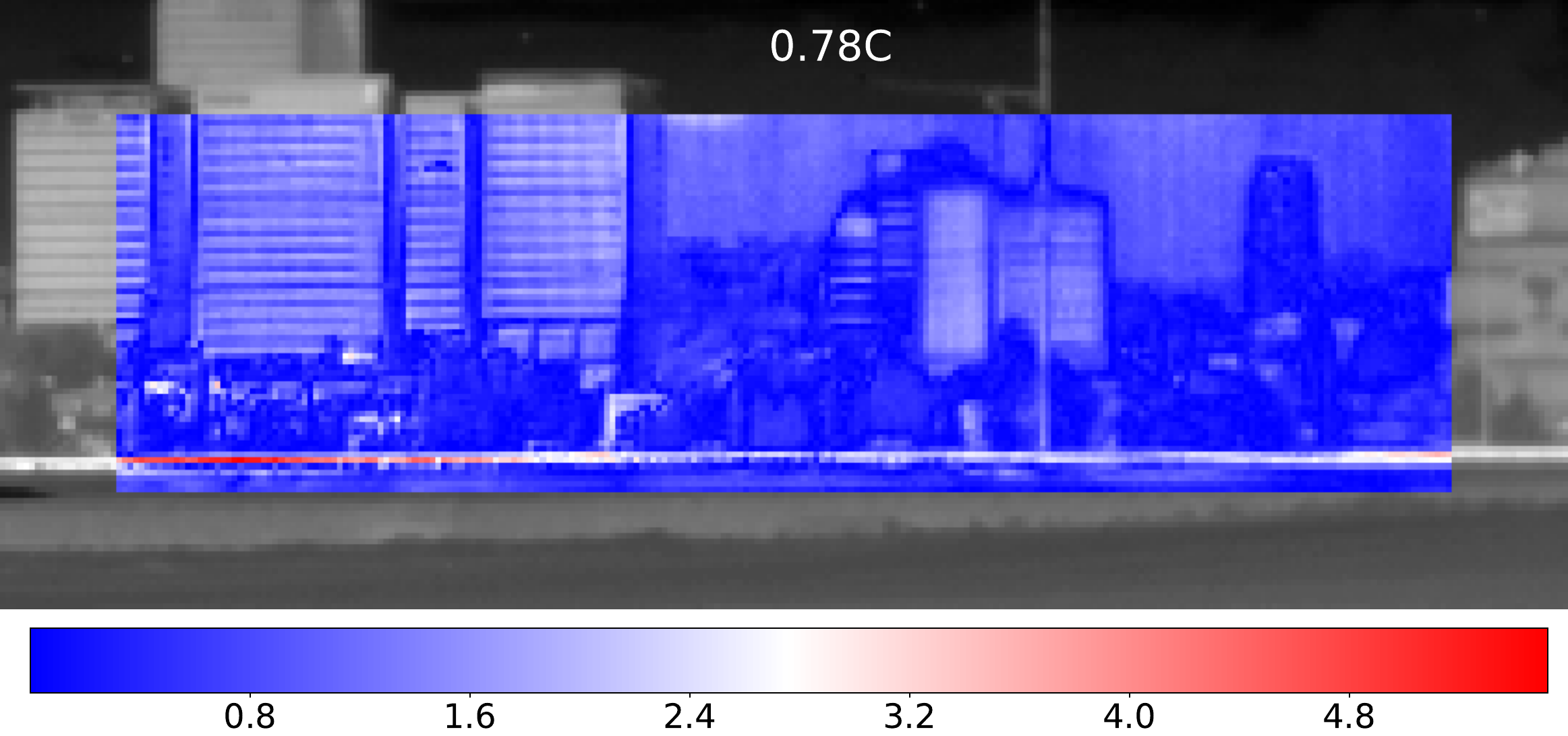}}
    \caption{Various results on real data. The temperature map taken with the \scientificCamera serves as the gray background, and the colored map is the difference between the results of GxPD with the temperature map. The number in white as the MAE in $\circ C$ between the temperature map and the results of GxPD.}
    \label{fig:results:realData}
\end{figure*}%
\noindent We captured the same scene with an accurate \scientificCamera scientific-level radiometric camera and with the \taucamera camera. The \scientificCamera outputs a temperature map and the \taucamera outputs gray levels corresponding to the radiation flux. The camera used for capturing these images was different from the one used for the calibration process.

The ambient temperature and emissivity of the \scientificCamera were tuned using an accurate temperature sensor placed in the scene. The scenes were registered by hand-picking correspondence points and performing a homography with OpenCV V4.5.4.

Six results are presented in \cref{fig:results:realData} and six more are presented in \cref{jeep,shed,building,shed2,building_sunlight,warehouse} in the supplementary material. The gray scales are the temperatures taken using the \scientificCamera. The blue patches in the frames are the per-pixel differences between the temperatures and the results of GxPD\@.
The numbers in white are the MAE between GxPD and the temperature map. We used the GxPD method because its MAE results were significantly better.
The two uppermost figures are cars taken at the morning. The hot areas with high errors stem from direct sunlight hitting the metal and glass surfaces of the cars\. the next two figures are buildings captured from a great distance. The last figure is a tree from a distance. 
Part of the error stems from registration errors between the two cameras, or from moving objects during acquisition (e.g, leaves in the lowest figure).

The range of the MAE is $0.15^\circ C-0.93^\circ C$.
This small error in temperature estimation is of the same order as the accuracy of the scientific \scientificCamera. 
This accurate result was achieved without any thermographic corrections or NUC from the \taucamera, only the radiation flux as gray levels. The exact configuration can be seen in \cref{tab:cameraParams} at the supplementary materiel.
These results are also on-par with the results on the validation set (\cref{tab:resultsWithWithoutAmb}) and with the visual results (\cref{fig:results:zoomin}).

\cref{fig:results:realDataCmp} compared between GxPD to SNRWDNN~\cite{snrwdnn} and to He et al.~\cite{He2018}.
Two registered frames were captured from a UAV simultaneously: a ground truth temperature map using \scientificCamera, and a gray-level frame using the \taucamera camera. The \taucamera used for capturing these images was \emph{different} from the one used in the calibration process.
Each subfigure in \cref{fig:results:realDataCmp} is the error between the estimation of a method to the ground truth temperature map in \cref{fig:results:realDataCmp:gt}. The number in white is the MAE in $\circ C$ between the temperature map and the estimation of the method.
The estimations from left to right - \cref{fig:results:realDataCmp:gxpd} GxPD (ours), \cref{fig:results:realDataCmp:he} He et al.~\cite{He2018} and \cref{fig:results:realDataCmp:snrwdnn} SNRWDNN~\cite{snrwdnn}.
The figure illustrates the generalization capabilities of our method.
While the other methods are trained on the same dataset as our method, they suffer significant degradation in performance when ran on frames acquired by other cameras.
Specifically, the temperature map estimated by SNRWDNN~\cite{snrwdnn} in \cref{fig:results:realDataCmp:snrwdnn} is full of artifacts, and suffer a significant increase in error as compared to the results on the synthetic dataset in \cref{fig:results:zoomin,tab:resultsWithWithoutAmb}.

%
\section{Conclusion}\label{sec:conclusion}
    \noindent A method to characterize the physical behavior of a system was demonstrated (\cref{sec:method:characterize}).
The characterization process allowed for supervised training of a neural network (\cref{sec:method:net}).
The temperature estimation performed by the network can be generalized to real data and different cameras (\cref{sec:experiments:realdata}).
We also showed that previously-suggested methods does not generalize well to different cameras (\cref{fig:results:realDataCmp}).
This allows for a faster NUC process that only requires a single collection of calibration data.
Moreover, the method not only estimates the temperature of the object, but also corrects the nonuniformity of the camera.

When using the proposed camera characterization and training, the backbone network (E2E) shows a significant improvement of roughly $1^\circ C$ compared to previous works~\cite{He2018}, producing a mean temperature error (MAE) of only $0.42^\circ C$ over the validation dataset.

We proposed a physical constraint on neural networks that incorporates the physical behavior of the system into the network (\cref{sec:method:net}).
The physically-constrained network (GxPD) achieved a MAE of $0.37^\circ C$, a significant improvement of $12\%$ in the accuracy of the temperature estimation compared to the backbone network (E2E).
Another notable result is that the ambient temperature of the camera has a significant effect on the accuracy of the temperature estimation, reducing the MAE by $13\%$.

Results on real-world experimental data achieve a MAE ranging in $0.15^\circ C-0.93^\circ C$ with a \emph{different camera} than the one used for training and validation. This shows that the proposed method can generalize to different cameras.
\section*{Acknowledgments}
\noindent The authors thank Dr. Yaffit Cohen and Dr. Eitan Goldstein for the UAV data used in this work; 
and Moti Barak, Lavi Rosenfeld and Liad Reshef for the design and construction of the environmental chamber.

\section*{Disclosures}
\noindent The authors declare no conflicts of interest.
\balance    
\FloatBarrier
\bibliographystyle{IEEEtran}
\bibliography{biblography}

\clearpage
\setcounter{figure}{0}  
\setcounter{table}{0}   
\setcounter{equation}{0}   
\setcounter{section}{0}   
\onecolumn
\makeatletter
\renewcommand{\thefigure}{S\@arabic\c@figure}
\renewcommand{\theequation}{S\@arabic\c@equation}
\renewcommand{\thesection}{S\@arabic\c@section}
\makeatother
\begin{center}
    {\Large \textbf{Supplementary material for:\\\titleOfArticle}}\\
    \href{mailto:navotoz@mail.tau.ac.il}{Navot Oz},
    \href{mailto:sochen@math.tau.ac.il}{Nir Sochen},
    \href{mailto:mend@eng.tau.ac.il}{David Mendelovich},
    and \href{mailto:iftach@volcani.agri.gov.il}{Iftach Klapp}
    
\end{center}

\section{Characterization of the nonuniformity}\label{sec:supp:characterization}
    \noindent The dataset used for characterization can be found \href{https://drive.google.com/drive/folders/1tu_hMJR1SPunttWM65EyuCs7DJs2K6ah?usp=drive_link}{Here}.
    \subsection{Object temperature dependence}\label{sec:method:characterization:temperatureDependency}
\noindent The camera response for a given object temperature and ambient temperature can be estimated as a polynomial for each pixel:
\begin{equation}\label{eq:supp:charactarization:simulation}
    \simulatedCameraResponse{\tamb}{\tobj}[x,y] = \sum_{m=0}^{\mGrayLevels}{\coefficientsPixelwise{m}(\tamb)[x,y]} \cdot \tobj^m[x,y]
\end{equation} where $\tobj$ is the temperature of the object, $\tamb$ is the ambient temperature of the camera, $\simulatedCameraResponse{\tamb}{\tobj}[x,y]$ is the estimated camera response, $\mGrayLevels$ is the degree of the polynomial fit, and $\coefficientsPixelwise{m}(\tamb)$ is the pixel-wise coefficient of the polynomial that depends on the ambient temperature.

Below, we outline the methods to estimate the coefficients of the camera response from measured data. In \cref{sec:method:characterization:applyToNew}, we show how these coefficients enable the synthesis of new images.

To estimate the coefficient vector $\coefficientsPixelwiseVec(\tamb)$, first the dependence of the response on ambient temperature was determined and fitted.
The dependence was determined from the \opPoint measurements by estimating the polynomial coefficients for each $\tamb$. A matrix of object temperatures at each \opPoint and a vector of camera responses is built for each $\tamb$, and the polynomial coefficients are found \emph{per pixel} using Least Squares. We denote:
\begin{align*}
    \mat{A}_O[\tamb]               & =
    \begin{bmatrix}
        \Tobj^0[1]           & \dots  & \Tobj^{\mGrayLevels}[1]           \\
        \vdots               & \ddots & \vdots                      \\
        \Tobj^0[\NumTempObj] & \dots  & \Tobj^{\mGrayLevels}[\NumTempObj]
    \end{bmatrix}_{\NumTempObj\times \mGrayLevels} \\
    \coefficientsRadiance{}[\tamb] & =
    \begin{bmatrix}
        c_0[\tamb] \\ \vdots \\ c_{\mGrayLevels}[\tamb]
    \end{bmatrix}_{\mGrayLevels\times1}                    \\
    \vec{R}[\tamb]                 & =
    \begin{bmatrix}
        R\left(\tamb,\Tobj[0]\right) \\\vdots\\R\left(\tamb,\Tobj[\NumTempObj]\right)
    \end{bmatrix}_{\NumTempObj\times1}
\end{align*} where $\vec{R}[\tamb]$ are the frames acquired by the low-cost IR camera, $\mGrayLevels$ is the degree of the polynomial to fit, and $\NumTempObj$ is the length of $\Tobj$.
Then the values of $\coefficientsRadiance{}[\tamb][x,y]$ are estimated by solving the inverse problem:
\begin{subequations}
    \begin{align}
        \vec{R}[\tamb][x,y]                 & = \mat{A}_O[\tamb] \cdot \coefficientsRadiance{}[\tamb][x,y] \label{eq:LsSolutionFPA} \rightarrow \\ 
        \coefficientsRadiance{}[\tamb][x,y] & = \mat{A}_O^{+}[\tamb]\cdot\vec{R}[\tamb][x,y] 
    \end{align}
\end{subequations}
where $\mat{A}_O^{+}$ is the Moore–Penrose inverse of $\mat{A}_O$.

A set of coefficients $\coefficientsRadiance{}\nolinebreak\in\nolinebreak\mathcal{R}^{\mGrayLevels}$ exists for each $\tamb\nolinebreak\in\nolinebreak\Tamb$. These coefficients are \textit{pixel-wise}, meaning there are $\mGrayLevels$ coefficient maps with spatial dimensions of $h\nolinebreak\times\nolinebreak w$ for $h,w$ the dimensions of each image. These coefficient maps were filtered using a Gaussian filter with $\sigma=1$ to remove high-frequency noise stemming from dead pixels in the camera.

\begin{figure*}[b]
    \centering
    \includegraphics[width=0.7\linewidth]{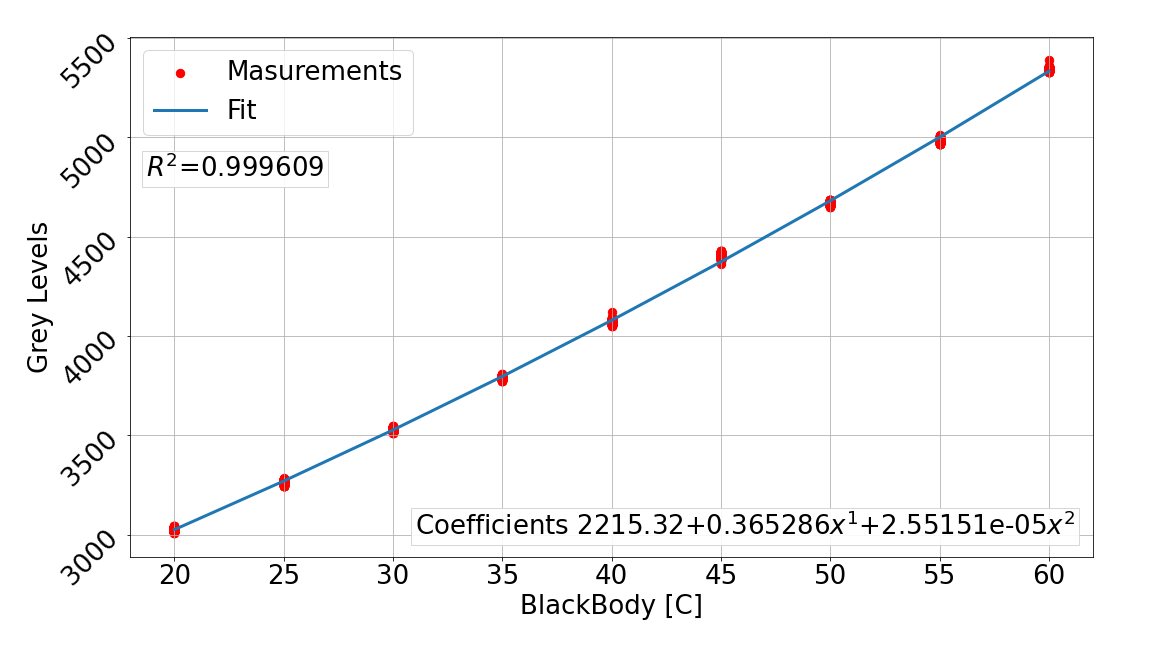}
    \caption{Example of quadratic fitting between gray level output of the \taucamera camera and the temperature of the blackbody as described in \cref{eq:LsSolutionFPA}.
        The measurements are taken at $\tamb=38.9^\circ C$ and the coefficients for the pixel in the middle of the frame are displayed.
        Specifically, the estimated gray level of the camera in the middle pixel is $\simulatedCameraResponse{38.9}{\tobj} \approx 2215.32+0.36\cdot \tobj+2.55\cdot\tobj^2$.}
    \label{fig:fitGl2BB}
\end{figure*}

\cref{fig:fitGl2BB} presents an example of the fitting results between the gray levels at an \opPoint an the real blackbody temperatures, as described in \cref{eq:LsSolutionFPA}. The number of coefficients was chosen empirically as $\mGrayLevels=3$. The fit provides a good estimation to the data ($R^2\geq0.99$).

\cref{fig:coefObjTemp} is a scheme of the coefficients for a given object temperature as described in \cref{eq:LsSolutionFPA}. Each coefficient map is two-dimensional.
\newcommand{\pathToFigsRadFromCoef}{supp_figures/coef/}
\newcommand{\widthFigsRadFromCoef}{2cm}
\newcommand{\fontSizeRadFromCoef}{\huge}
\newcommand{\fontCaptionBelowRadFromCoef}{\Large}
\begin{figure*}
    \centering
    \begin{tikzpicture}[node distance=centered, auto, thick]
        \node[inner sep=0pt] (beta) {\includegraphics[width=\widthFigsRadFromCoef]{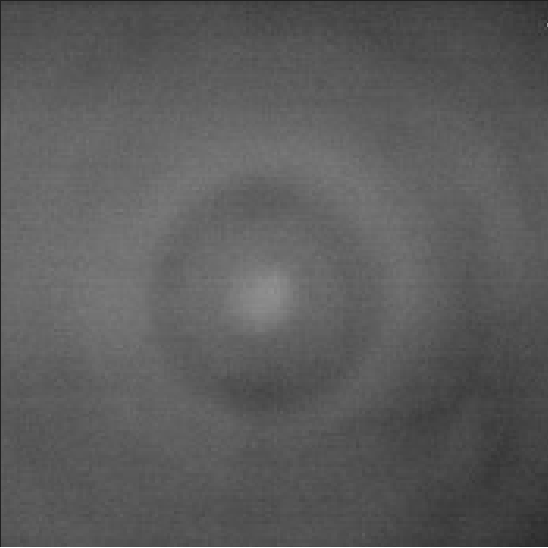}};
        \node[below =0.0ex of beta, font=\fontCaptionBelowRadFromCoef] {$\simulatedCameraResponse{\tamb}{\tobj}$};
        \node[inner sep=0pt, font=\fontSizeRadFromCoef] (approx) at ($(beta.east)+(1em,0)$) {$=$};
        \node[inner sep=0pt, font=\fontSizeRadFromCoef] (p0) at ($(approx.east)+(2em,0)$) {$\tobj^0$};
        \node[inner sep=0pt, font=\fontSizeRadFromCoef] (p0Fig) at ($(p0.east)+(3.5em,0)$) {\includegraphics[width=\widthFigsRadFromCoef]{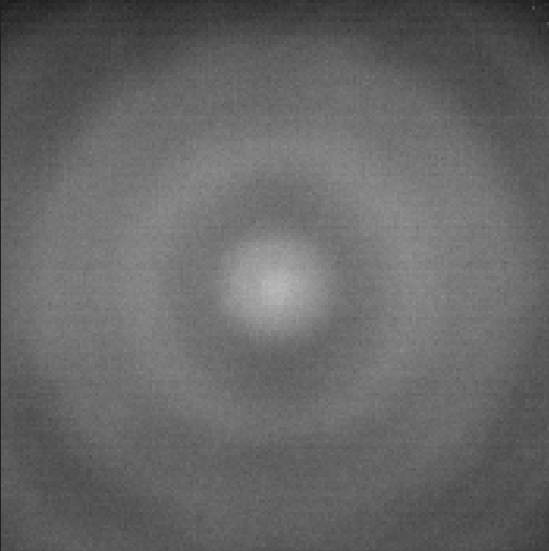}};
        \node[below =0.cm of p0Fig, font=\fontCaptionBelowRadFromCoef] {$\coefficientsPixelwise{0}$};
        \node[inner sep=0pt, font=\fontSizeRadFromCoef] (p1) at ($(p0Fig.east)+(2.5em,0)$) {$+\tobj^1$};
        \node[inner sep=0pt, font=\fontSizeRadFromCoef] (p1Fig) at ($(p1.east)+(3.5em,0)$) {\includegraphics[width=\widthFigsRadFromCoef]{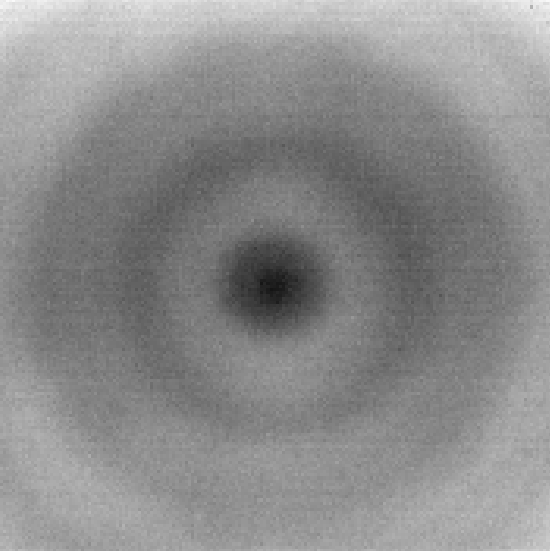}};
        \node[below =0.cm of p1Fig, font=\fontCaptionBelowRadFromCoef] {$\coefficientsPixelwise{1}$};
        \node[inner sep=0pt, font=\fontSizeRadFromCoef] (p2) at ($(p1Fig.east)+(2.5em,0)$) {$+\tobj^2$};
        \node[inner sep=0pt, font=\fontSizeRadFromCoef] (p2Fig) at ($(p2.east)+(3.5em,0)$) {\includegraphics[width=\widthFigsRadFromCoef]{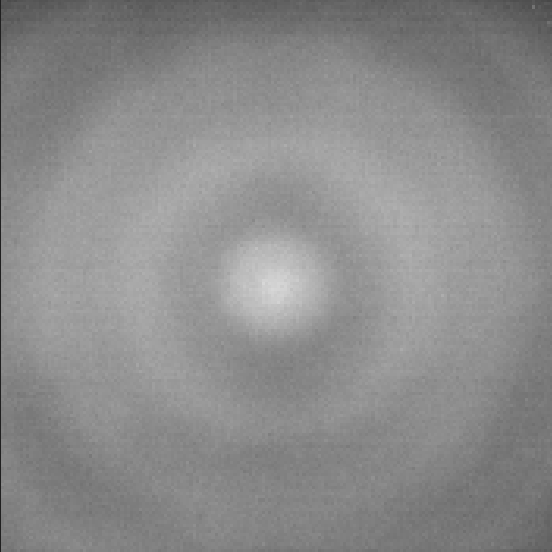}};
        \node[below =0.cm of p2Fig, font=\fontCaptionBelowRadFromCoef] {$\coefficientsPixelwise{2}$};
    \end{tikzpicture}
    \caption{Example of applying the nonuniformity simulator by using \cref{eq:supp:charactarization:simulation}.
        To estimate the radiance $\simulatedCameraResponse{\tamb}{\tobj}$ in gray levels, we find the pixel-wise sum of the coefficients $\coefficientsPixelwiseVec$ times the object temperature $\tobj$ with the appropriate power.
        The coefficients are unique for each ambient temperature.}
    \label{fig:coefObjTemp}
\end{figure*}

\begin{figure}
    \centering
    \subfloat[\fontfamily{ptm}\selectfont \small \centering (a) Real]{\includegraphics[width=0.48\linewidth]{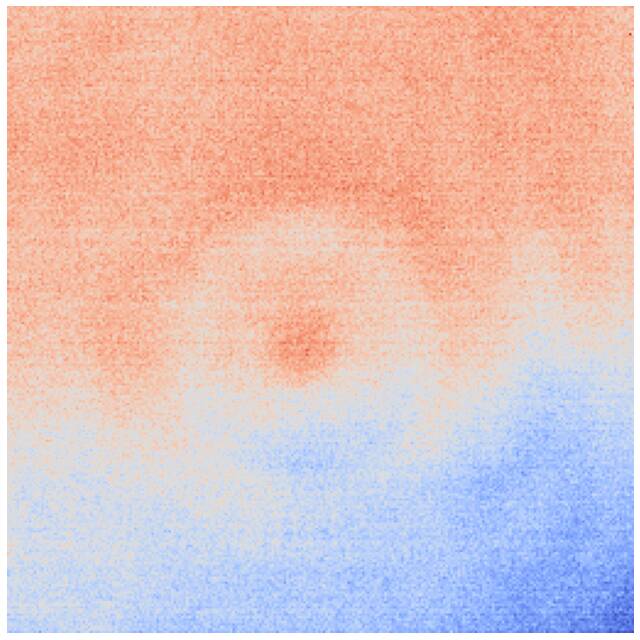}}
    \hfill
    \subfloat[\fontfamily{ptm}\selectfont \small \centering (b) Fit]{\includegraphics[width=0.48\linewidth]{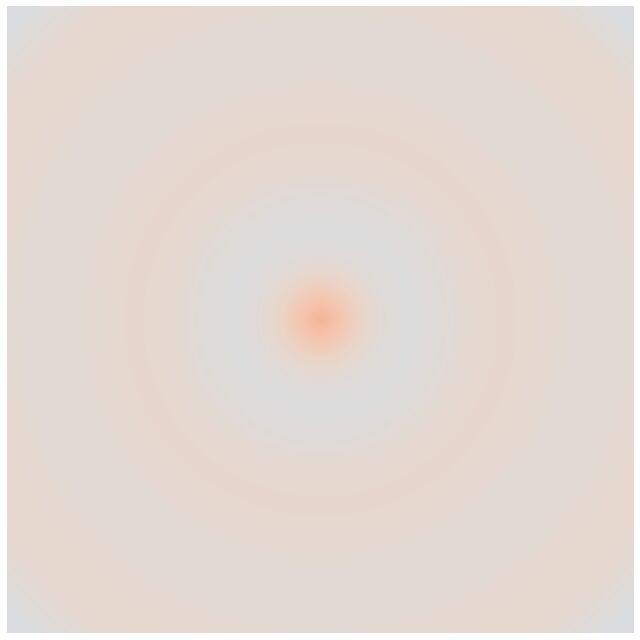}}
    \\
    \subfloat[\fontfamily{ptm}\selectfont \small \centering (c) Side view of the skewing effect in the measurements, the second-order quadratic fit, the fine fit and the subtraction between the coefficients, as detailed in \cref{eq:spatialFitSubtraction}]{\includegraphics[width=\linewidth]{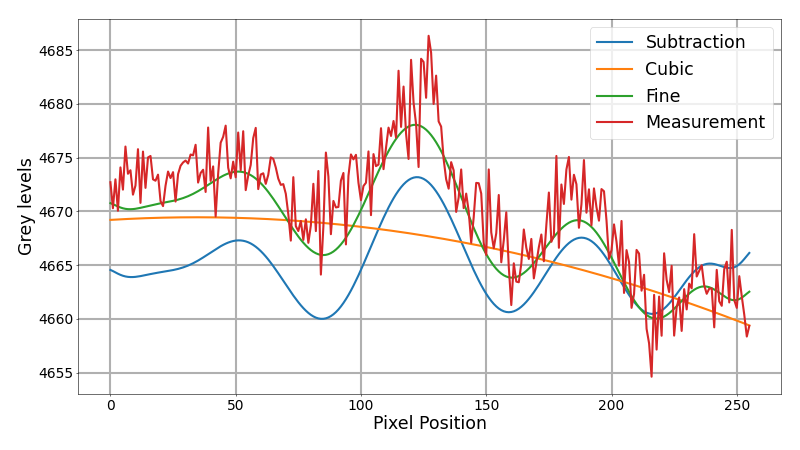}\label{fig:spatialFitSideView}}
    \caption{Example of skewing in the measurements, and the spatial fitting is according to \cref{eq:spatialFitSubtraction}. Nonuniformity in the real image is $1.4\%$.}
    \label{fig:skew}
\end{figure}

The measurements are expected to be symmetrical around the middle of the image~\cite{Tempelhahn2016}, but practical effects can create skew. The skewing limits the usability of the model because the skewed model does not accurately depict a general symmetrical case.
The effect of the skewing on real data can be seen in \cref{fig:skew}.

\subsection{Spatial dependence}\label{sec:method:characterization:spatialDependency}
\noindent So far, the polynomial dependence of the camera's readings on $\tobj,\tamb$\ were found for each pixel. To overcome the skewing, the coefficients from \cref{eq:LsSolutionFPA} are fitted to a spatial function. The spatial fitting is performed separately on each set of coefficients~$\coefficientsRadiance{m},\forall m\in[0,\ldots,\mGrayLevels]$.
The spatial fitting is performed twice: for a quadratic polynomial and for a high-degree polynomial. The coefficients that have the most profound effect on the skewing are the quadratic coefficients. The ideal form of nonuniformity is expected to be axis-symmetric, and a quadratic function can be viewed as a low-frequency distortion of this symmetry. Thus, subtraction of the polynomials up to the quadratic coefficient removes the low-frequency distortion and alleviates the skew. An example of the skewing effect on real data and fitting results can be seen in \cref{fig:skew}.

Under these assumptions, we intend to find a skew-less axis-symmetric polynomial approximation of the measurements. This will be achieved by first fitting the results to a spatial function, and then fitting again to a radial function.

The first step, fitting the coefficients of the camera response to a spatial function, exploits the correlation between neighboring pixels. The spatial fitting reduces the number of coefficients considerably, from $\propto h\times s$ - the number of pixels - to $\propto\mSpatial$ - the number of coefficients in the spatial fit where $\mSpatial<<h\times w$.

To fit to a spatial function, we first define two matrices of dimensions $h\times w$. The matrices are built from vectors in the range $[-0.5,0.5]$, in $\mat{H}$ as columns and in $\mat{W}$ as rows:
$$\mat{W} = \begin{bmatrix}
        -0.5   & \ldots & 0.5    \\
        \vdots & \ddots & \vdots \\
        -0.5   & \ldots & 0.5
    \end{bmatrix}_{h,w},
    \quad \mat{H}=\mat{W}^T $$

The spatial fit is defined as:
\begin{equation} \label{eq:spatialFit}
    \coefficientsSpatialRaw[\tamb]  =\argmin_{\coefficientsSpatialRaw[\tamb]}\left(\coefficientsRadiance{m}[\tamb] - \sum_{q=0}^{\mSpatial}\sum_{z=0}^{\mSpatial}  \coefficientsSpatialRaw[\tamb][q,z]\cdot\mat{H}^{q}\cdot \mat{W}^{z}\right), \qquad\coefficientsSpatialRaw[\tamb]\in\mathcal{R}^{\mSpatial\times\mSpatial}
\end{equation} where $\mSpatial$ is the number of coefficients in the spatial fit. The powers $q,z$ are applied respectively to matrices $\mat{H},\mat{W}$ element-wise.

To reduce the skewing (\cref{fig:skew}), we define $\coefficientsSpatialSkewless$ a skew-less fit, which is the subtraction between the quadratic fit and the fine fit. We also define $\coefficientsSpatialQuadratic$ as the quadratic fit with $\mSpatial\nolinebreak=\nolinebreak2$, and $\coefficientsSpatialFine$ as the fine fit with $\mSpatial\nolinebreak>>\nolinebreak2$:
\begin{equation}\label{eq:spatialFitSubtraction}
    \coefficientsSpatialSkewless[\tamb] =
    \begin{cases}
        \text{Mean}(\coefficientsSpatialQuadratic[q,z],\coefficientsSpatialFine[q,z]), & q=z=0            \\
        \coefficientsSpatialFine[q,z]-\coefficientsSpatialQuadratic[q,z],              & \forall q,z\neq0
    \end{cases}, \qquad\coefficientsSpatialSkewless[\tamb]\in\mathcal{R}^{\mSpatial\times\mSpatial}
\end{equation}
The bias coefficient is averaged between the fits. Empirically, this is found to produce better results.

\cref{fig:spatialFitSideView} shows a horizontal cross-sectional view of \cref{fig:skew}, along with the results of the spatial fitting in \cref{eq:spatialFitSubtraction}. The cross-section of the measurements, fine fit, quadratic fit and subtraction fitting are presented. The subtraction fitting is calculated by subtracting the quadratic polynomial from the fine polynomial. The number of coefficients for the fine fit were set to $\mSpatial=15$. The final fit does indeed alleviate the skewing, while remaining faithful to the measurements.

\subsection{Axis-symmetric fitting}\label{sec:method:characterization:axisSymmetry}
\noindent To exploit the radial symmetry around the middle of the image, the spatial fit results of \cref{eq:spatialFitSubtraction} are fitted to a radial kernel. For each ambient temperature $\tamb\in\TambAsVec$ in the discrete set of measurements, there is a unique vector of radial coefficients $\coefficientsRadiiVec[\tamb]$:
\begin{subequations}\label{eq:spatialRadiiFit}
    \begin{align}
        \mat{P}                      & = \sqrt{\mat{H}^2 + \mat{W}^2},\quad\quad \mat{P}\in\mathcal{R}^{h\times w}                                                                                                                                            \\
        \coefficientsRadiiVec[\tamb] & =\argmin_{\coefficientsRadiiVec[\tamb]}\left(\sum_{q=0}^{\mSpatial}\sum_{z=0}^{\mSpatial}  \coefficientsSpatialSkewless[\tamb][q,z]\cdot\mat{H}^{q}\cdot \mat{W}^{z} - \sum_{r=0}^{\mRadial} \coefficientsRadii{r}[\tamb]\cdot\mat{P}^r\right),\qquad\coefficientsRadiiVec[\tamb]\in\mathcal{R}^{\mRadial}
    \end{align}
\end{subequations}
where $\mRadial$ is the number of coefficients in the radial fit and $\coefficientsRadii{r}\nolinebreak\in\nolinebreak\mathcal{R}^{\mRadial}$ are the radial fitting coefficients.

    
The last step in the estimation process is to express a polynomial approximation of the radial coefficients by $\tamb$; specifically, to approximate $\coefficientsRadiiVec(t)$ in the continuous range $t\in\TambAsVec$ for $\NumFPA$ the length of $\Tamb$.
We make several denotations. $\mat{A}_C$ is a matrix containing the powers of all $\tamb\in\TambAsVec$:
\begin{equation}
    \mat{A}_C =
    \begin{bmatrix}
        \Tamb^0[1]       & \cdot & \Tamb^{\mFPA}[1]       \\
        \vdots           & \cdot & \vdots                 \\
        \Tamb^0[\NumFPA] & \cdot & \Tamb^{\mFPA}[\NumFPA]
    \end{bmatrix}_{\NumFPA\times \mFPA}
\end{equation}

$\coefficientsRadiiVec\{m\}$ is a vector of the $m$th coefficient in $\coefficientsRadiiVec$ for all $\tamb\in\TambAsVec$:
\begin{equation}
    \coefficientsRadiiVec\{r\} =
    \begin{bmatrix}
        \coefficientsRadiiVec\left[\Tamb[0]\right][r] \\
        \vdots                                        \\
        \coefficientsRadiiVec\left[\Tamb[\NumFPA]\right][r]
    \end{bmatrix}_{\NumFPA\times1}
\end{equation}

$\mat{R}_{C}$ is a matrix of all the $\coefficientsRadiiVec\{r\}$:
\begin{equation}
    \mat{R}_{C} =
    \begin{bmatrix}
        \coefficientsRadiiVec\left[\Tamb[0]\right][0] & \hdots & \coefficientsRadiiVec\left[\Tamb[0]\right][\mRadial] \\
        \vdots           & \ddots & \vdots                  \\
        \coefficientsRadiiVec\left[\Tamb[\NumFPA]\right][0] & \hdots & \coefficientsRadiiVec\left[\Tamb[\NumFPA]\right][\mRadial] \\
    \end{bmatrix}_{\NumFPA\times \mRadial},\qquad\forall r\in[0,\ldots,\mRadial]
\end{equation}

$\Gamma$, the radial coefficient matrix, is comprised of $\gamma_{i,j}$ coefficients:
\begin{equation}
    \Gamma =
    \begin{bmatrix}
        \gamma_{0,0}     & \hdots & \gamma_{0,\mRadial}     \\
        \vdots           & \ddots & \vdots                  \\
        \gamma_{\mFPA,0} & \hdots & \gamma_{\mFPA,\mRadial}
    \end{bmatrix}_{\mFPA\times\mRadial} \label{eq:fitRadiiFPA:coef}
\end{equation}

Then the values of $\Gamma$ can be found by solving the inverse problem:
\begin{equation}\label{eq:fitRadiiFPA}
    \mat{R}_C = \mat{A}_C \cdot \Gamma \quad \rightarrow\quad
    \Gamma = \mat{A}_C^{+}\cdot\mat{R}_C,\qquad\Gamma\in\mathcal{R}^{\mFPA\times \mRadial}
\end{equation}
$\mat{A}_C^{+}$ is the Moore–Penrose inverse of $\mat{A}_C$.

We denote the approximation of $\coefficientsRadiiVec[\tamb]$ by $\tamb$ as $\coefficientsAmbRadiiVec(\tamb)$.
\cref{fig:coefRadial} demonstrates how $\coefficientsPixelwiseVec(\tamb)$ are found using $\Gamma$.
The algorithm used to find $\Gamma$ the $\tamb$-dependent radial nonuniformity coefficients is depicted in \cref{alg:estimateNonUniformity}.
\newcommand{\pathToFigsRadiiCoefsGamma}{supp_figures/coef/}
\newcommand{\widthFigs}{1cm}
\newcommand{\fontSize}{\huge}
\newcommand{\fontCaptionBelow}{\Large}
\begin{figure*}
    \centering
    \begin{tikzpicture}[node distance=centered, auto, thick]
        \node[inner sep=0pt] (beta) {\includegraphics[width=\widthFigs]{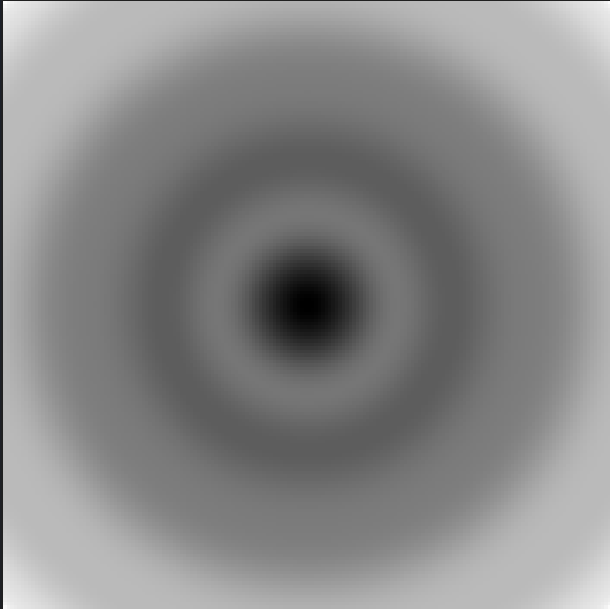}};
        \node[below =-0.1cm of beta, font=\fontCaptionBelow] {$\coefficientsPixelwiseVec(\tamb)$};
        \node[inner sep=0pt, font=\fontSize] (approx) at ($(beta.east)+(1em,0)$) {$\approx$};
        \node[inner sep=0pt, font=\fontSize] (p0) at ($(approx.east)+(2.5em,0)$) {$\coefficientsAmbRadii{0}\cdot($};
        \node[inner sep=0pt, font=\fontSize] (p0Fig) at ($(p0.east)+(1.5em,0)$) {\includegraphics[width=\widthFigs]{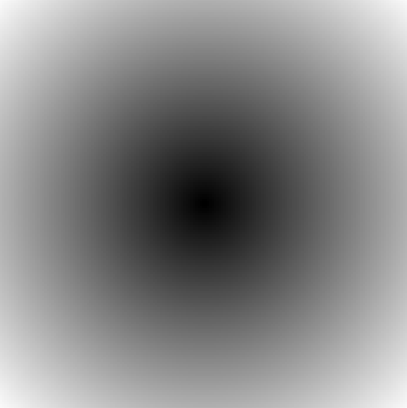}};
        \node[below =-0.1cm of p0Fig, font=\fontCaptionBelow] {$\mat{P}^0$};
        \node[inner sep=0pt, font=\fontSize] (p0Parentheses) at ($(p0Fig.east)+(3.4ex,0)$) {$)^0+$};
        \node[inner sep=0pt, font=\fontSize] (p1) at ($(p0Parentheses.east)+(2.5em,0)$) {$\coefficientsAmbRadii{1}\cdot($};
        \node[inner sep=0pt, font=\fontSize] (p1Fig) at ($(p1.east)+(1.5em,0)$) {\includegraphics[width=\widthFigs]{\pathToFigsRadiiCoefsGamma/p3.png}};
        \node[below =-0.1cm of p1Fig, font=\fontCaptionBelow] {$\mat{P}^1$};
        \node[inner sep=0pt, font=\fontSize] (p1Parentheses) at ($(p1Fig.east)+(3.4ex,0)$) {$)^1+$};
        \node[inner sep=0pt, font=\fontSize] (p2) at ($(p1Parentheses.east)+(2.5em,0)$) {$\coefficientsAmbRadii{2}\cdot($};
        \node[inner sep=0pt] (p2Fig) at ($(p2.east)+(1.5em,0)$) {\includegraphics[width=\widthFigs]{\pathToFigsRadiiCoefsGamma/p3.png}};
        \node[below =-0.1cm of p2Fig, font=\fontCaptionBelow] {$\mat{P}^2$};
        \node[inner sep=0pt, font=\fontSize] (p2Parentheses) at ($(p2Fig.east)+(0.8em,0)$) {$)^2+$};
        \node[inner sep=0pt, font=\fontSize] (p3) at ($(p2Parentheses.east)+(2.5em,0)$) {$\coefficientsAmbRadii{3}\cdot($};
        \node[inner sep=0pt] (p3Fig) at ($(p3.east)+(1.5em,0)$) {\includegraphics[width=\widthFigs]{\pathToFigsRadiiCoefsGamma/p3.png}};
        \node[below =-0.1cm of p3Fig, font=\fontCaptionBelow] {$\mat{P}^3$};
        \node[inner sep=0pt, font=\fontSize] (p3Parentheses) at ($(p3Fig.east)+(0.8em,0)$) {$)^3$};
    \end{tikzpicture}
    \caption{Example of applying the radial coefficients $\coefficientsAmbRadiiVec$ in \cref{eq:estimationOfMeas:fpa}.
    The number of coefficients in the radial fit $\mRadial=3$.
    The results of the summation in the figure are the coefficients used to approximate the radiance $\coefficientsPixelwiseVec\in\mathcal{R}^{\mGrayLevels\times h\times w}$.}
    \label{fig:coefRadial}
\end{figure*}

\subsection{Applying the camera simulator to new data}\label{sec:method:characterization:applyToNew}
\noindent We can now use the coefficients $\Gamma$ to simulate the low-cost IR camera response in gray-levels from a new object temperature $\tobj$ and ambient temperature $\tamb$.
First, the radial coefficients are estimated for a given ambient temperature $\tamb$:
\begin{equation}\label{eq:estimationOfMeas:sp}
    \coefficientsAmbRadiiVec(\tamb) = \sum_{m=0} ^{\mFPA}{\Gamma[m] \cdot \tamb^m},\qquad\coefficientsAmbRadiiVec(\tamb)\in\mathcal{R}^{\mRadial}
\end{equation}

Next, the pixel-wise coefficients are found from the radial coefficients $\coefficientsAmbRadiiVec(\tamb)$ and the radial kernel $\mat{P}$:
\begin{equation} \label{eq:estimationOfMeas:fpa}
    \coefficientsPixelwiseVec(\tamb)[x,y] = \sum_{m=0}^{\mRadial}{\coefficientsAmbRadii{m}(\tamb)\cdot P^m},\qquad\coefficientsPixelwiseVec(\tamb)\in\mathcal{R}^{\mGrayLevels\times h\times w}
\end{equation} Notice that the coefficients $\coefficientsPixelwiseVec(\tamb)$ are pixel-wise and that for $P^m$, the power $m$ is applied element-wise on the matrix $P$.

Lastly, the estimated response of the camera $\simulatedCameraResponse{\tamb}{\tobj}$ is approximated using the given object temperature and the pixel-wise coefficients $\coefficientsPixelwiseVec(\tamb)$. The equation is stated at the beginning of \cref{sec:method:characterization:temperatureDependency}, as \cref{eq:supp:charactarization:simulation}:
$$ \simulatedCameraResponse{\tamb}{\tobj}[x,y] = \sum_{m=0}^{\mGrayLevels}{\coefficientsPixelwise{m}(\tamb)[x,y]} \cdot \tobj^m[x,y],\qquad\simulatedCameraResponse{\tamb}{\tobj}\in\mathcal{R}^{h\times w} $$
where $\tobj$ can have dimensions similar to a frame. The power $m$ is applied element-wise.

The simulator is illustrated in \cref{fig:method:simulator} of the main article.
%
%
%
\begin{algorithm}
    \SetAlgoLined
    \KwData{Images of \blackbody blackbody at different \opPoints.}
    \KwIn{$\mathbf{\mGrayLevels}$ is the degree of the polynomial of the object's temperature.\newline
        $\mathbf{\mFPA}$ is the degree of the polynomial.\newline
        $\mathbf{M_F}$ is the degree of the fine spatial fit.\newline
        $\mathbf{\mRadial}$ is the degree of the radial fit.\newline
        $\mathbf{\mFPA}$ is the degree of the camera temperature fit for the radial coefficients.}
    \KwOut{The $\tamb$-dependent radial nonuniformity coefficients $\Gamma\in\mathcal{R}^{\mFPA\times \mRadial}$.}
    \For{$\tamb\in \Tamb  $}{
        $\coefficientsRadiance{}[\tamb][x,y]\longleftarrow$ \cref{eq:LsSolutionFPA}, $\forall x,y\in$ image}
    \For{$m\in[0,\ldots,\mGrayLevels]$}{
        \For{$\tamb\in \Tamb  $}{
            Gaussian filter on $\coefficientsRadiance{m}[\tamb]$\\
            $\coefficientsSpatialQuadratic[\tamb][m]\longleftarrow$ Quadratic spatial fit (\cref{eq:spatialFit})\\
            $\coefficientsSpatialFine[\tamb][m]\longleftarrow$ Fine spatial fit (\cref{eq:spatialFit})\\
            $\coefficientsSpatialSkewless[\tamb][m]\longleftarrow$ Subtract the spatial fit (\cref{eq:spatialFitSubtraction})\\
            $\coefficientsRadiiVec[\tamb][m]\longleftarrow$ Radial spatial fit (\cref{eq:spatialRadiiFit})}
        $\Gamma[m]\longleftarrow$ Fit radial coefficients to $\tamb$ (\cref{eq:fitRadiiFPA})}
    \caption{Estimation of nonuniformity maps.}
    \label{alg:estimateNonUniformity}
\end{algorithm}

\begin{algorithm}
    \SetAlgoLined
    \KwIn{$\mathbf{\tobj}$ is an accurate temperature map of arbitrary dimensions.\newline
        $\mathbf{\tamb}$ to simulate the camera ambient temperature.\newline
        $\mathbf{\Gamma}$ the $\tamb$-dependent radial nonuniformity coefficients.}
    \KwOut{$\simulatedCameraResponse{\tamb}{\tobj}$ a gray-level map simulation at temperature $\tamb$.}
    $\simulatedCameraResponse{\tamb}{\tobj}\longleftarrow\mat{0}_{h\times w}$\\
    \For{$m\in[0,\ldots,\mGrayLevels]$}{
        $\coefficientsAmbRadii{m}(\tamb)\longleftarrow$  Calc.\ radial coefficients (\cref{eq:estimationOfMeas:sp})\\
        $\coefficientsPixelwise{m}(\tamb)\longleftarrow$ Calc.\ per-pixel coefficients using the radial coefficients $\coefficientsAmbRadii{m}$ (\cref{eq:estimationOfMeas:fpa})\\
        \tcp{Multiplication and power of matrices are applied element-wise.}
        $\simulatedCameraResponse{\tamb}{\tobj}\longleftarrow\simulatedCameraResponse{\tamb}{\tobj}+\coefficientsPixelwise{m}(\tamb)\cdot \tobj^m$}
    \caption{Transform an accurate temperature map of arbitrary dimensions into a nonuniform radiation flux.}
    \label{alg:applyNonUniformity}
\end{algorithm}

\FloatBarrier

\section{Network}\label{sec:supp:network}
    The nonuniformity is space-variant (\cref{sec:intro}), so the network needs a large receptive field to correct the entire image. The U-Net architecture is a tradeoff between the receptive field of the network and the computational requirements. $\nLevels$ is the coefficient of the spatial resolution and $\nChannels$ is the coefficient for the number of channels. At each level, the spatial resolution decreases by $2^\nLevels$\ and the number of channels increases by $2^\nChannels$. The deeper levels of the network increase the receptive field because each convolution kernel affects a larger area in the original frame. The computation requirements decrease by a factor of 2 for each encoder level~\cite{Oz2020}.
    \begin{figure*}
        \centering
        \includegraphics[width=\linewidth]{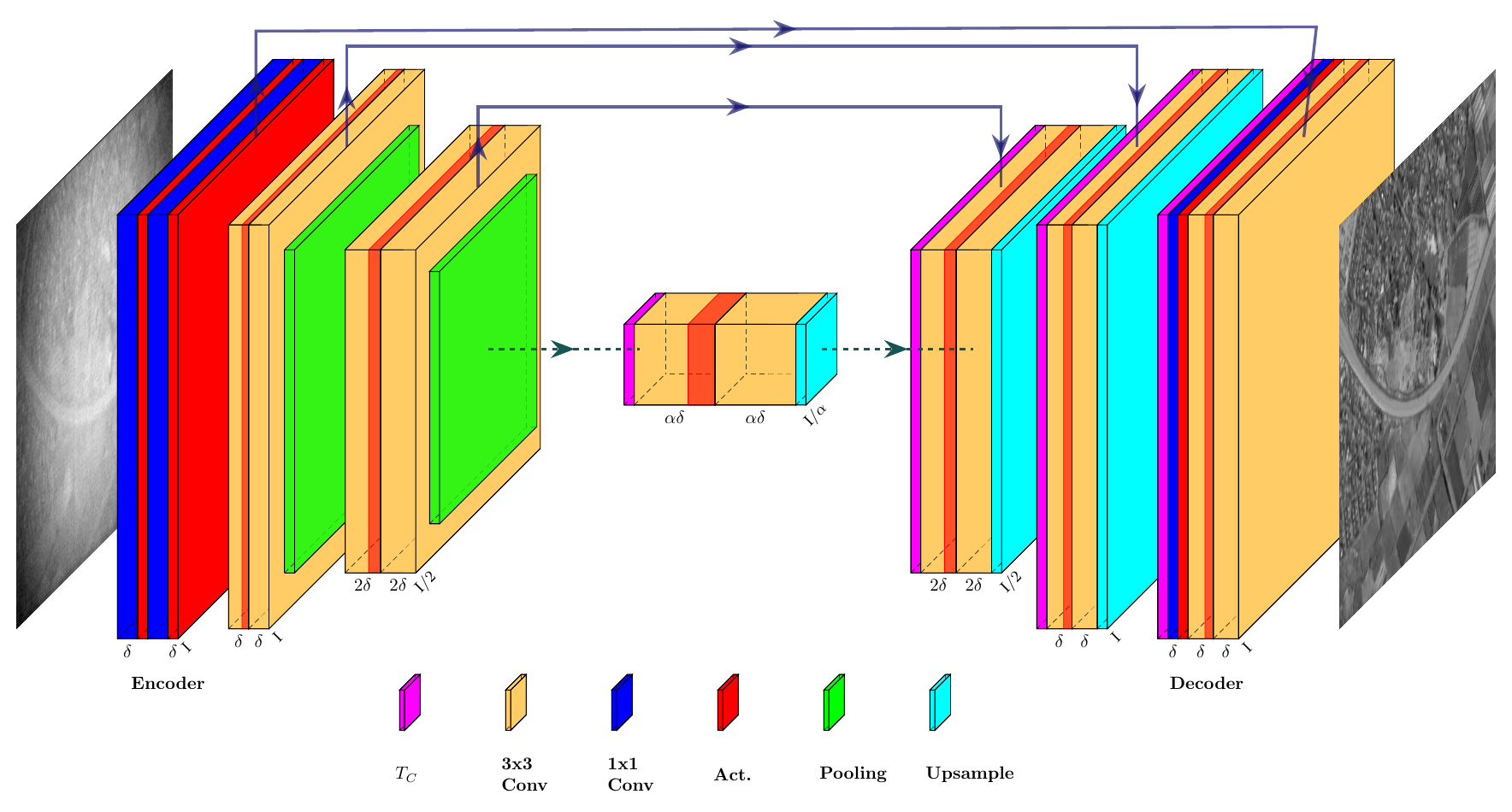}
        \caption{Architecture of the backbone U-Net estimation network~\cite{unet}.
        The input is a frame with gray level values, and the output is a temperature map. This figure shows the end-to-end (E2E) configuration. In the network with a linear physical constraint (GxPD), the final block is replaced by two identical blocks as described in \cref{sec:method:net}.}
        \label{fig:supp:network}
    \end{figure*}
\FloatBarrier

\section{List of figures}
\begin{figure*}
    \centering
    \includegraphics[width=\linewidth]{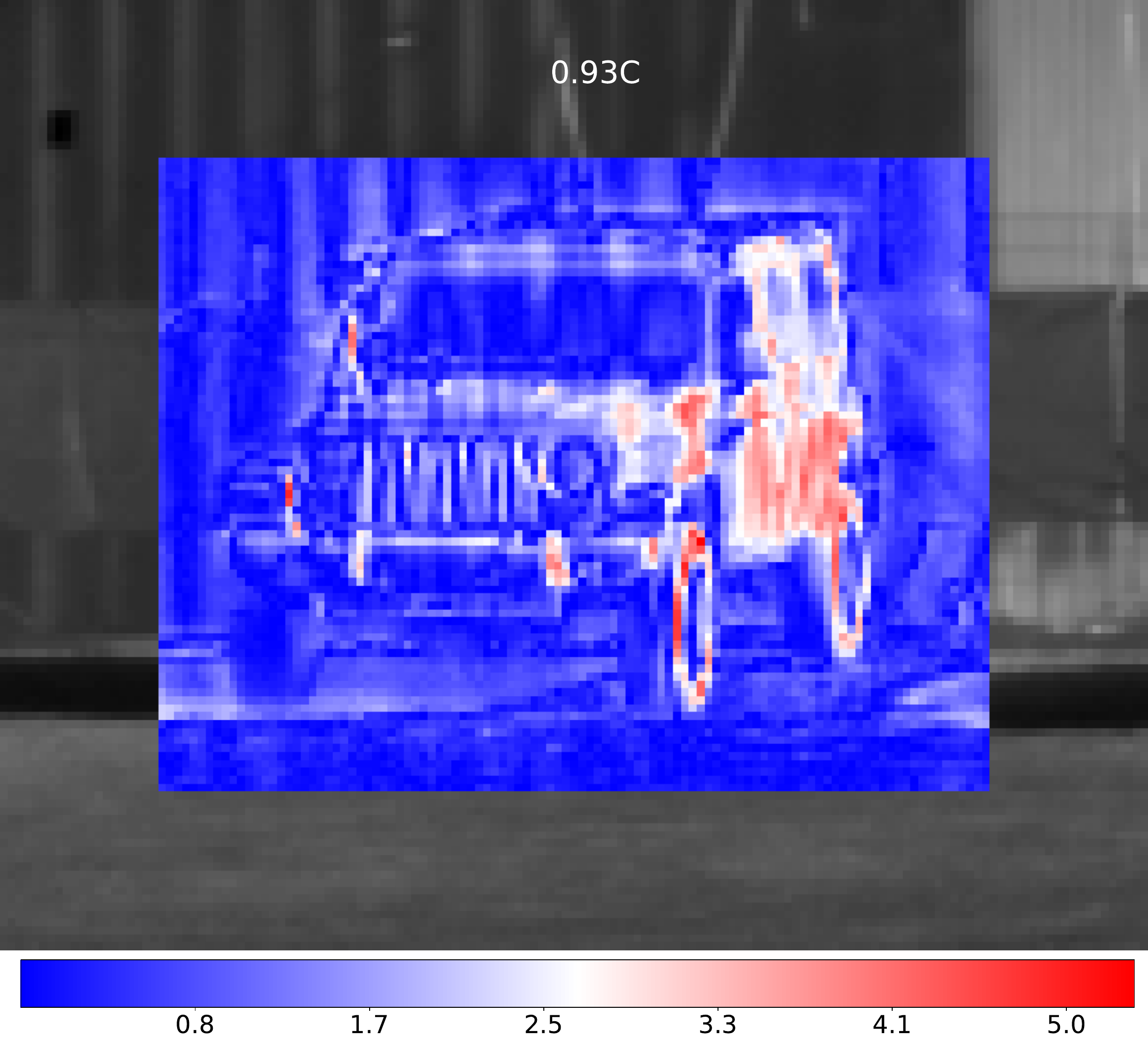}
    \caption{Results of the proposed method on an image of a jeep. The metallic surfaces of the jeep reflects direct sun from the east (right side of the frame), stemming in higher error.}
    \label{jeep}
\end{figure*}

\begin{figure*}
    \centering
    \includegraphics[width=\linewidth]{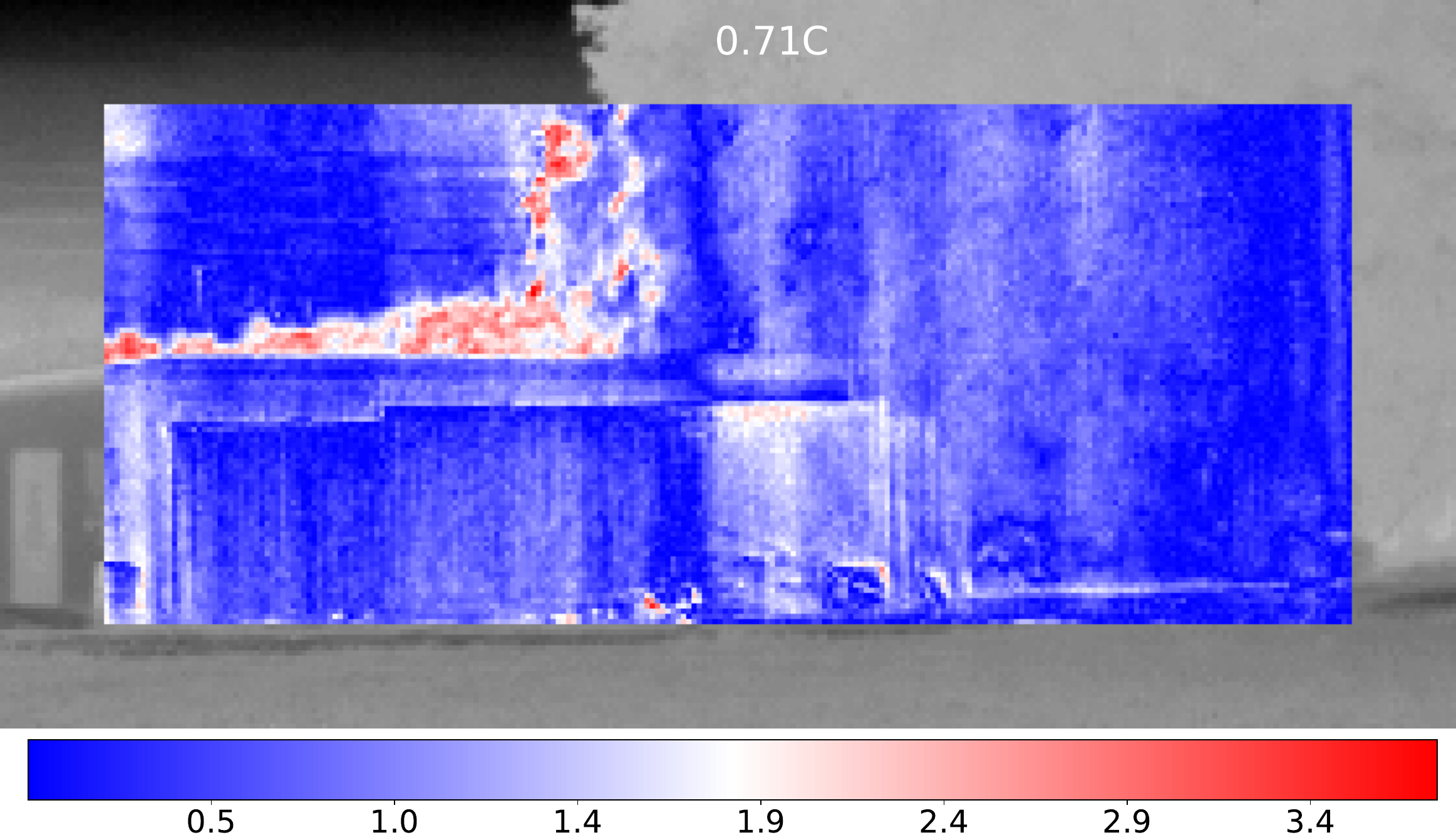}
    \caption{Results of the proposed method on an image of a shed. The background of the shed in the frame is a huge dense tree. The high error in the frame stems from moving leaves on the edge of the tree.}
    \label{shed}
\end{figure*}

\begin{figure*}
    \centering
    \includegraphics[width=\linewidth]{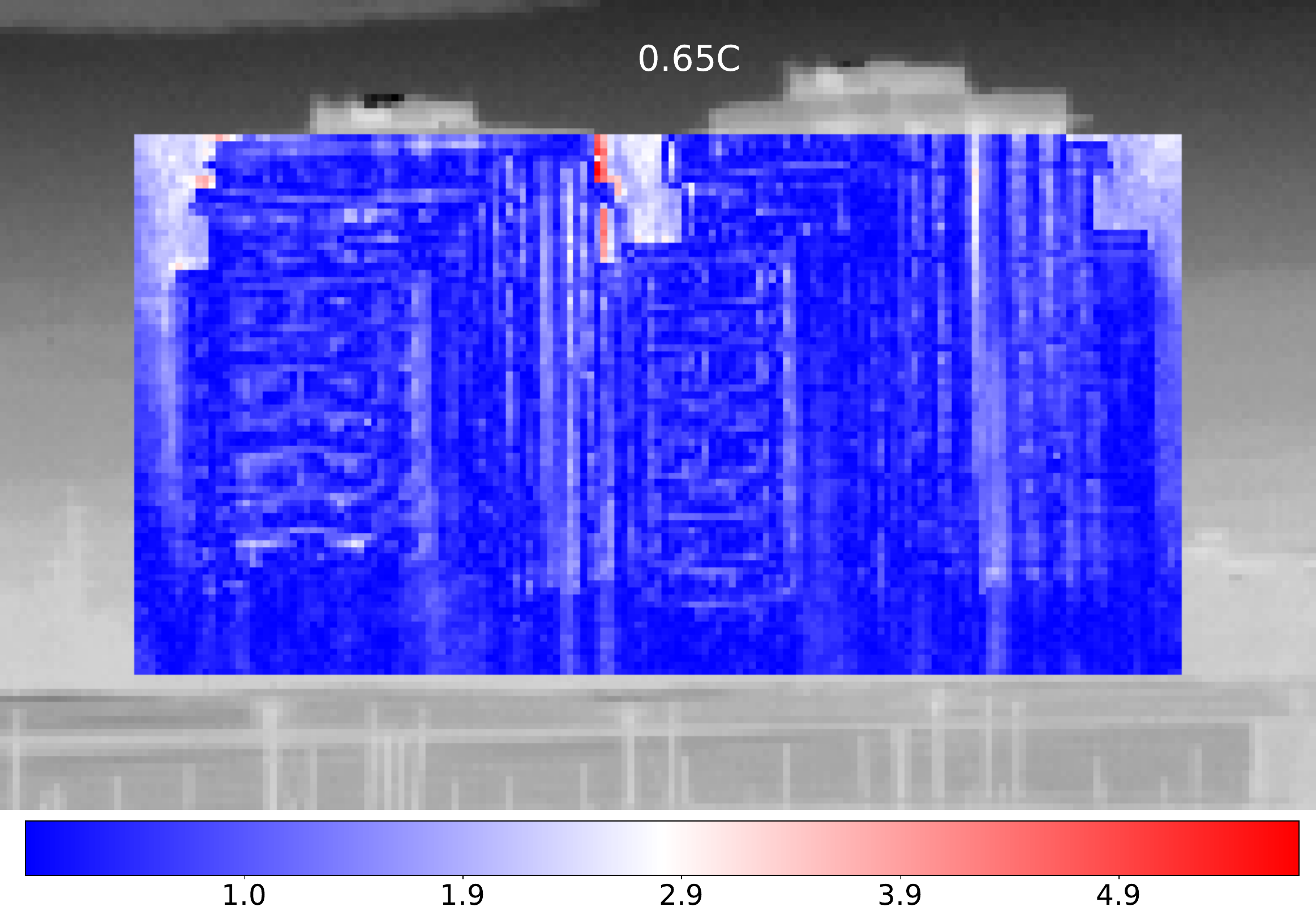}
    \caption{Results of the proposed method on an image of a building.}
    \label{building}
\end{figure*}

\begin{figure*}
    \centering
    \includegraphics[width=\linewidth]{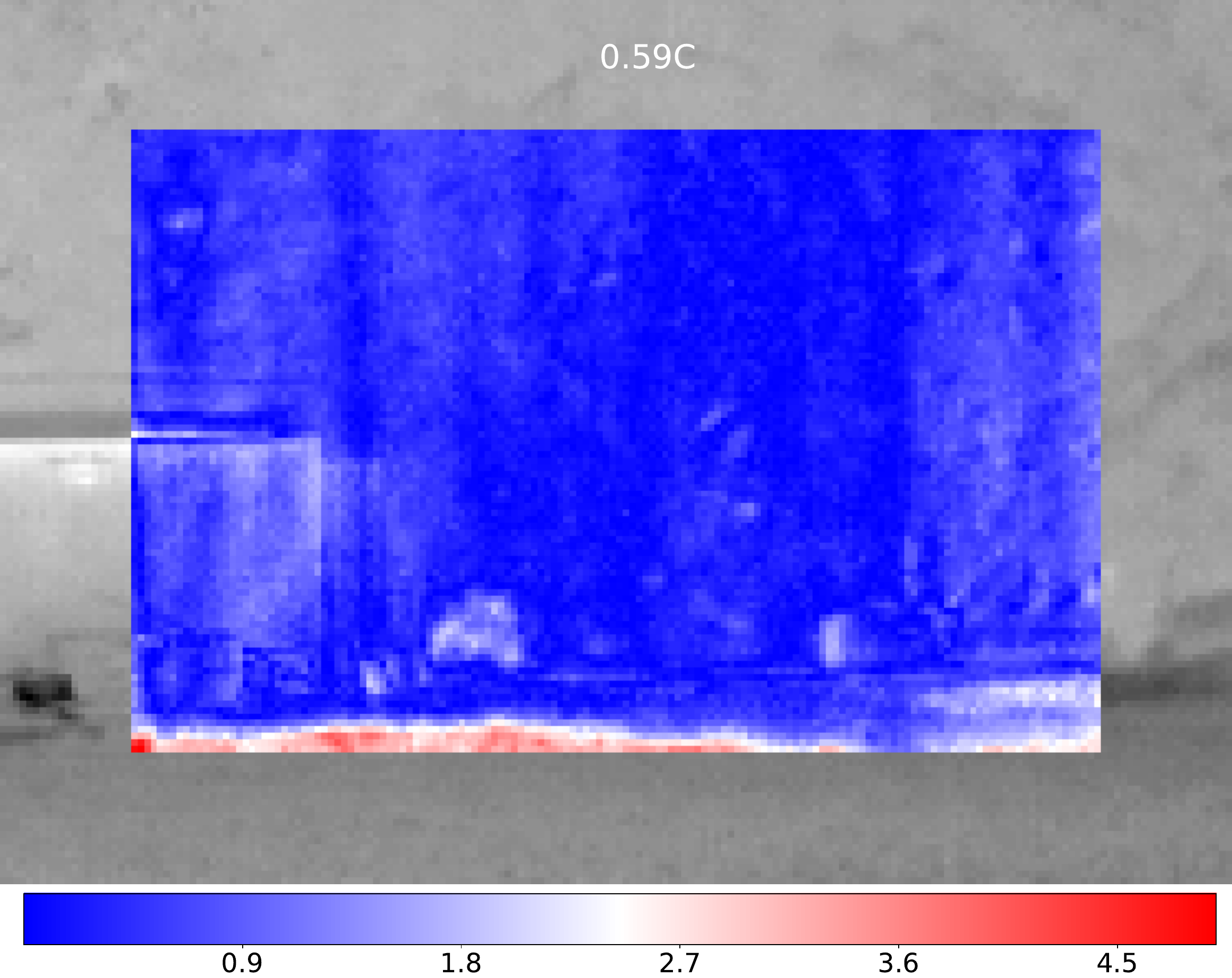}
    \caption{Results of the proposed method on an image of a shed.}
    \label{shed2}
\end{figure*}

\begin{figure*}
    \centering
    \includegraphics[width=\linewidth]{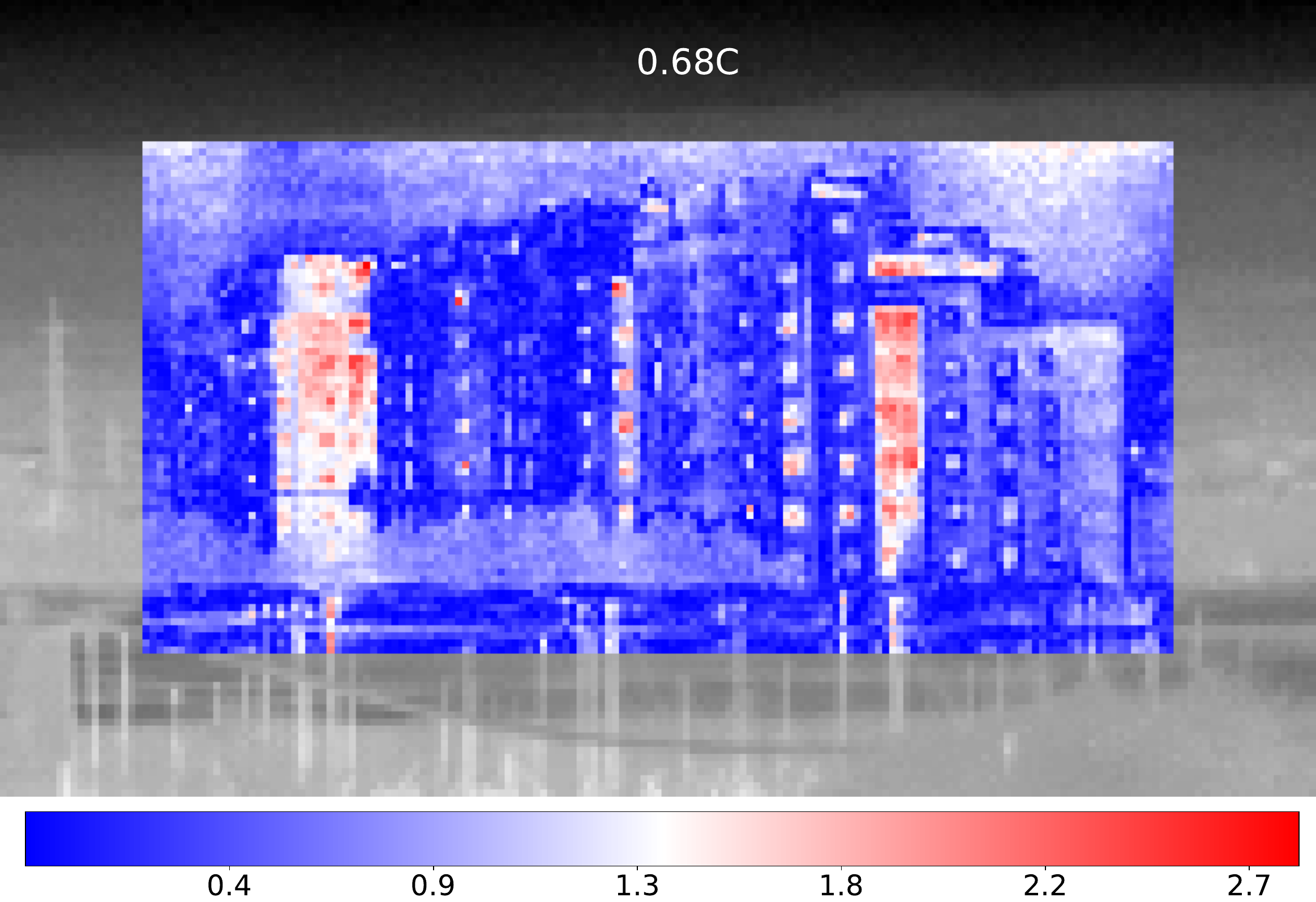}
    \caption{Results of the proposed method on an image of a building. The errors stems from direct sunlight reflecting from glass surfaces (the sun is in the back of the camera).}
    \label{building_sunlight}
\end{figure*}

\begin{figure*}
    \centering
    \includegraphics[width=\linewidth]{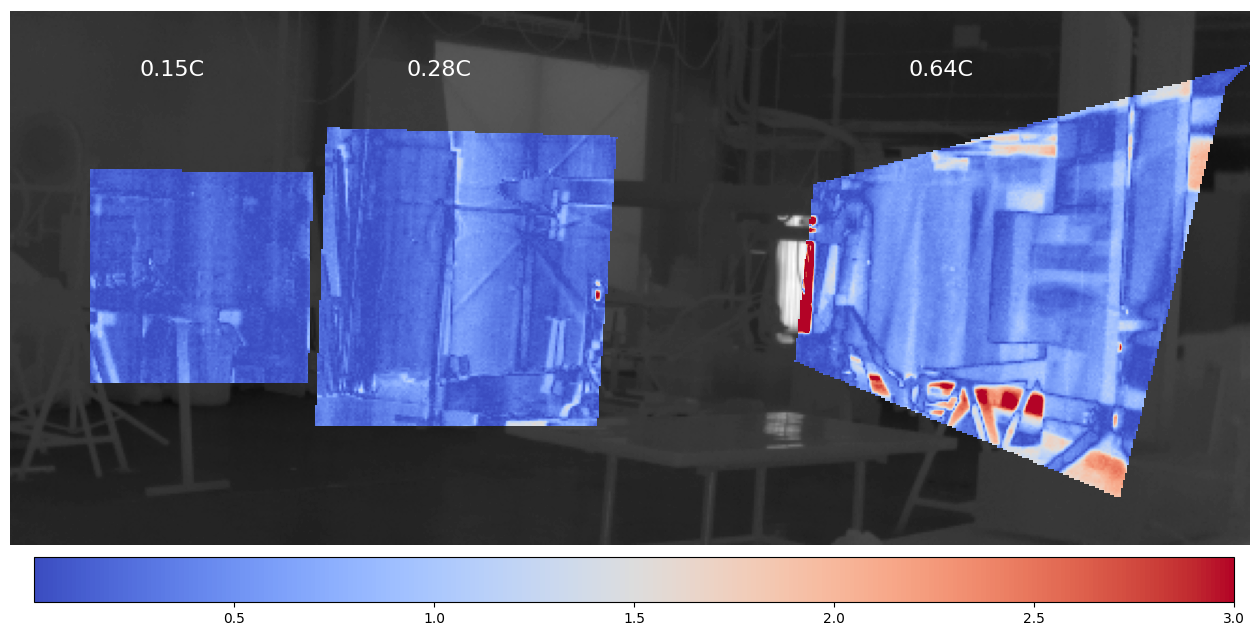}
    \caption{Results of the proposed method inside a warehouse. The image was taken from a close distance ($15_m-20_m$), so the registration between the two cameras was imperfect. The high error on the right frame stems from the registration error.}
    \label{warehouse}
\end{figure*}

\begin{figure*}
    \centering
    \includegraphics[width=\linewidth]{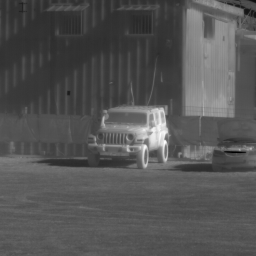}
    \caption{The original frame taken by \taucamera of the jeep in \cref{jeep} in the supplementary materiel.}
    \label{jeep_real}
\end{figure*}

\begin{figure*}
    \centering
    \includegraphics[width=\linewidth]{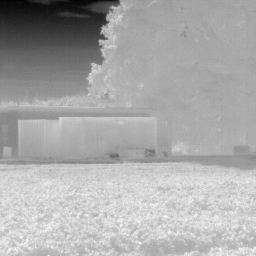}
    \caption{The original frame taken by \taucamera of the shed in \cref{shed} in the supplementary materiel. Notice the leaves at the edge of the tree.}
    \label{shed_real}
\end{figure*}

\begin{figure*}
    \centering
    \includegraphics[width=\linewidth]{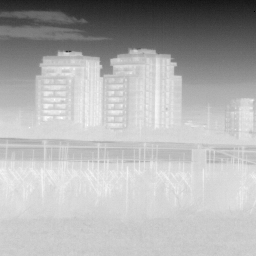}
    \caption{The original frame taken by \taucamera of the building in \cref{building} in the supplementary materiel.}
    \label{building_real}
\end{figure*}

\begin{figure*}
    \centering
    \includegraphics[width=\linewidth]{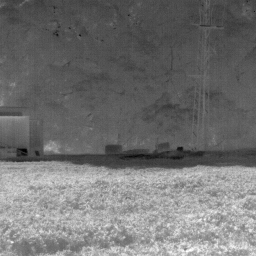}
    \caption{The original frame taken by \taucamera of the shed in \cref{shed2} in the supplementary materiel.}
    \label{shed2_real}
\end{figure*}

\begin{figure*}
    \centering
    \includegraphics[width=\linewidth]{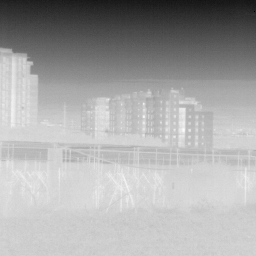}
    \caption{The original frame taken by \taucamera of the building with reflections from the sunlight in \cref{building_sunlight} in the supplementary materiel.}
    \label{building_sunlight_real}
\end{figure*}

\begin{figure*}
    \centering
    \subfloat{\includegraphics[width=0.31\linewidth]{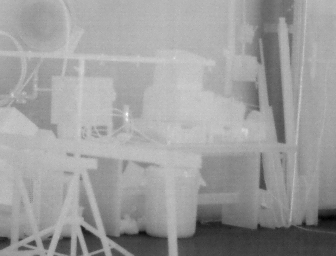}}
    \hfill
    \subfloat{\includegraphics[width=0.31\linewidth]{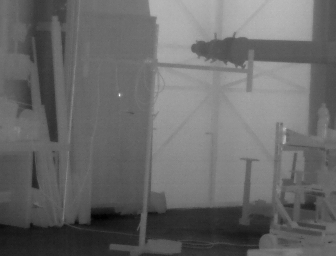}}
    \hfill
    \subfloat{\includegraphics[width=0.31\linewidth]{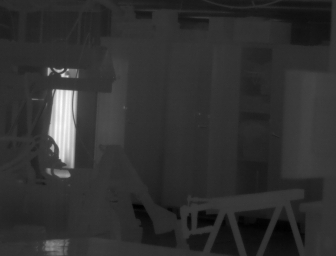}}
    \caption{The original frame taken by \taucamera of the warehouse in \cref{warehouse} in the supplementary materiel.}
    \label{warehouse_tau_real}
\end{figure*}

\begin{figure*}
    \centering
    \includegraphics[width=\linewidth]{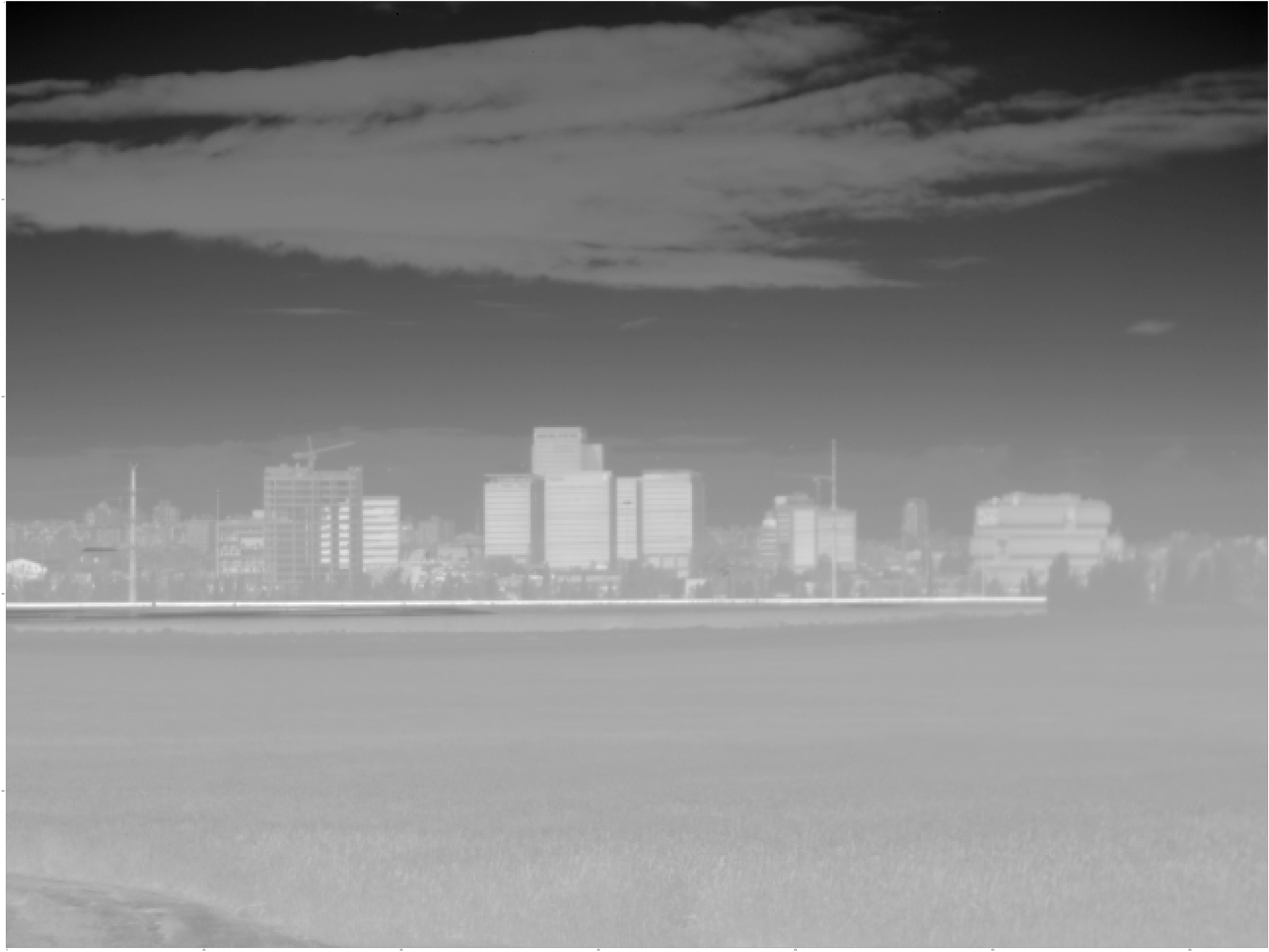}
    \caption{The original frame taken by \scientificCamera of the building \cref{fig:results:realData} in the article.}
    \label{gt_lab}
\end{figure*}

\begin{figure*}
    \centering
    \includegraphics[width=\linewidth]{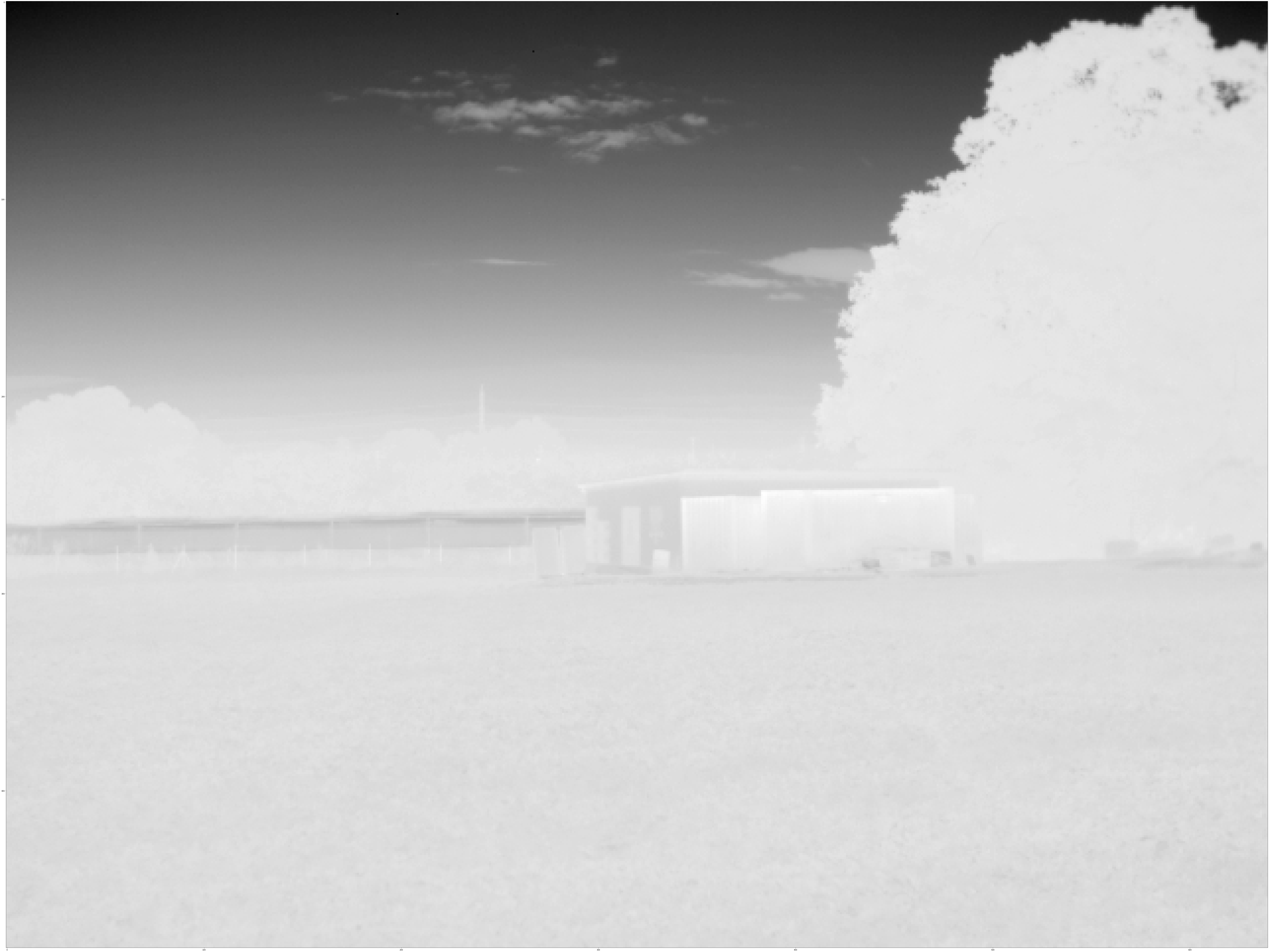}
    \caption{The original frame taken by \scientificCamera of the shed in \cref{shed,shed2,shed_real,shed2_real} in the supplementary materiel.}
    \label{gt_shed}
\end{figure*}

\begin{figure*}
    \centering
    \includegraphics[width=\linewidth]{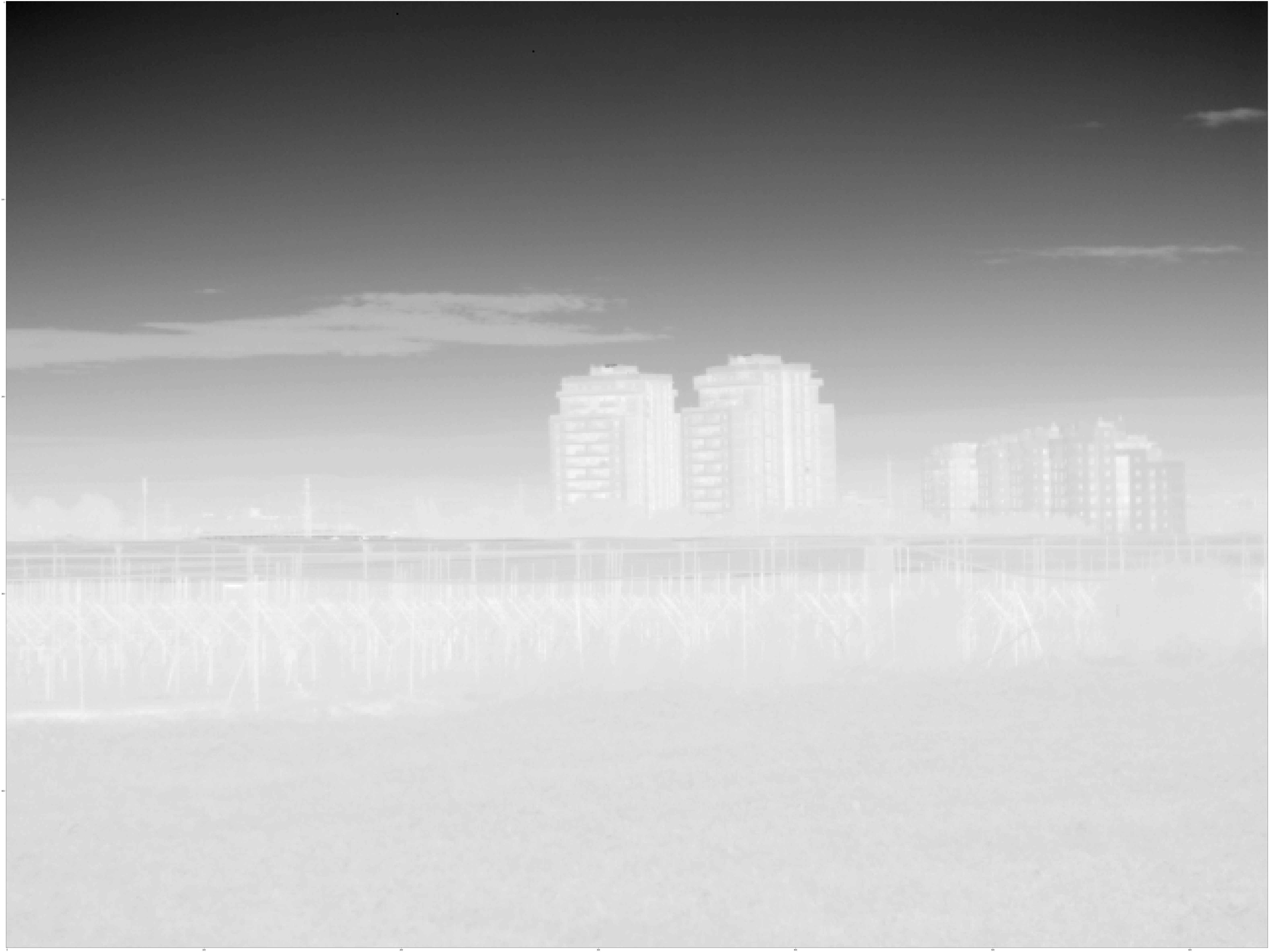}
    \caption{The original frame taken by \scientificCamera of the shed in \cref{building_sunlight,building_sunlight_real} in the supplementary materiel.}
    \label{gt_building_sunlight}
\end{figure*}

\begin{figure*}
    \centering
    \includegraphics[width=\linewidth]{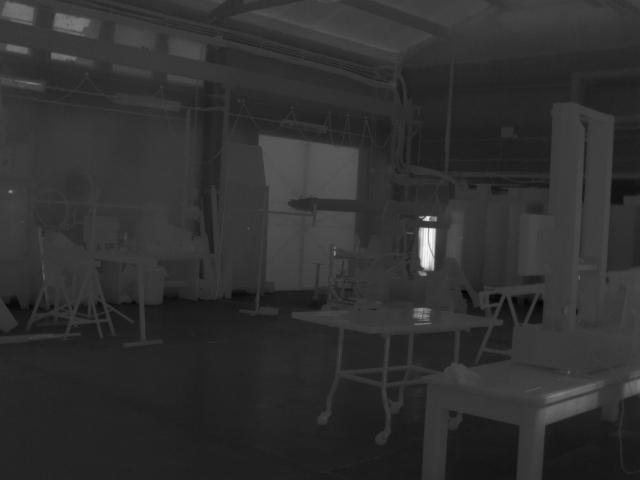}
    \caption{The original frame taken by \scientificCamera of the warehouse in \cref{warehouse} in the supplementary materiel.}
    \label{gt_warehouse}
\end{figure*}

\begin{figure*}[b]
    \centering
    \includegraphics[width=0.8\linewidth]{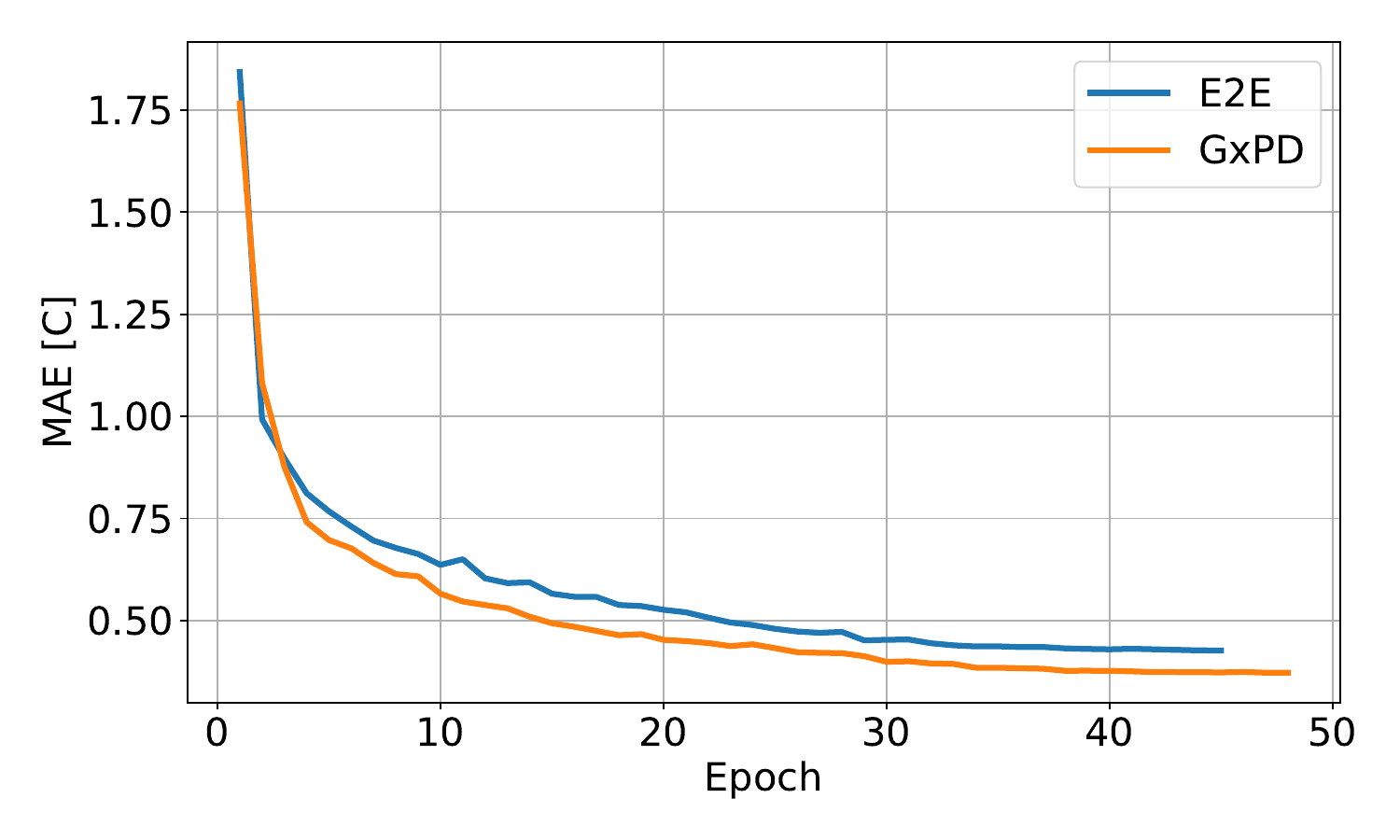}
    \caption{Validation mean estimation error (MAE) in $^\circ C$ for the training of E2E and GxPD.}
    \label{fig:results:convergence}
\end{figure*}

\begin{enumerate}
    \item \cref{jeep,shed,building,shed2,building_sunlight,warehouse} shows the results of the temperature estimation on real-world data. The temperature map taken with the \scientificCamera serves as the gray background, and the colored map is the difference between the results of GxPD with the temperature map. The number in white as the MAE in $\circ C$ between the temperature map and the results of GxPD.
    \item \cref{jeep_real,shed_real,building_real,shed2_real,building_sunlight_real,warehouse_tau_real} are some of the original frames taken by the \taucamera.
    \item \cref{gt_lab,gt_shed,gt_building_sunlight,gt_warehouse} are some of the ground truth frames taken by the \scientificCamera. Notice that not all of the original frames could be displayed due to privacy limitations.
    \item \cref{fig:supp:patch:1,fig:supp:patch:2,fig:supp:patch:3,fig:supp:patch:4,fig:supp:patch:5,fig:supp:patch:6,fig:supp:patch:7,fig:supp:patch:8,fig:supp:patch:9,fig:supp:patch:10,fig:supp:patch:11,fig:supp:patch:12} are results of our GxPD model. A zoomed patch from the sample is presented (surrounded by a red square). From left to right - (a) the input sample, (b) the ground truth temperature map, (c) GxPD (ours), (d) ADMIRE~\cite{admire} (e) He et al.~\cite{He2018} and (f) SNRWDNN~\cite{snrwdnn}.
    \item \cref{fig:results:convergence} shows the convergence of the training for both E2E and GxPD networks.
    \item \cref{tab:hyperparams} are the hyperparameters of the final networks used for all the metrics and figures.
    \item \cref{tab:cameraParams} are the parameters of the \taucamera used throughout the experiments.
\end{enumerate}

\section{Data}
The training dataset was $12,897$ frames of different agricultural fields with dimensions $640\times 480$ pixels.
The validation set was comprised of $4,723$ frames. All frames were of different agricultural fields in Israel, taken from an unmanned aerial vehicle (UAV) flying $70_m-100_m$ above the ground.
369 frames of corn fields taken in \textit{Tzora} village,
646 frames of peach trees taken in \textit{Nir Eliyahu} village,
$1,162$ frames of wheat fields taken in \textit{Neve Yaar} research station,
372 frames of vineyards taken in \textit{Mevo Beitar} village,
765 frames of vineyards taken in \textit{Mevo Beitar} village on a different day,
$1,048$ frames of cotton fields taken in \textit{Neve Yaar} research station,
and 361 frames of wheat fields taken in \textit{Gilat} research station - for a total of $4,723$ frames for validation.

The collection of data for the comparison between \scientificCamera and \taucamera is elaborated in the article.

\begin{table}[h]
    \centering
    \caption{Hyperparameters used in end-to-end (E2E) and physically constrained (GxPD) networks.}
    \begin{tabular}{|c|c|c|}
        \hline
        Network                               & E2E                                   & GxPD                          \\
        \hline
        $\nChannels$ (number of filters)      & \multicolumn{2}{c|}{32}                                               \\
        \hline
        $\nLevels$ (number of layers in UNet) & \multicolumn{2}{c|}{6}                                                \\
        \hline
        Normalization                         & None                                  & Instance \cite{instance_norm} \\
        \hline
        Activation                            & \multicolumn{2}{c|}{GeLU \cite{GeLU}}                                 \\
        \hline
        $\beta$ (DSSIM loss)                  & \multicolumn{2}{c|}{0.01}                                             \\
        \hline
        $\gamma$ (TV loss)                    & 0.001                                 & 0.0001                        \\
        \hline
    \end{tabular}
    \label{tab:hyperparams}
\end{table}
\begin{table}[h]
    \centering
    \caption{The FLIR \taucamera settings as described in Tau2 Quark Software IDD}
    \begin{tabular}{|c|c||c|c|}
        \hline
        Function   & State  & Function        & State                  \\
        \hline\hline
        FFC Mode   & Auto   & FPS             & 4 ($60_{Hz}$)          \\
        \hline
        FFC Period & 0      & CMOS Depth      & 0 ($14_{bit}$ w/o AGC) \\
        \hline
        Isotherm   & 0      & LVDS            & 0                      \\
        \hline
        DDE        & 0      & LVDS Depth      & 0 ($14_{bit}$)         \\
        \hline
        T-Linear   & 0      & XP              & 2 ($14_{bit}$)         \\
        \hline
        AGC        & Manual & Brightness Bias & 0                      \\
        \hline
        Contrast   & 0      & Brightness      & 0                      \\
        \hline
        ACE        & 0      & SSO             & 0                      \\
        \hline
        Gain       & High   &                 &                        \\
        \hline
    \end{tabular}
    \label{tab:cameraParams}
\end{table}


\begin{figure*}
    \figPatchesLabelSuppMat{180725_Ramon}{0}{3}{5}{8}{9}{10}
    \label{fig:supp:patch:1}
\end{figure*}
\begin{figure*}
    \figPatchesLabelSuppMat{180725_Ramon}{14}{15}{27}{29}{36}{68}
    \label{fig:supp:patch:2}
\end{figure*}
\begin{figure*}
    \figPatchesLabelSuppMat{180725_Ramon}{84}{100}{108}{132}{11}{82}
    \label{fig:supp:patch:3}
\end{figure*}
\begin{figure*}
    \figPatchesLabelSuppMat{Gilat_210809}{15}{25}{33}{56}{64}{80}
    \label{fig:supp:patch:4}
\end{figure*}
\begin{figure*}
    \figPatchesLabelSuppMat{Gilat_210809}{1}{2}{4}{6}{7}{9}
    \label{fig:supp:patch:5}
\end{figure*}
\begin{figure*}
    \figPatchesLabelSuppMat{180805_Peach}{2}{4}{5}{8}{17}{84}
    \label{fig:supp:patch:6}
\end{figure*}
\begin{figure*}
    \figPatchesLabelSuppMat{NeveYaar_210520}{7}{35}{73}{122}{179}{492}
    \label{fig:supp:patch:7}
\end{figure*}
\begin{figure*}
    \figPatchesLabelSuppMat{YanivReshef_190816}{11}{31}{51}{60}{80}{215}
    \label{fig:supp:patch:8}
\end{figure*}
\begin{figure*}
    \figPatchesLabelSuppMat{Gilat_210809}{94}{126}{133}{143}{10}{92}
    \label{fig:supp:patch:9}
\end{figure*}
\begin{figure*}
    \figPatchesLabelSuppMat{YanivReshef_190816}{479}{592}{295}{NeveYaar_210520_899}{MevoBytar_210818_0}{MevoBytar_210818_76}
    \label{fig:supp:patch:10}
\end{figure*}
\begin{figure*}
    \figPatchesLabelSuppMat{NirEliyho_211005}{103}{140}{168}{MevoBytar_210818_78}{Tzora_210523_61}{Tzora_210523_164}
    \label{fig:supp:patch:11}
\end{figure*}
\begin{figure*}
    \figPatchesFiveSuppMat{180805_Peach}{159}{161}{170}{152}
    \label{fig:supp:patch:12}
\end{figure*}

\end{document}